%% file: main.tex
\documentclass[10pt,twocolumn,letterpaper]{article}
\usepackage[pagenumbers]{cvpr}
\usepackage{graphicx}
\usepackage{amsmath}
\usepackage{amssymb}
\usepackage[pagebackref,breaklinks,colorlinks]{hyperref}
\usepackage{multirow}
\usepackage{makecell}
\usepackage{wrapfig,boxedminipage,lipsum}
\usepackage[export]{adjustbox}
\usepackage{graphicx}
\usepackage[dvipsnames]{xcolor}
\usepackage{balance}

\def\cvprPaperID{2974}
\def\confName{CVPR}
\def\confYear{2023}

\input{figures}
\newcommand{\Sref}[1]{Section \ref{#1}}
\newcommand{\Eref}[1]{Eq. \ref{#1}}
\newcommand{\Tref}[1]{Table \ref{#1}}
\newcommand{\Fref}[1]{Figure \ref{#1}}

\newcommand{\red}[1]{\textcolor{red}{#1}}

\begin{document}
\title{Interactive Cartoonization with Controllable Perceptual Factors}

\author{
Namhyuk Ahn\textsuperscript{1}
\quad
Patrick Kwon\textsuperscript{1}
\quad
Jihye Back\textsuperscript{1}
\quad
Kibeom Hong\textsuperscript{2}
\quad
Seungkwon Kim\textsuperscript{1}\\
\textsuperscript{1} NAVER WEBTOON AI \quad \textsuperscript{2} Yonsei University\\
}

\twocolumn[{
\renewcommand\twocolumn[1][]{#1}
\maketitle
\begin{center}
\centering
\captionsetup{type=figure}
\includegraphics[width=\linewidth, height=7cm]{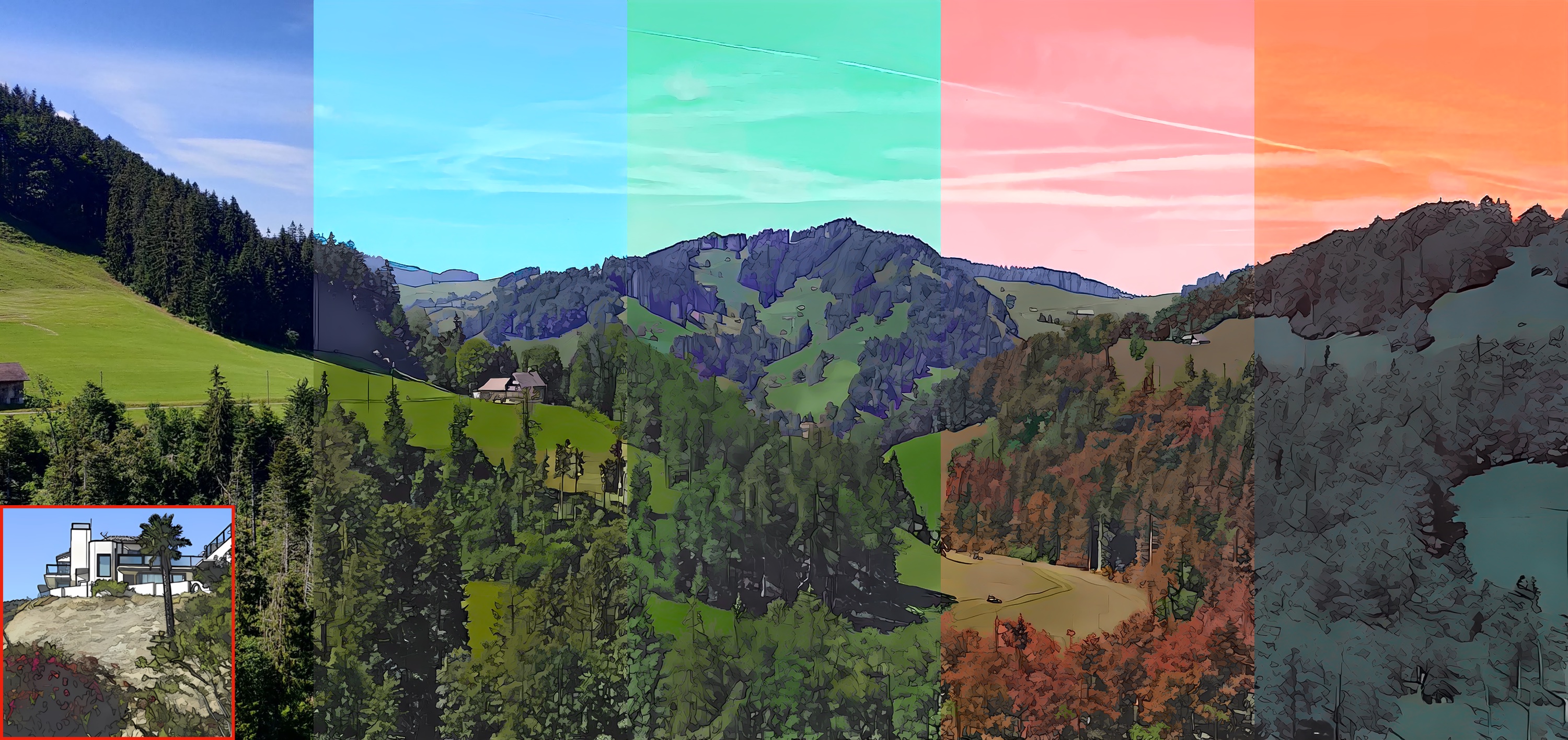}
\caption{\textbf{Interactive cartoonization.} The proposed cartoonization method allows a user interaction over texture and color, and can generate diverse outputs that meet users' demands. Leftmost: a source photography and an exemplar image of target cartoon.}
\label{fig:teaser}
\end{center}
}]

\input{sections/0_abstract}
\input{sections/1_introduction}
\input{sections/2_related_work}
\input{sections/3_method}
\input{sections/4_experiment}
\input{sections/5_application}
\input{sections/6_conclusion}

\clearpage
\newpage

\appendix
\input{sections/7_appendix}

\clearpage
\newpage
\balance

{
\small
\bibliographystyle{ieee_fullname}
\bibliography{}
}
\end{document}

%% file: figures.tex
\newcommand{\figBackgroundProcess}{
\begin{figure*}[t]
\newcommand{\h}{34.0mm}
\newcommand{\himg}{0.3mm}
\centering
\includegraphics[width=\h]{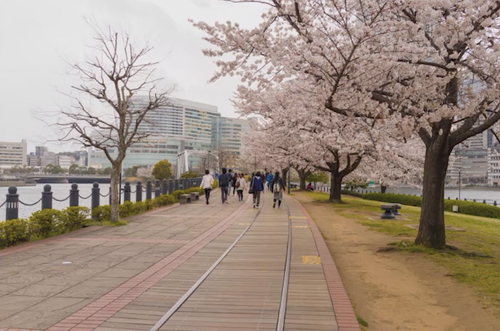}\hspace{\himg}
\includegraphics[width=\h]{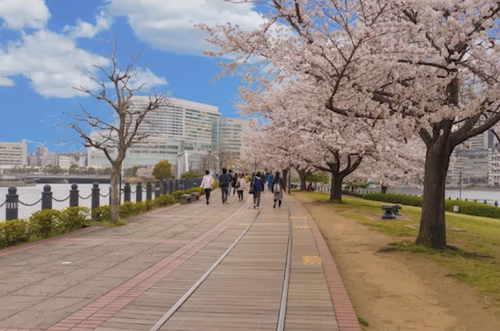}\hspace{\himg}
\includegraphics[width=\h]{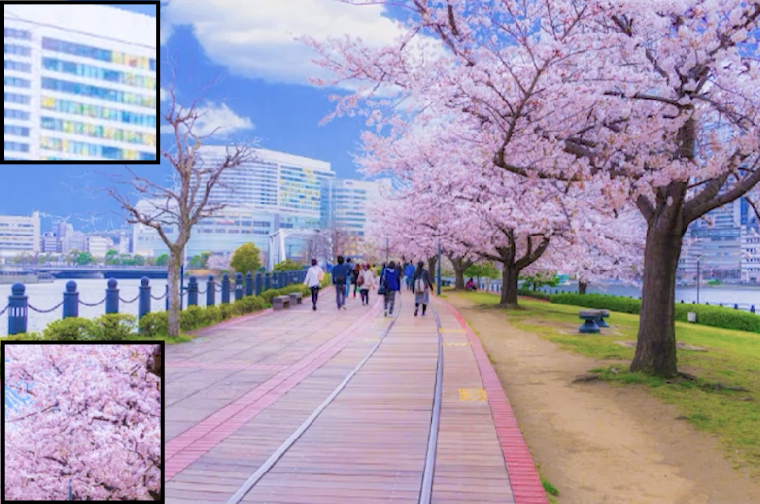}\hspace{\himg}
\includegraphics[width=\h]{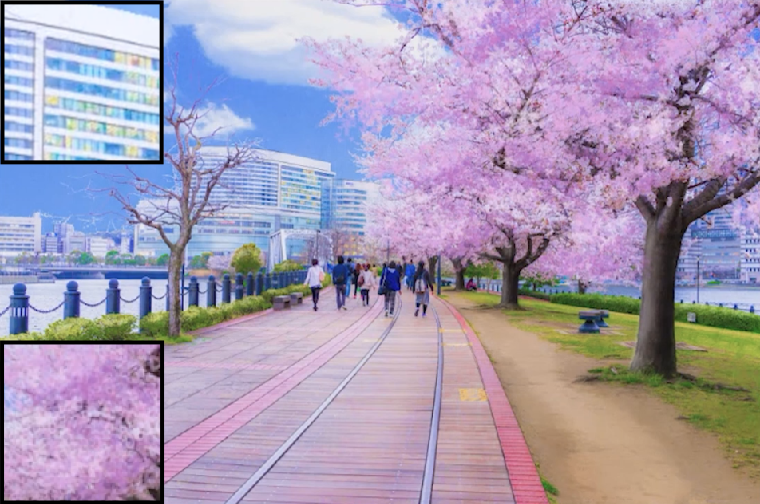}\hspace{\himg}
\includegraphics[width=\h]{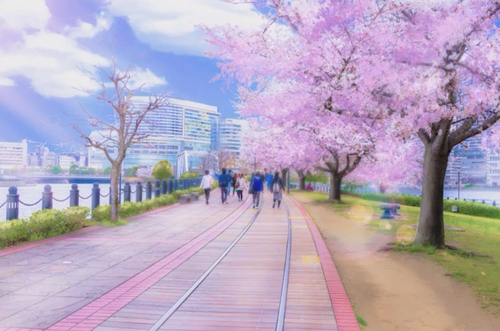}\hfill
\\
\makebox[\h][c]{(a) Source photo}\hspace{\himg}
\makebox[\h][c]{(b) Sky synthesis}\hspace{\himg}
\makebox[\h][c]{(c) Color stylization}\hspace{\himg}
\makebox[\h][c]{(d) Texture stylization}\hspace{\himg}
\makebox[\h][c]{(e) Post-processing}\hfill
\caption{\textbf{Example of the background making process.} We visualize how the artist creates the background scene by the step-by-step procedure. Note that some steps can be skipped or changed in the order depending on the artist. \textcopyright Kawaii Studio}
\label{fig:background_process}
\end{figure*}
}

\newcommand{\figModelOverview}{
\begin{figure}[t]
\centering
\includegraphics[width=\linewidth]{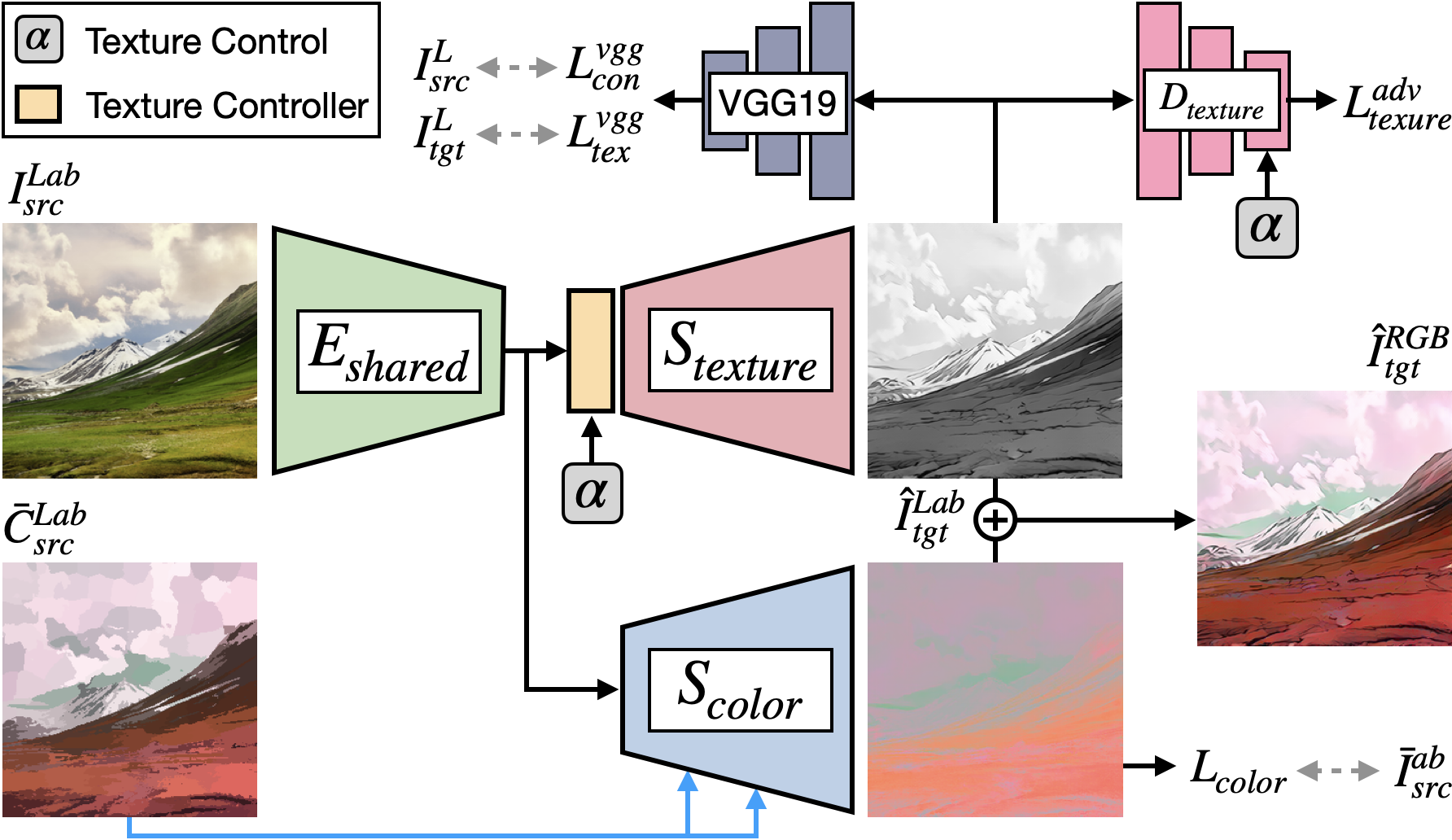}
\caption{\textbf{Model overview.} Given a photo, \textsc{Cartooner} estimates the stylized texture and color images, which are then composed for the final product. We design decomposed texture/color paths, a texture controller, and a multi-texture discriminator for interaction.}
\label{fig:model_overview}
\end{figure}
}

\newcommand{\figPreAnalysis}{
\begin{figure}[t]
\centering
\includegraphics[width=\linewidth]{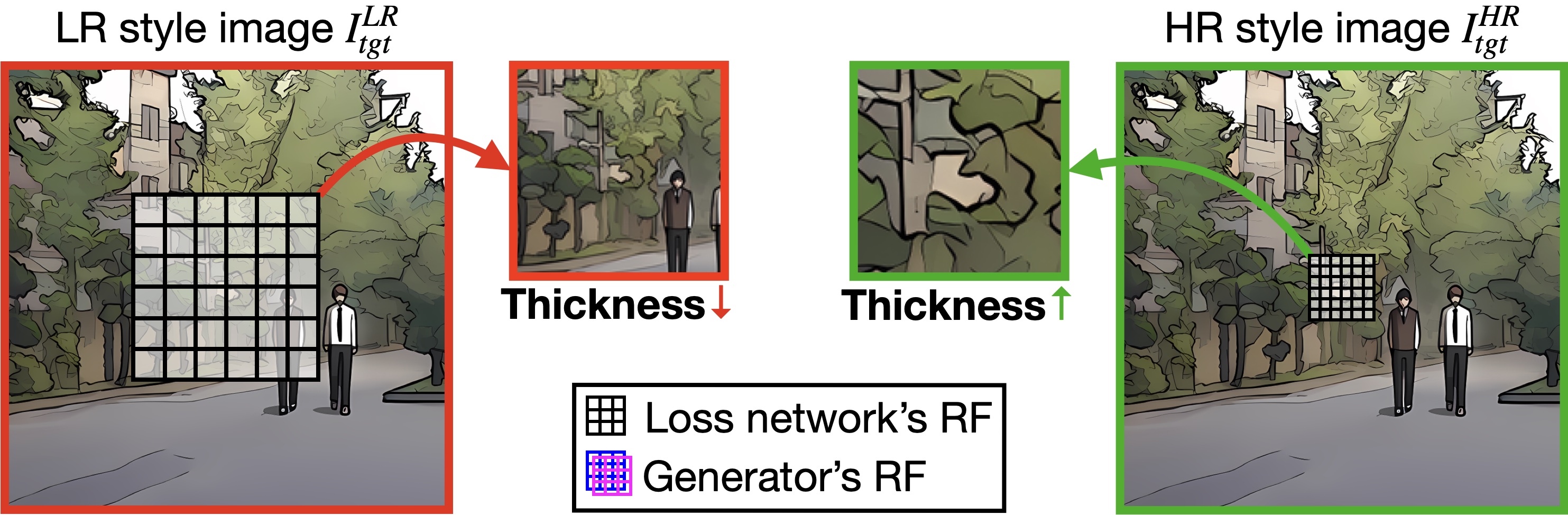}\\
\makebox[\linewidth][c]{(a) Stroke thickness}\\
\includegraphics[width=\linewidth]{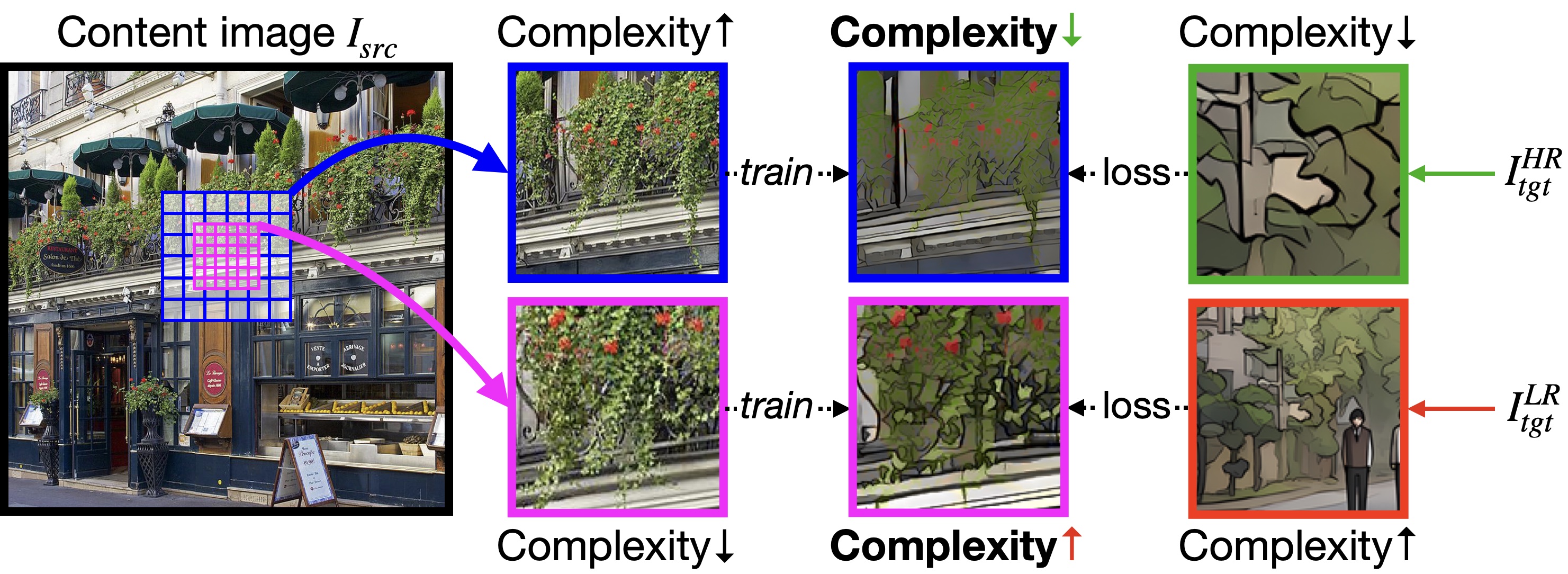}
\makebox[\linewidth][c]{(b) Abstraction}\\
\caption{\textbf{Why affects texture levels?} (\textbf{a}) With a fixed receptive field (RF) of loss network, as the resolution of target domain images changes, the stroke thickness within an RF window varies. (\textbf{b}) With a fixed resolution of a content image, as the RF of the generator changes, the scene complexity within an RF window varies. When we provide style images with disparate complexity, the generator alters the scene complexity of the cartoonized results.}
\label{fig:preanalysis}
\end{figure}
}

\newcommand{\figResultPreAnalysis}{
\begin{figure}[t]
\centering
\includegraphics[width=\linewidth]{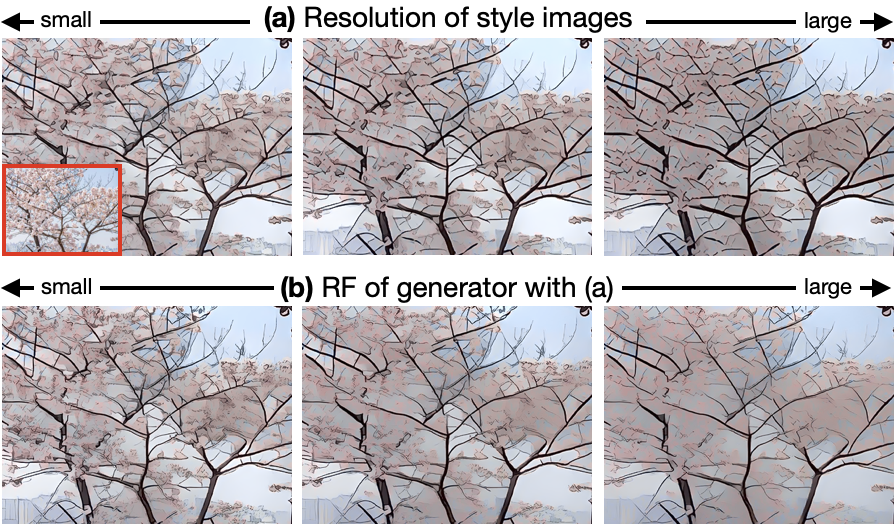}
\caption{\textbf{What affects texture levels?} (\textbf{a}) As the resolution of target cartoon (style) images is increased, the stroke is thickened. (\textbf{b}) As both the resolution of images and the receptive field (RF) of the generator is increased, the abstraction becomes a high degree.}
\label{fig:result_preanalysis}
\end{figure}
}

\newcommand{\figTextureController}{
\begin{figure}[t]
\centering
\includegraphics[width=\linewidth]{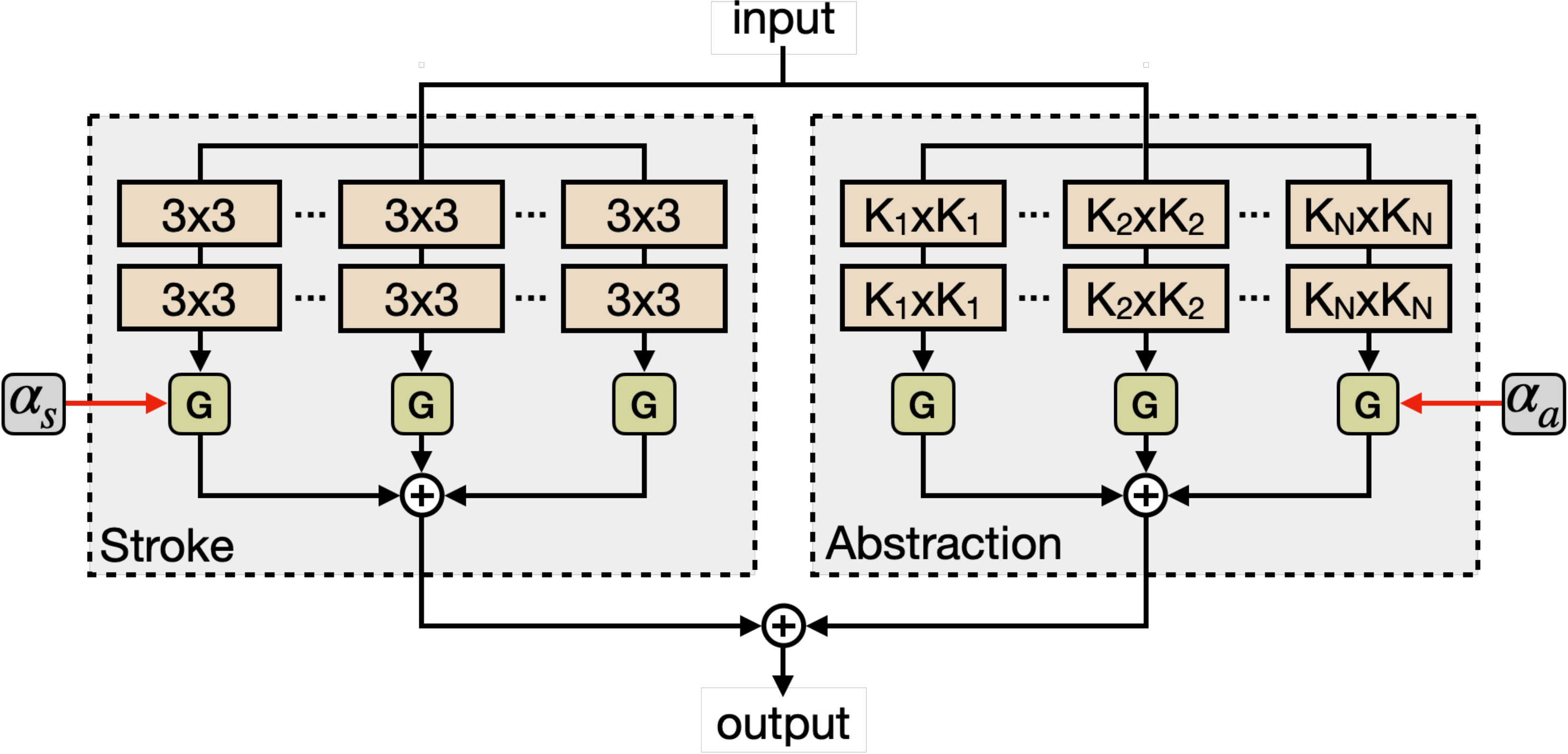}
\caption{\textbf{Texture controller.} This module consists of the stroke and abstraction control units, which both are designed as multi-branch. Intermediate features from branches are fused through a gating unit, which is controlled by texture factors, $\{\alpha_s, \alpha_a\}$. In the abstraction unit, kernel sizes of $K_1,...,K_N$ are increasing order.}
\label{fig:texture_controller}
\end{figure}
}

\newcommand{\figPreprocessTrain}{
\begin{figure}[t]
\centering
\includegraphics[width=\linewidth]{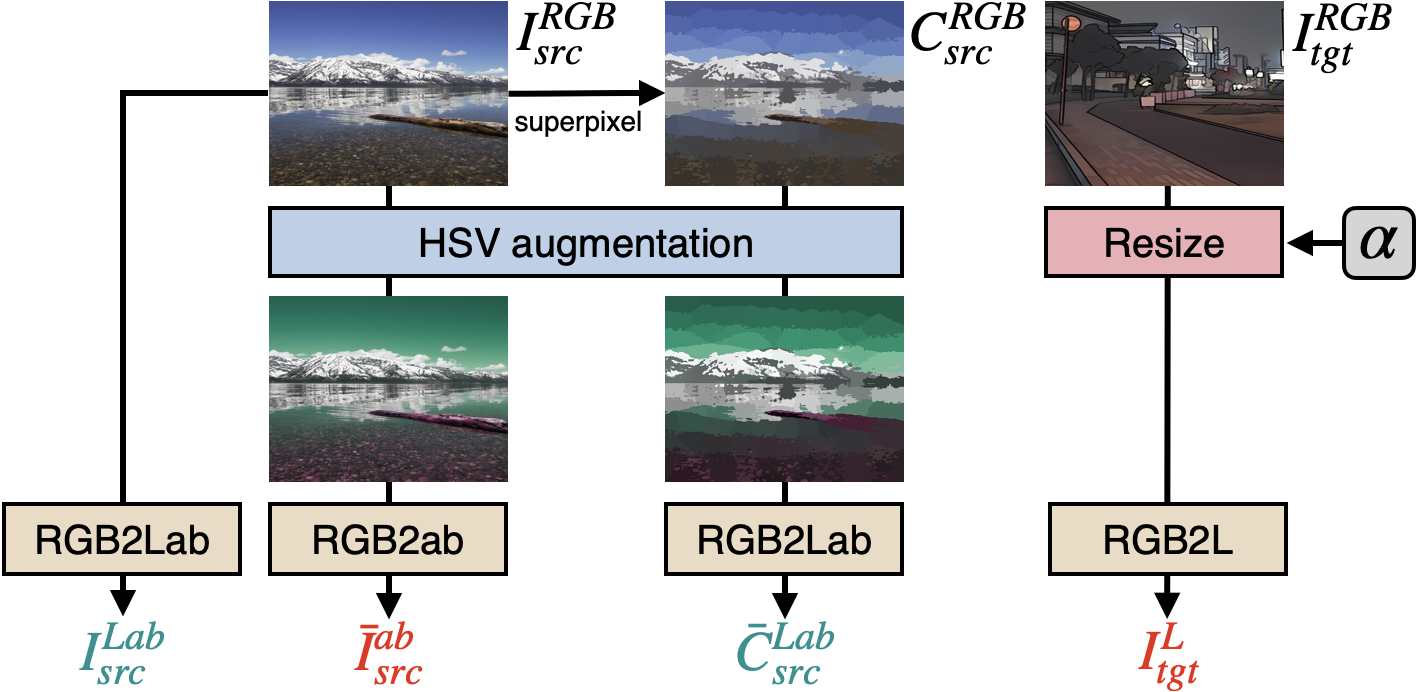}
\vspace{-1.4em}
\caption{\textbf{Data preprocessing on training.} Given a source photo $I^{RGB}_{src}$, we prepare an input color map $C^{RGB}_{src}$. Then, the HSV augmentation is operated on both $I^{RGB}_{src}$ and $C^{RGB}_{src}$. Target domain (cartoon) images $I^{RGB}_{tgt}$ are resized by referring to the prefixed texture control levels $\pmb{\alpha}$. All the processed images are then converted to Lab, L, or ab space. The \textcolor{ForestGreen}{green symbols} indicate the input data and \textcolor{red}{red ones} represent data used in the loss calculation.}
\label{fig:preprocess_train}
\end{figure}
}

\newcommand{\figCompQual}{
\begin{figure*}[t]
\newcommand{\h}{34mm}
\newcommand{\himg}{0.0mm}
\centering
\includegraphics[width=\h]{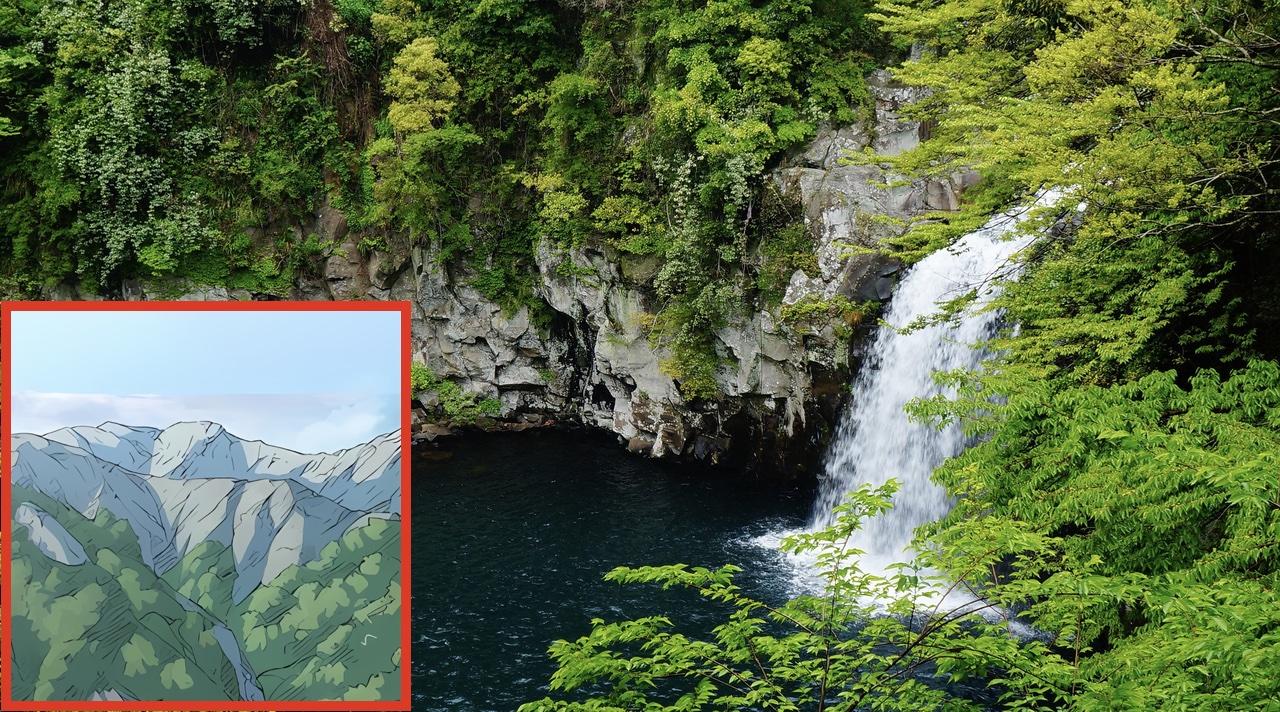}\hspace{\himg}
\includegraphics[width=\h]{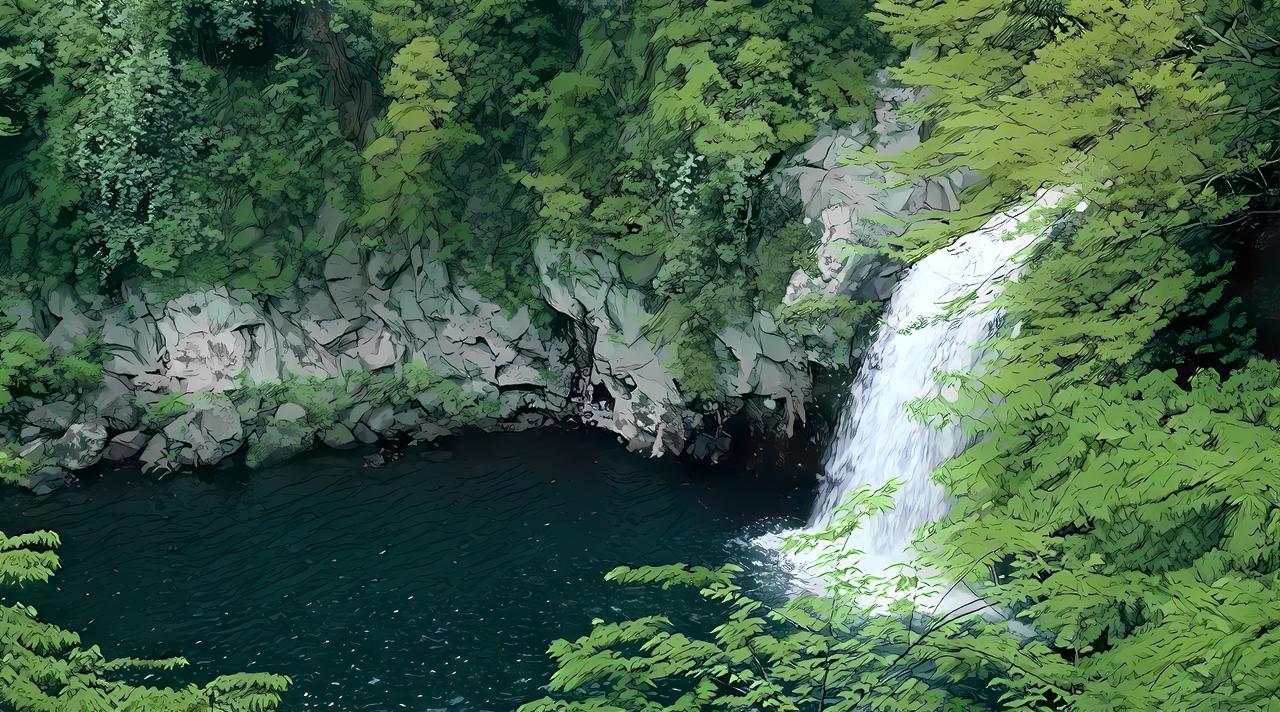}\hspace{\himg}
\includegraphics[width=\h]{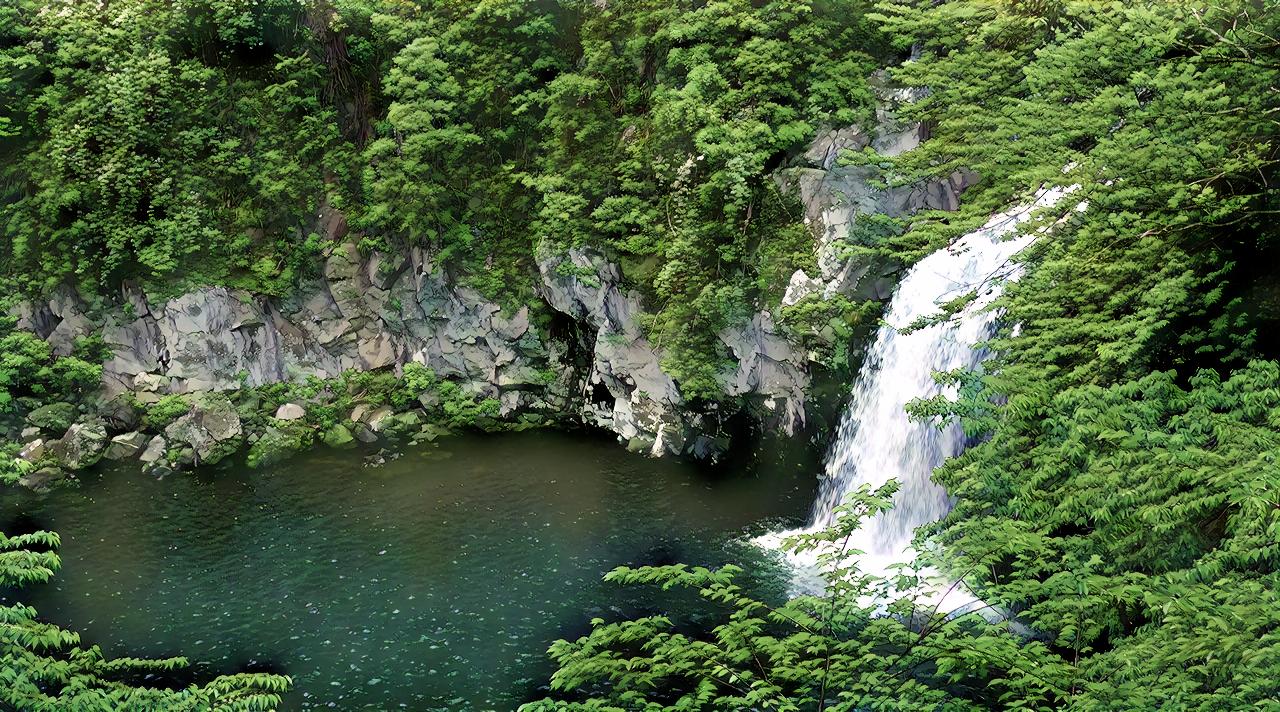}\hspace{\himg}
\includegraphics[width=\h]{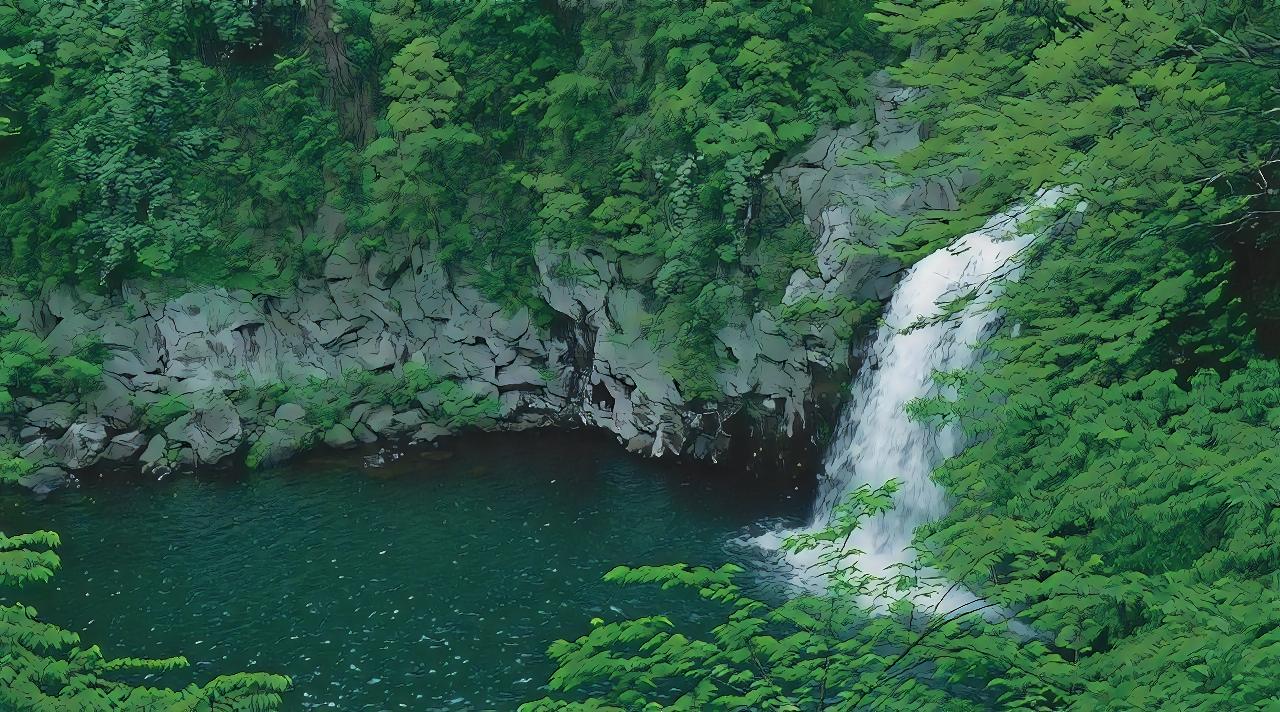}\hspace{\himg}
\includegraphics[width=\h]{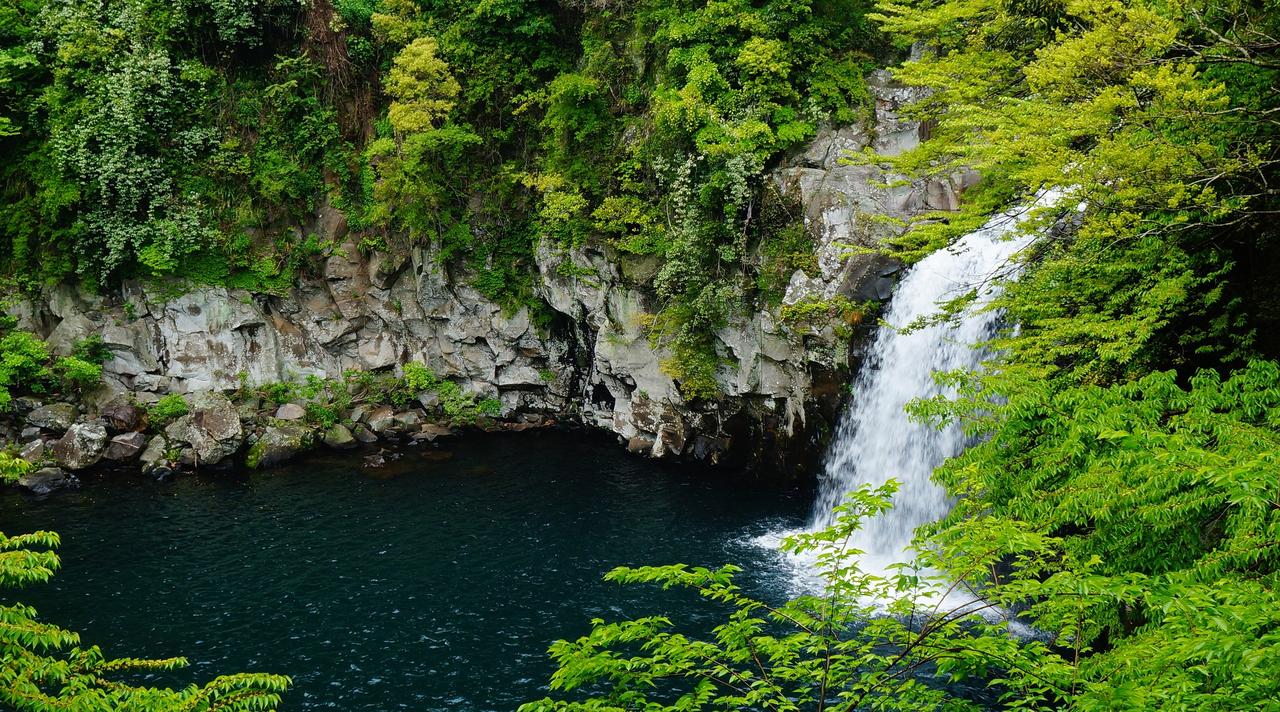}\hfill
\\
\includegraphics[width=\h]{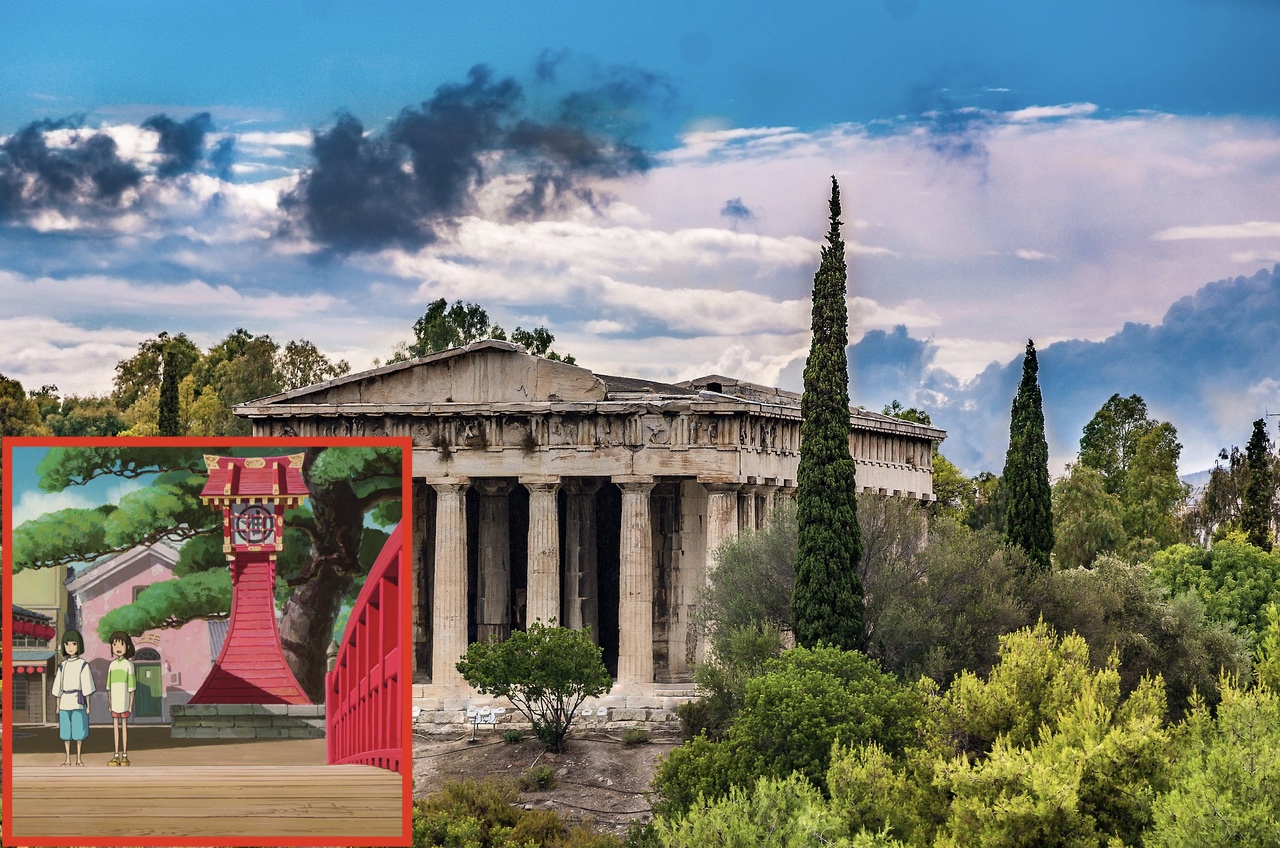}\hspace{\himg}
\includegraphics[width=\h]{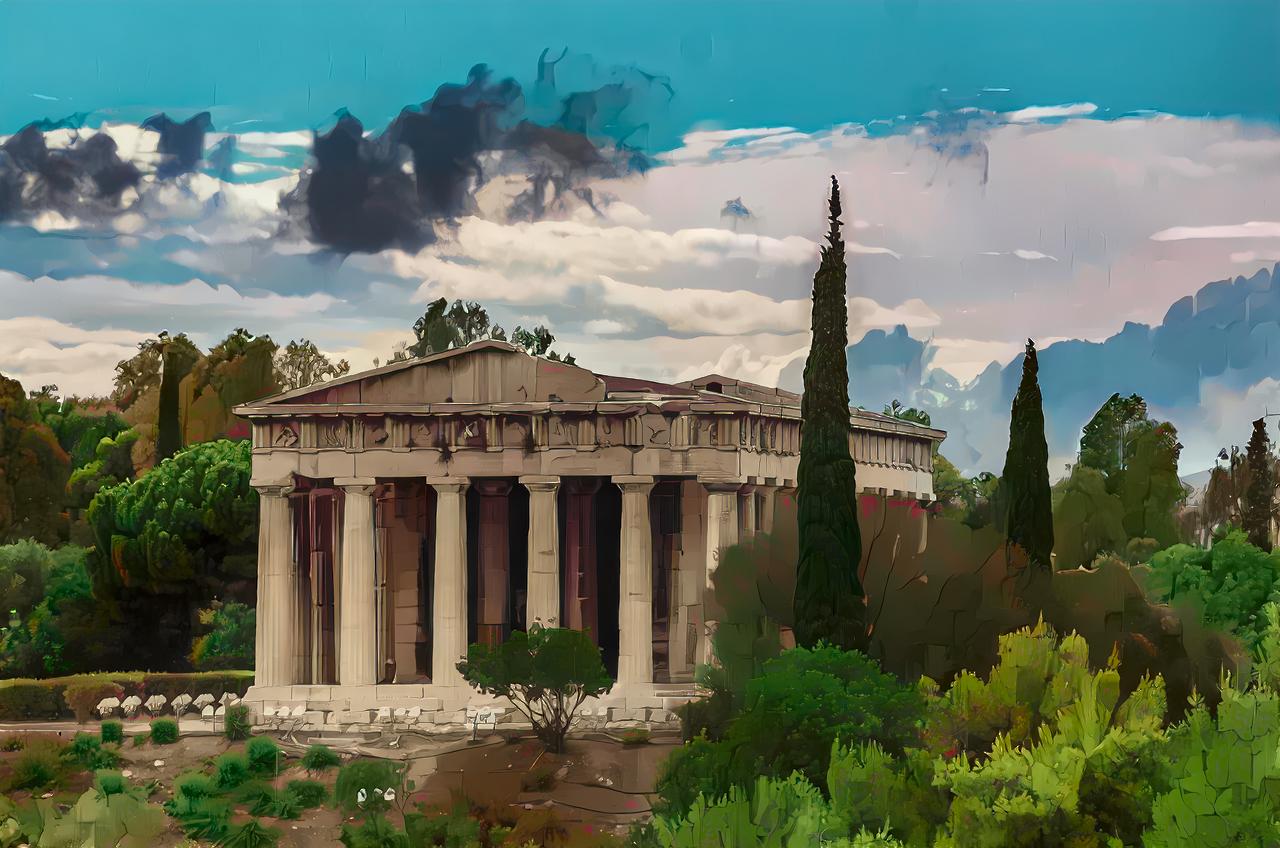}\hspace{\himg}
\includegraphics[width=\h]{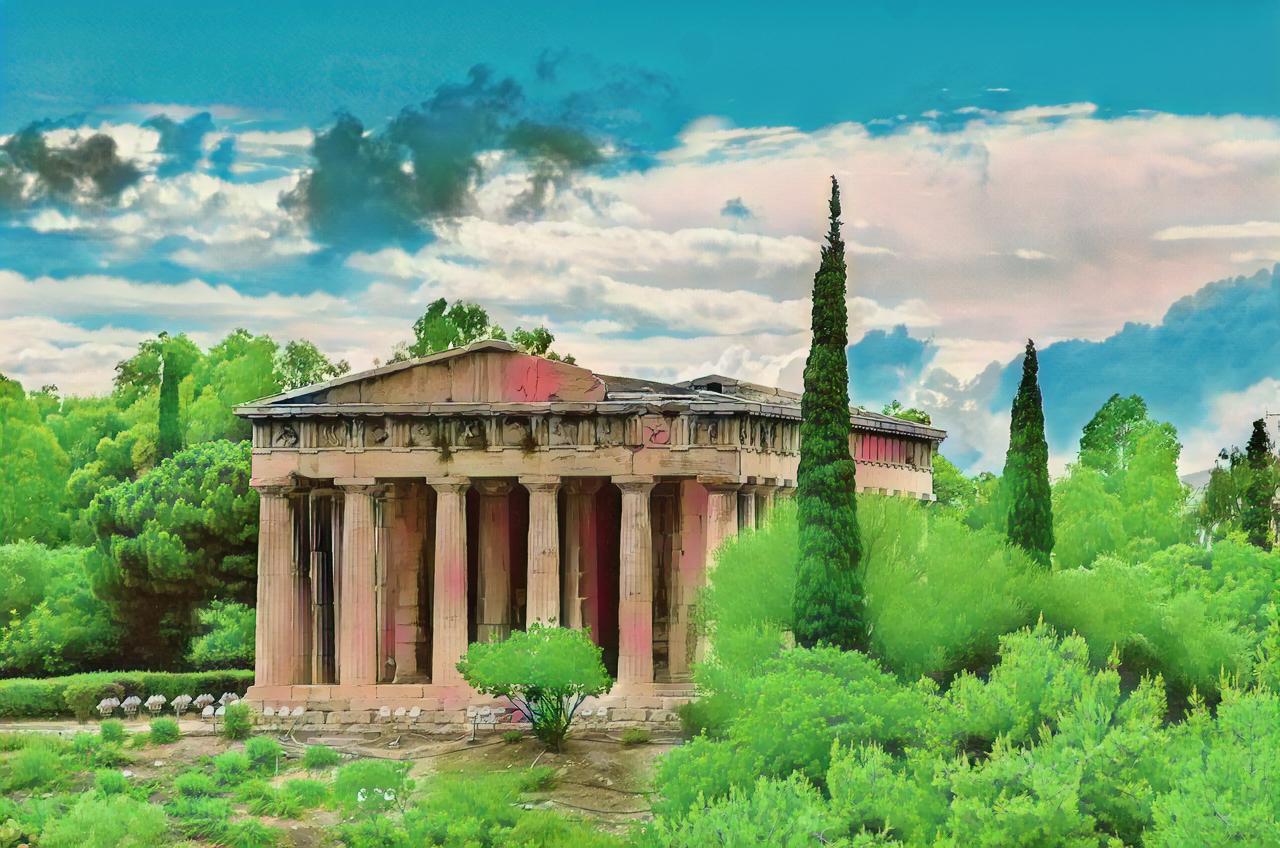}\hspace{\himg}
\includegraphics[width=\h]{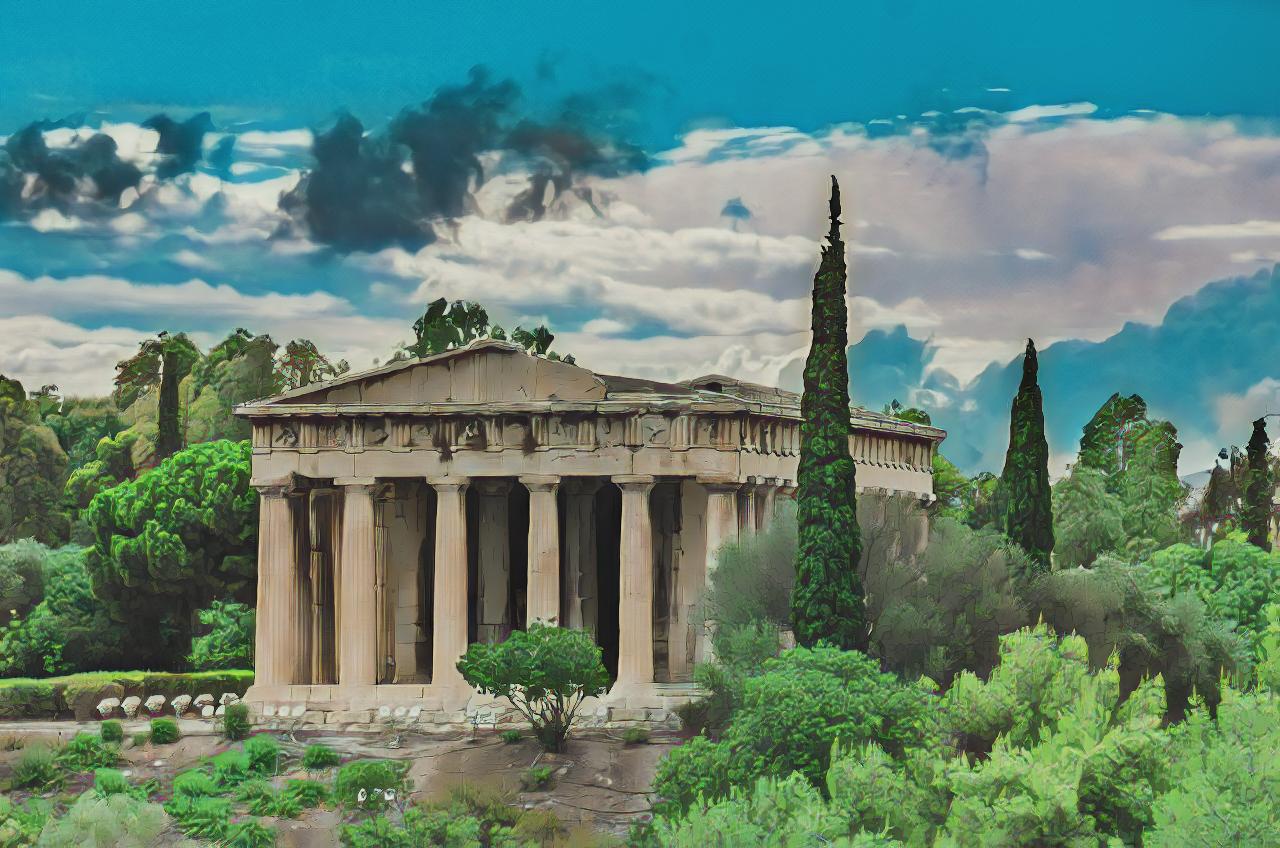}\hspace{\himg}
\includegraphics[width=\h]{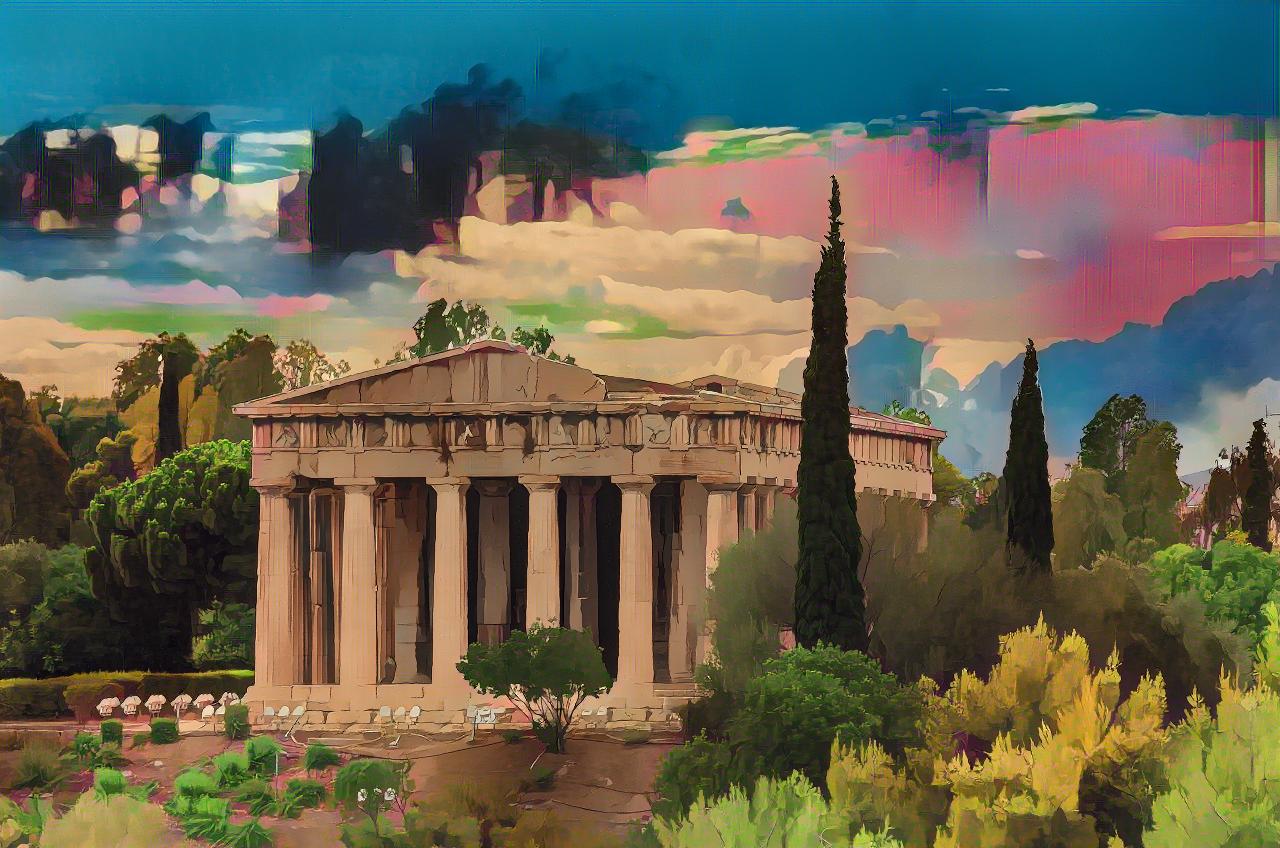}\hfill
\\
\includegraphics[width=\h]{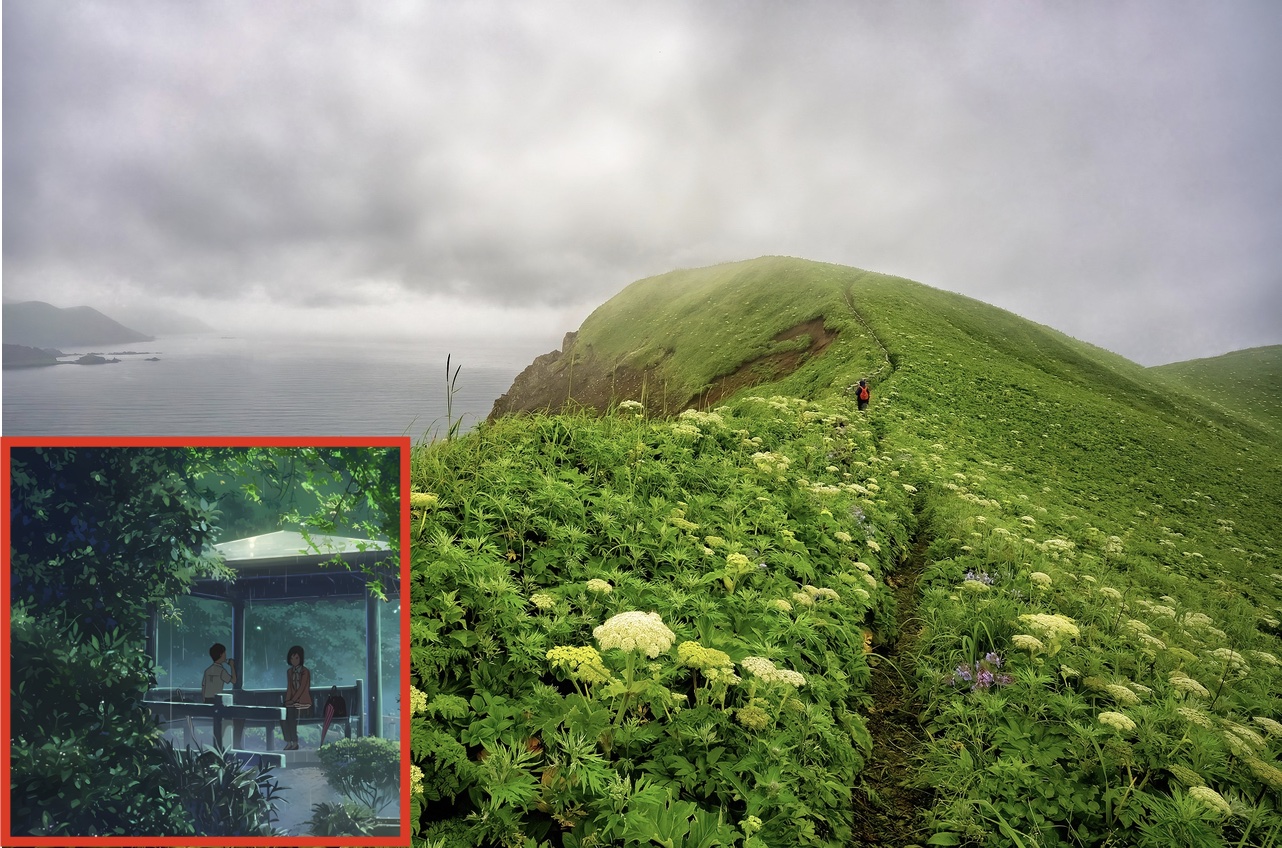}\hspace{\himg}
\includegraphics[width=\h]{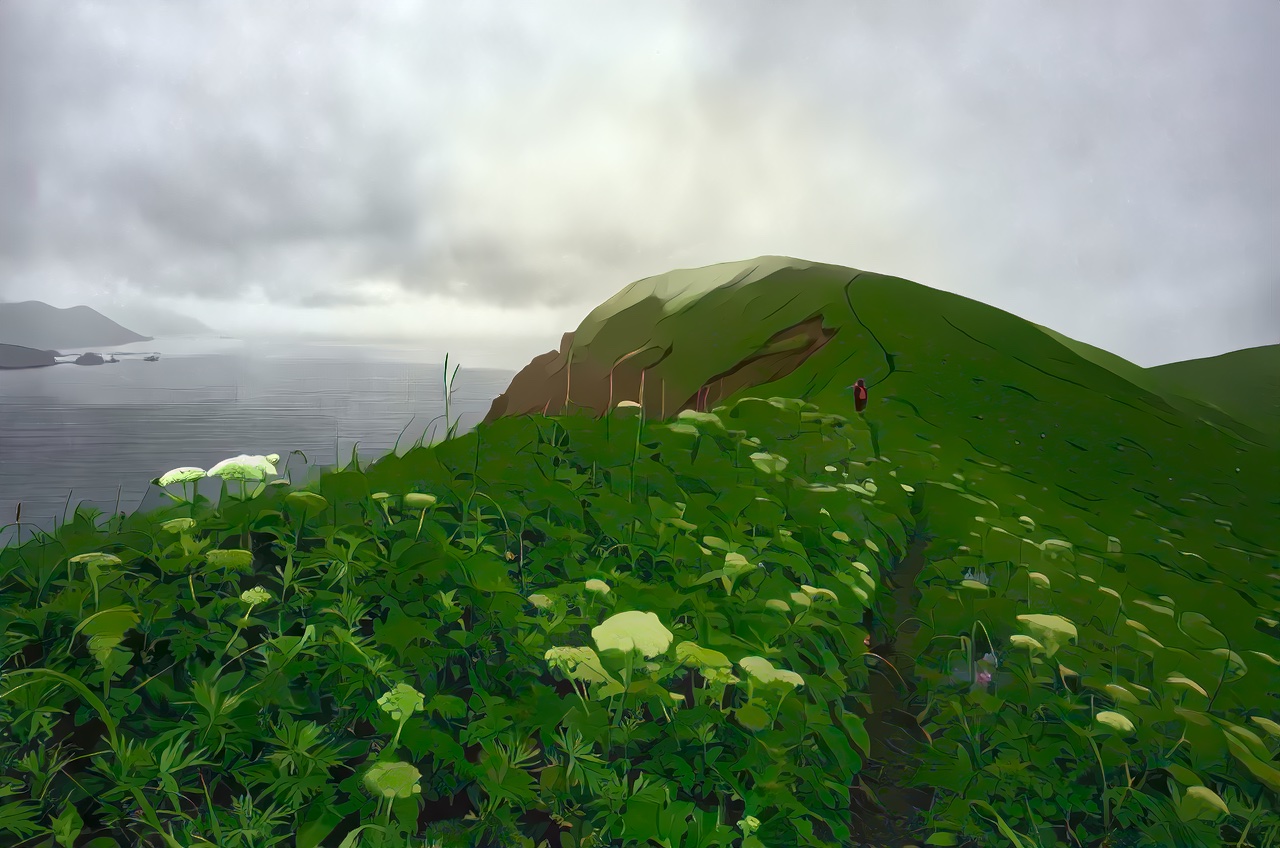}\hspace{\himg}
\includegraphics[width=\h]{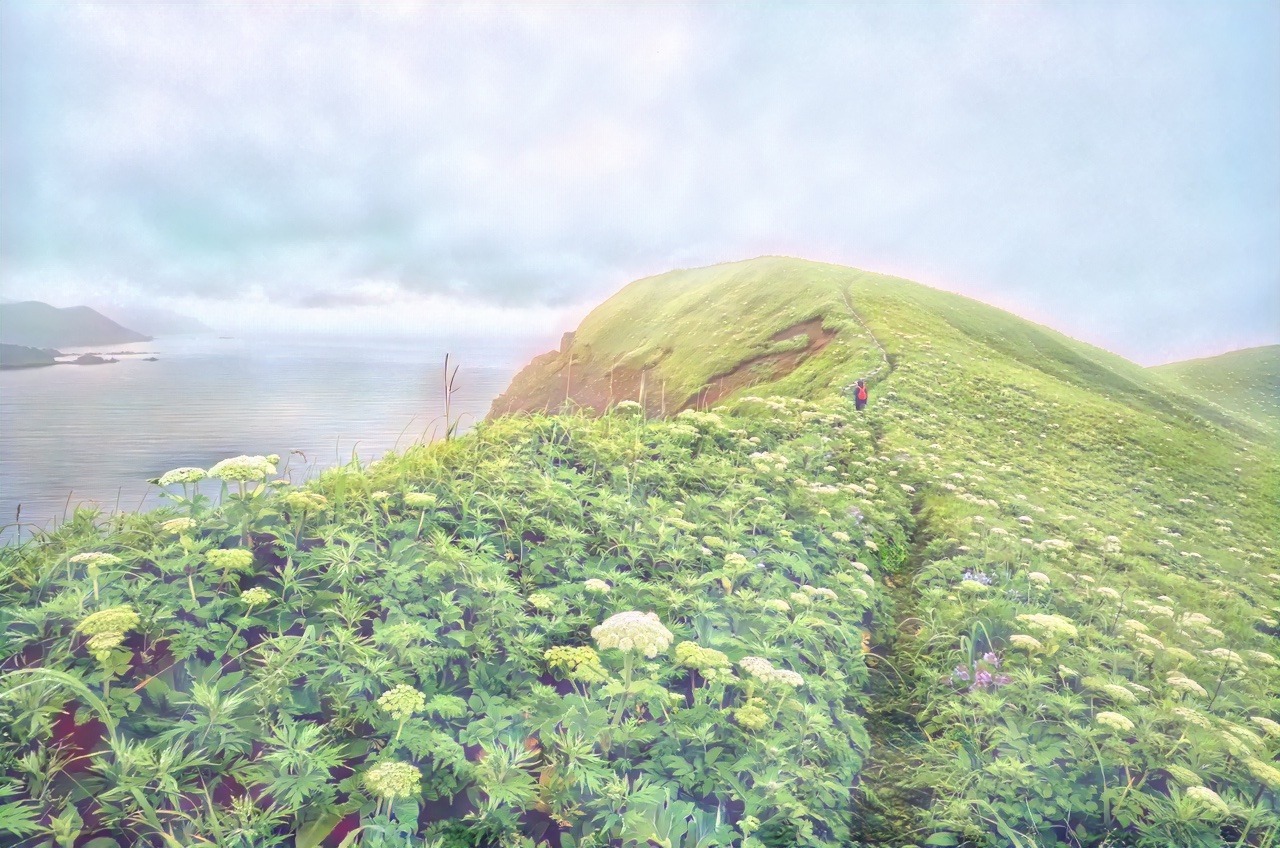}\hspace{\himg}
\includegraphics[width=\h]{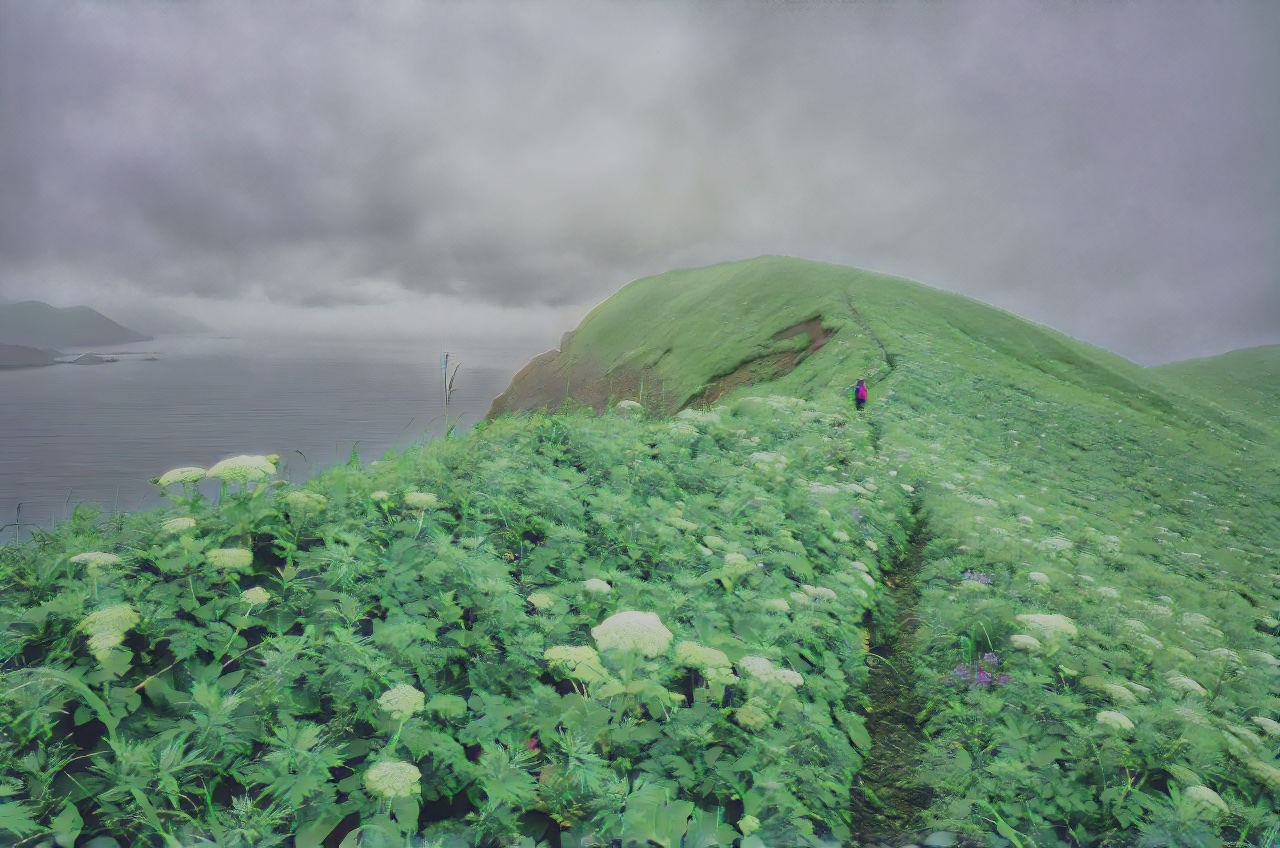}\hspace{\himg}
\includegraphics[width=\h]{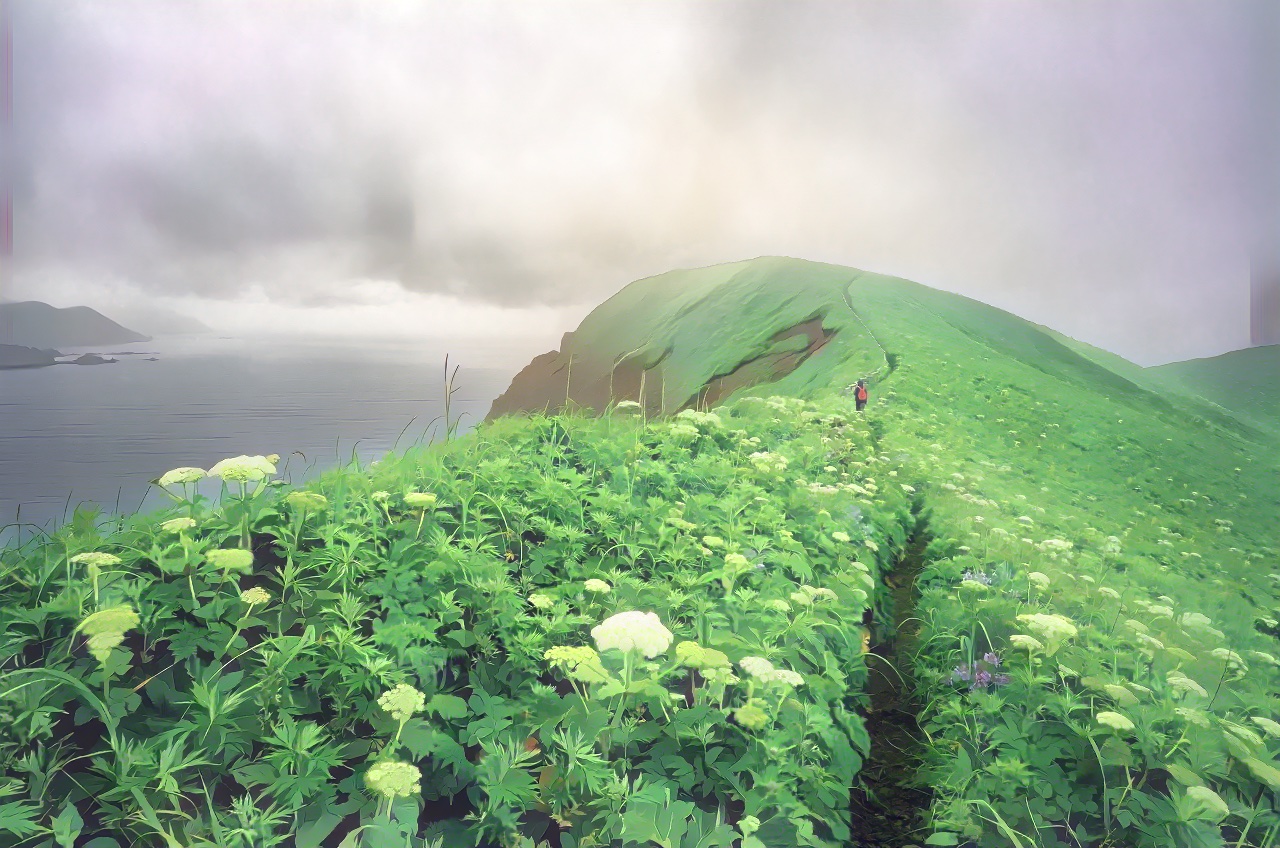}\hfill
\\
\includegraphics[width=\h]{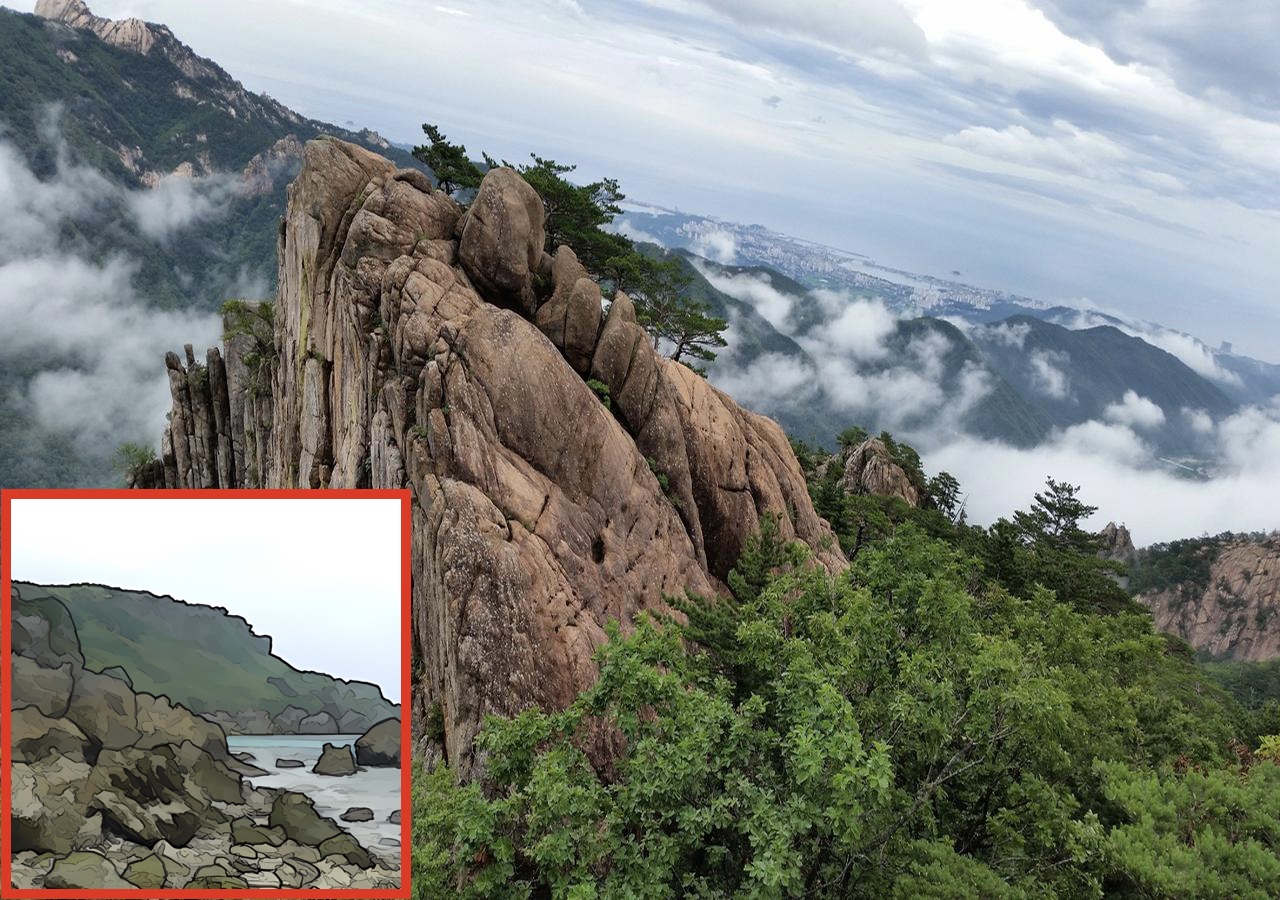}\hspace{\himg}
\includegraphics[width=\h]{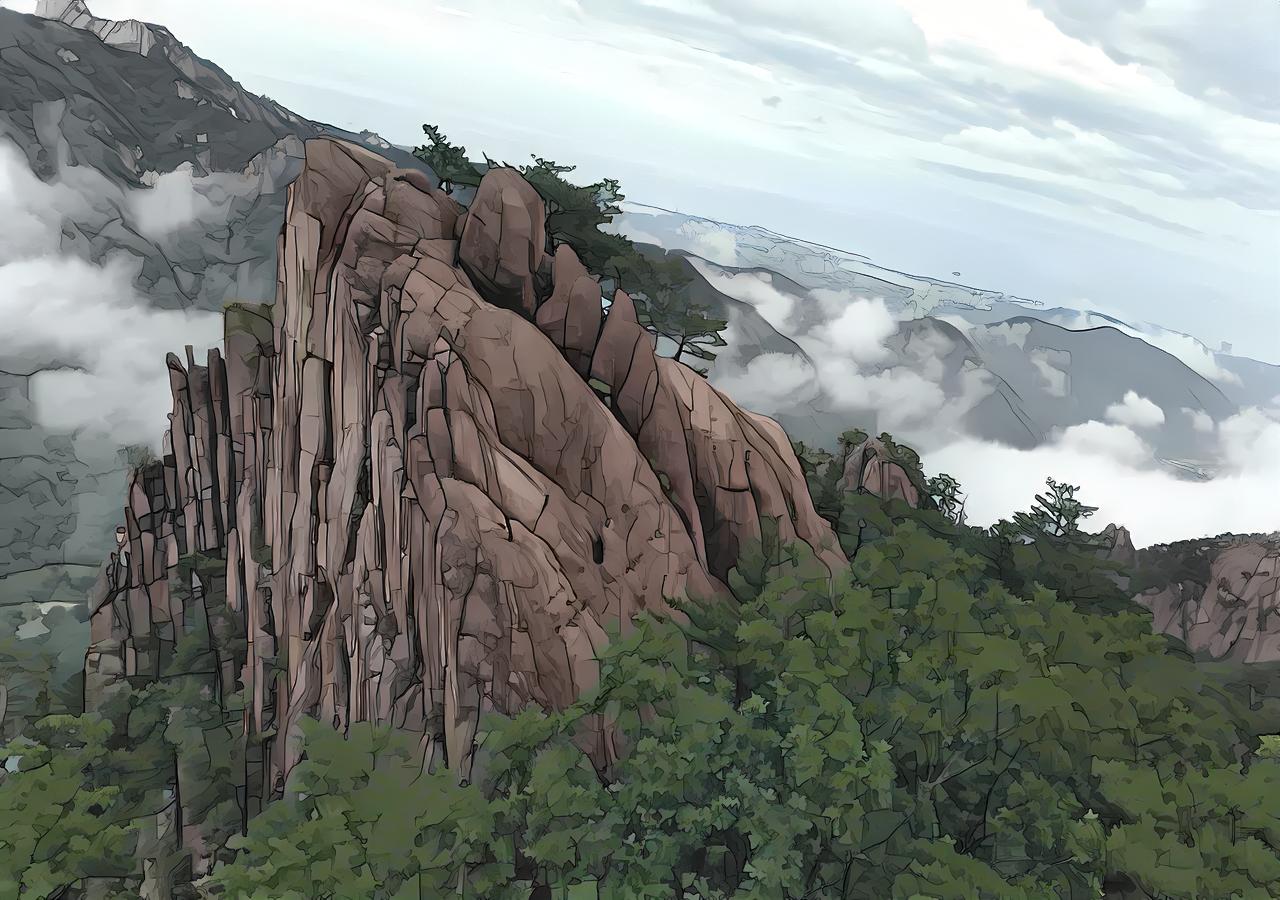}\hspace{\himg}
\includegraphics[width=\h]{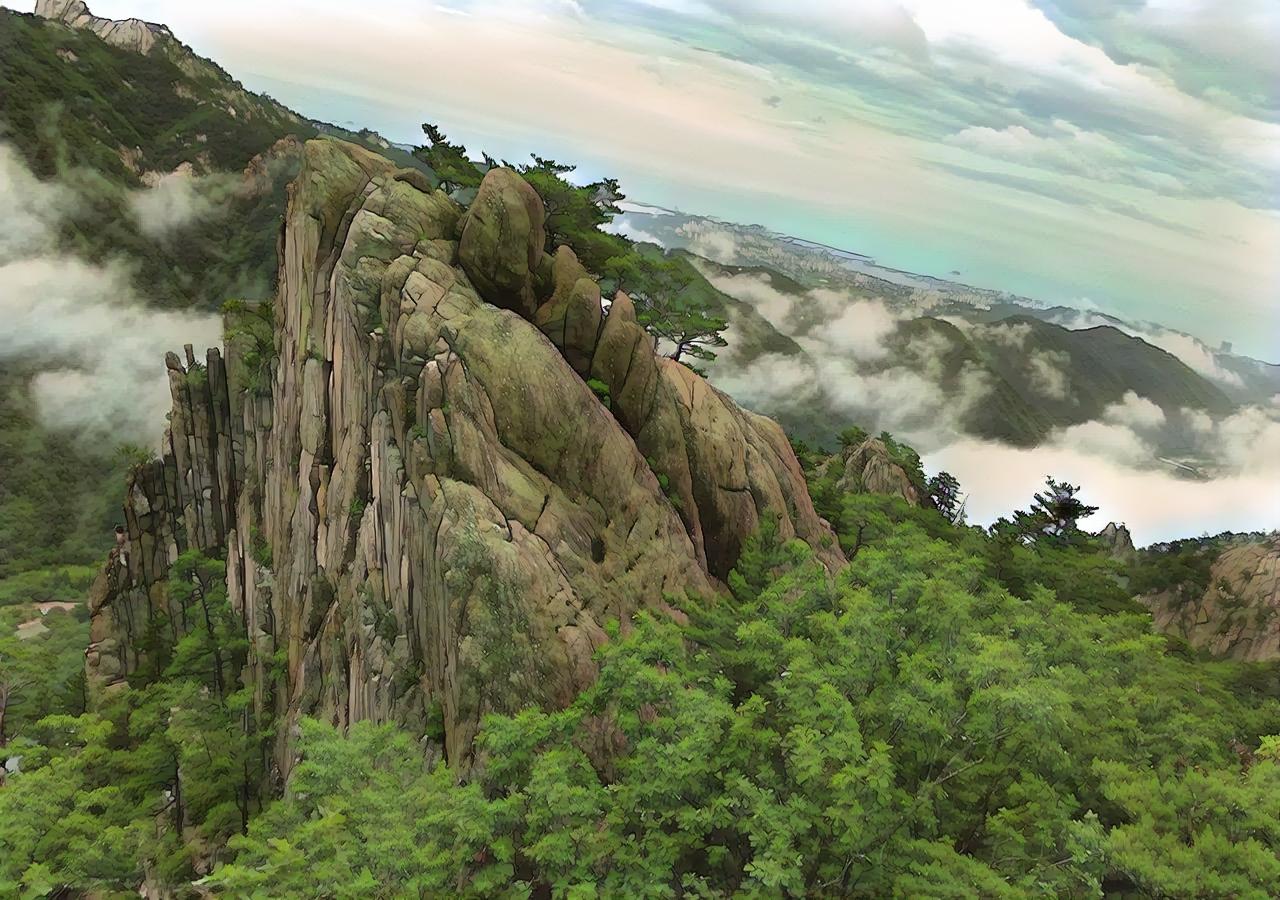}\hspace{\himg}
\includegraphics[width=\h]{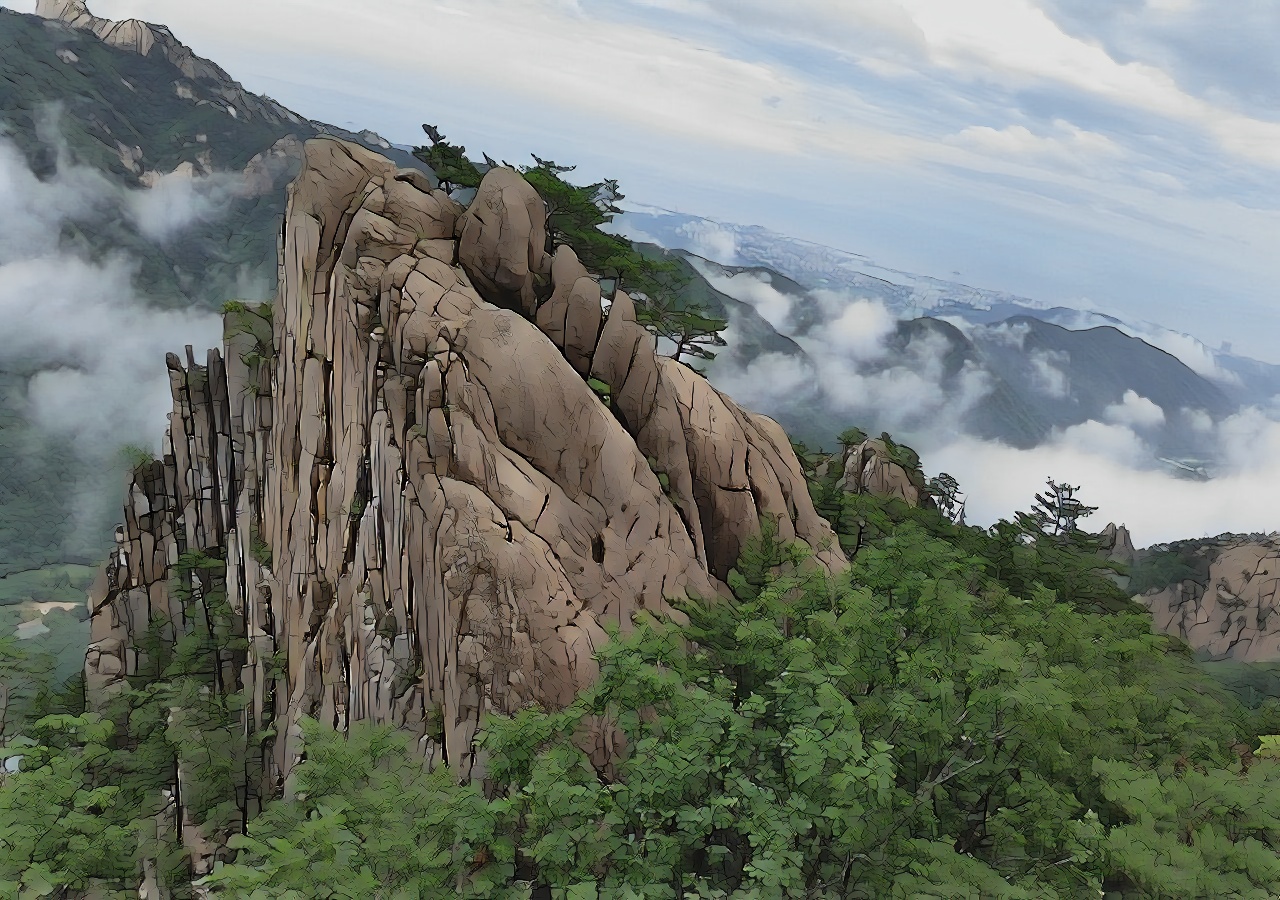}\hspace{\himg}
\includegraphics[width=\h]{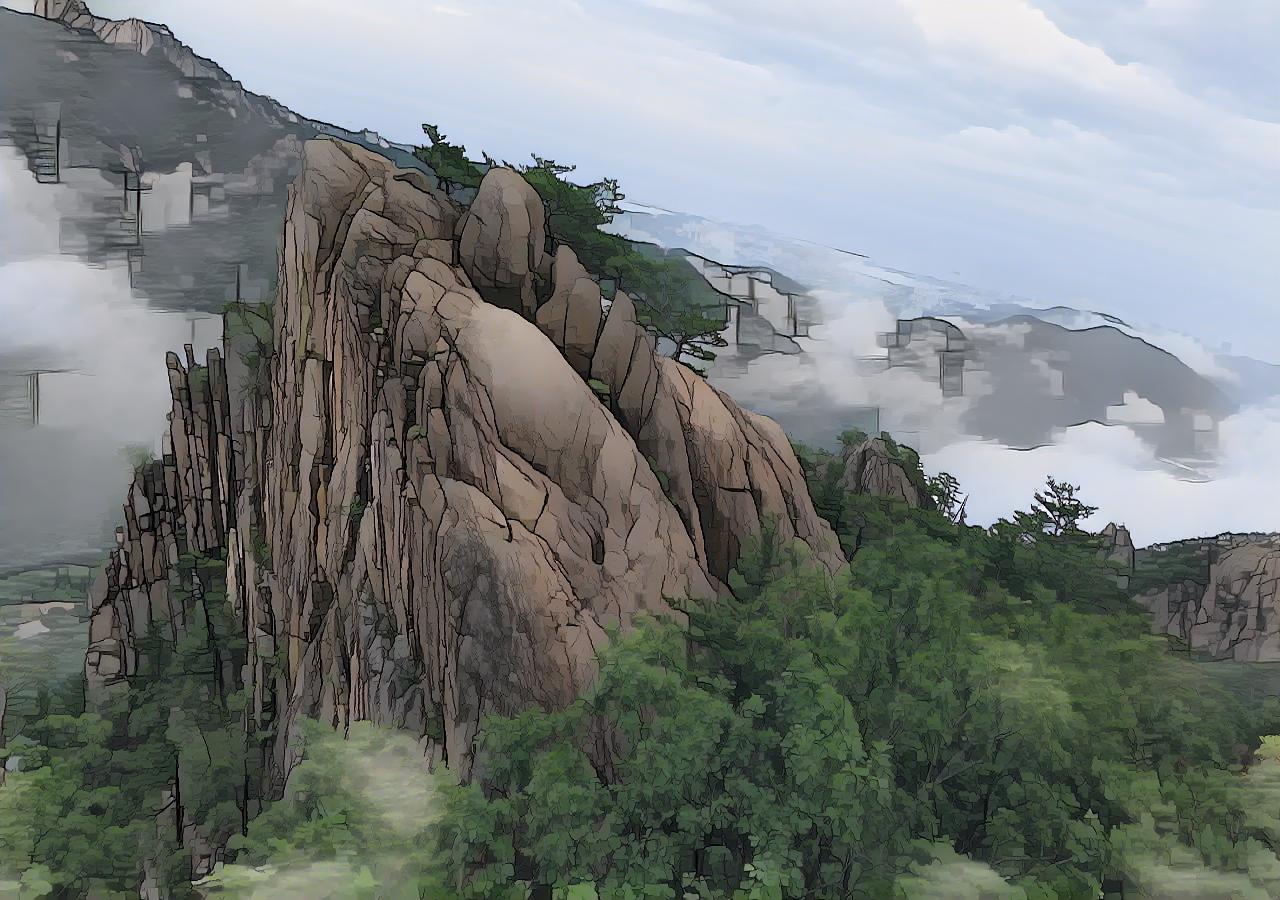}\hfill
\\
\makebox[\h][c]{(a) Source photo}\hspace{\himg}
\makebox[\h][c]{(b) \textsc{Cartooner}}\hspace{\himg}
\makebox[\h][c]{(c) CartoonGAN~\cite{chen2018cartoongan}}\hspace{\himg}
\makebox[\h][c]{(d) AnimeGANv2~\cite{chen2019animegan}}\hspace{\himg}
\makebox[\h][c]{(e) WhiteboxGAN~\cite{wang2020learning}}\hfill
\caption{\textbf{Visual comparison.} Images along with source photos indicate exemplar images of the target cartoon. Best viewed in zoom.}
\label{fig:comp_qual}
\end{figure*}
}

\newcommand{\figFeatmap}{
\begin{figure}[t]
\newcommand{\h}{27mm}
\newcommand{\himg}{0.0mm}
\newcommand{\hh}{20mm}
\centering
\includegraphics[width=\h, height=\hh]{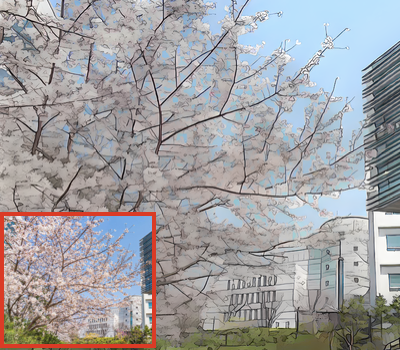}\hspace{\himg}
\includegraphics[width=\h, height=\hh]{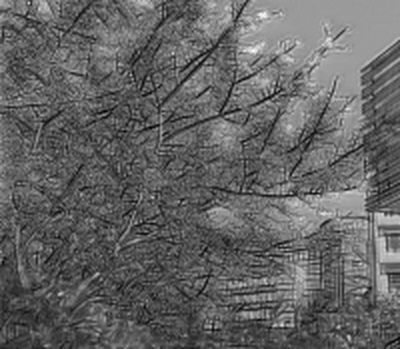}\hspace{\himg}
\includegraphics[width=\h, height=\hh]{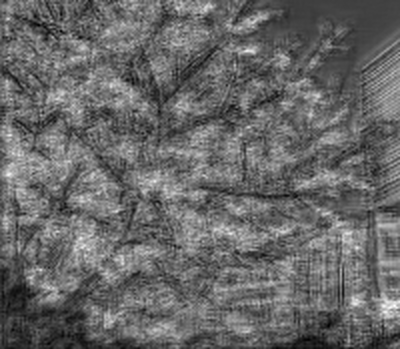}\hfill
\\
\includegraphics[width=\h, height=\hh]{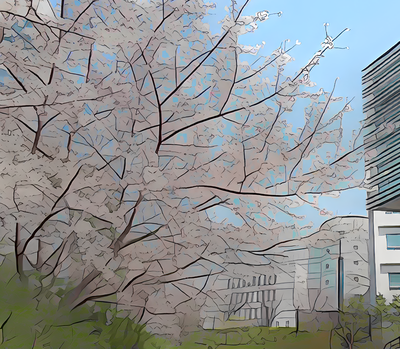}\hspace{\himg}
\includegraphics[width=\h, height=\hh]{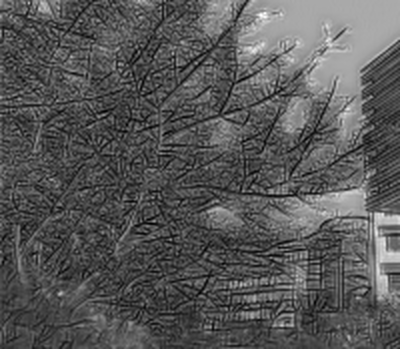}\hspace{\himg}
\includegraphics[width=\h, height=\hh]{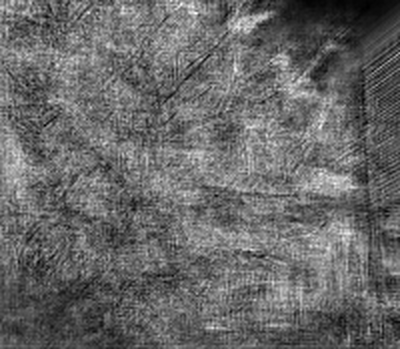}\hfill
\\
\makebox[\h][c]{(a) Result}\hspace{\himg}
\makebox[\h][c]{(b) Stroke}\hspace{\himg}
\makebox[\h][c]{(c) Abstraction}\hspace{\himg}
\caption{\textbf{Output feature of texture controller.} (\textbf{Top}) Lower stroke thickness and abstraction levels. (\textbf{Bottom}) Higher texture levels. The stroke unit focuses on fine details regards to edge region while the abstraction unit concentrates on the overall region.}
\label{fig:featmap}
\end{figure}
}

\newcommand{\figAblStroke}{
\begin{figure}[t]
\centering
\includegraphics[width=\linewidth]{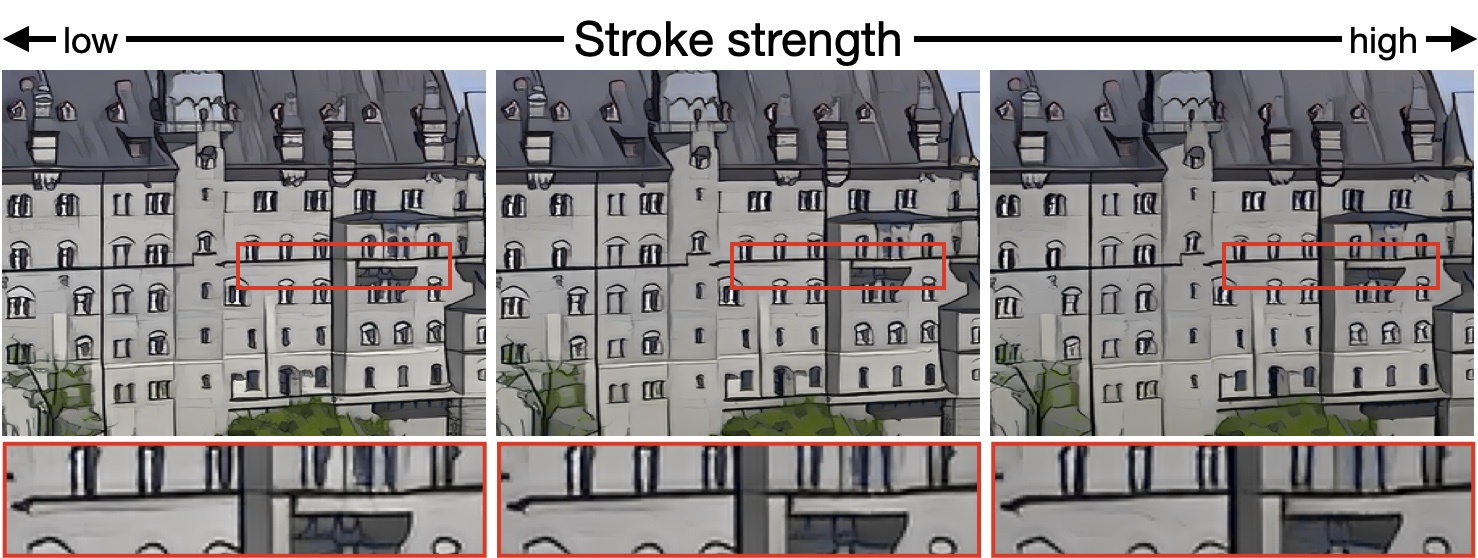}
\\
\makebox[\linewidth][c]{(a) \# levels = 2}
\\
\includegraphics[width=\linewidth]{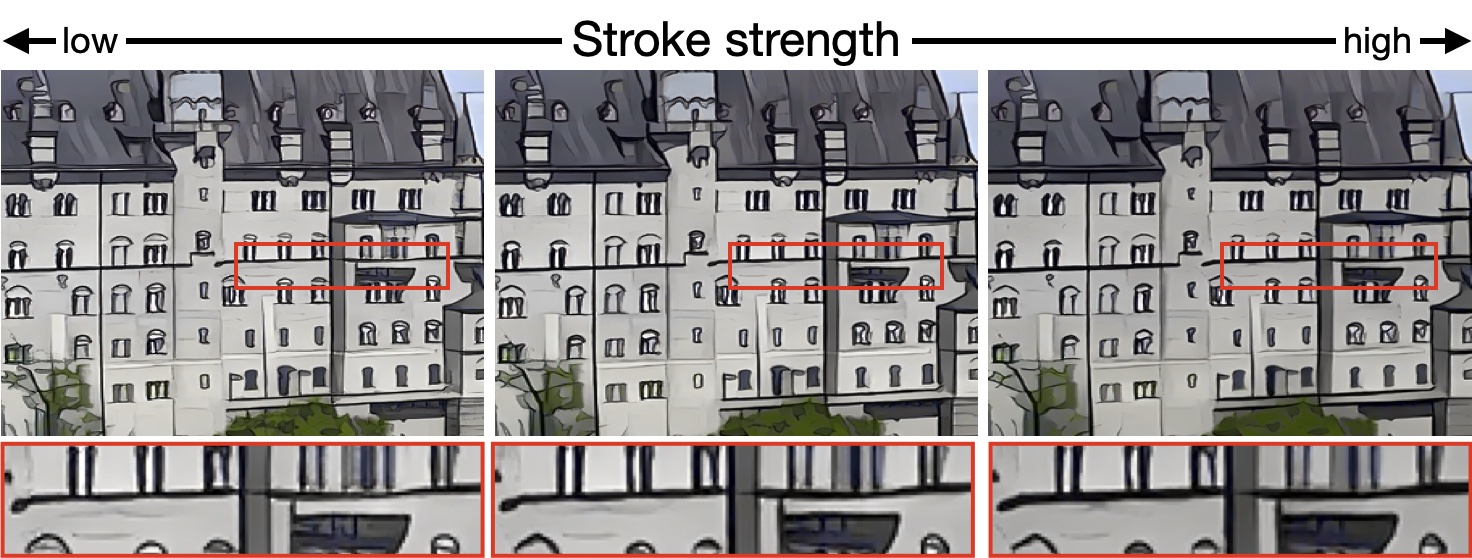}
\\
\makebox[\linewidth][c]{(b) \# levels = 5 (ours)}
\\
\caption{\textbf{Study on the stroke levels.} We compare the model trained with 2-level and 5-level stroke thickness. The 2-level case cannot adequately reflect the subtle change in stroke thickness.}
\label{fig:abl_stroke}
\end{figure}
}

\newcommand{\figAblColor}{
\begin{figure}[t]
\centering
\newcommand{\h}{27mm}
\newcommand{\hh}{40.3mm}
\newcommand{\himg}{0.0mm}
\centering
\includegraphics[width=\h, height=\hh]{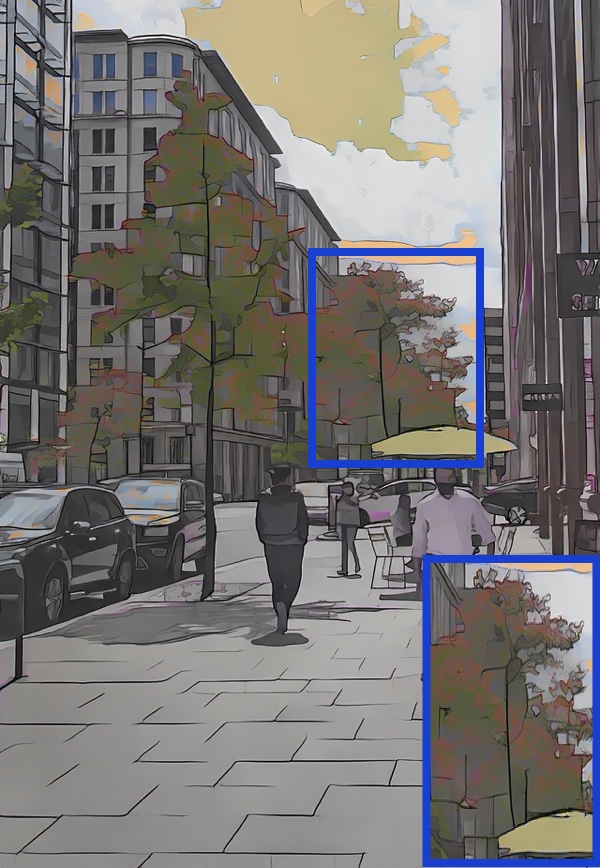}\hspace{\himg}
\includegraphics[width=\h, height=\hh]{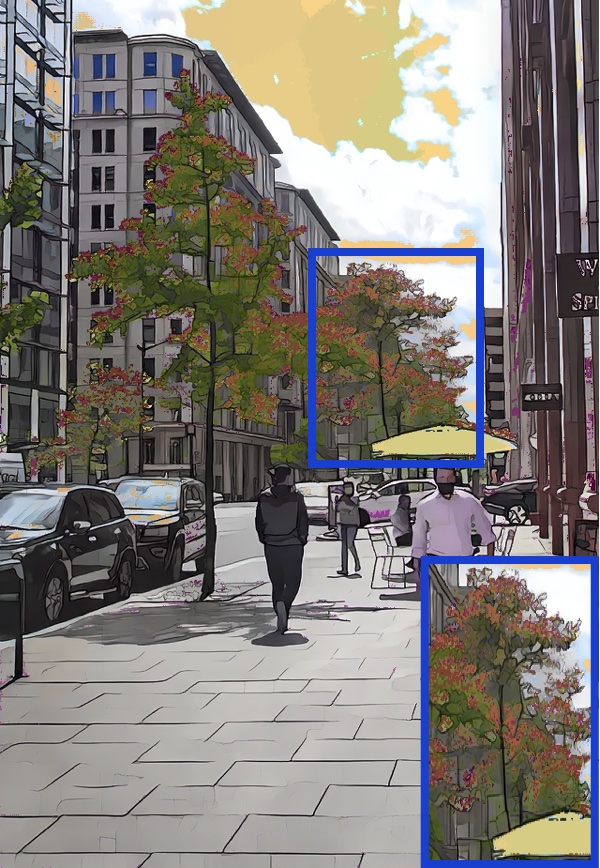}\hspace{\himg}
\includegraphics[width=\h, height=\hh]{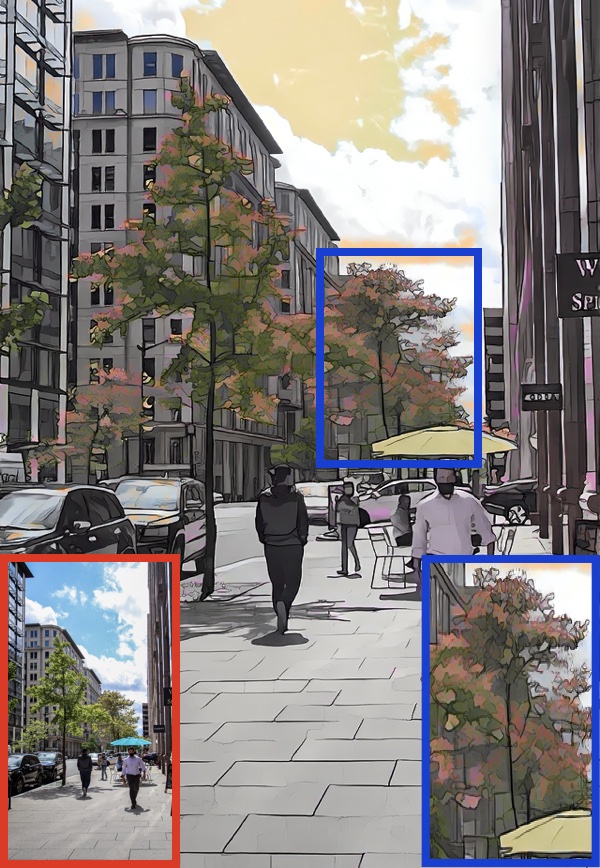}\hfill
\\
\makebox[\h][c]{(a) Coloring $\rightarrow$}\hspace{\himg}
\makebox[\h][c]{(b) $\rightarrow$ Coloring}\hspace{\himg}
\makebox[\h][c]{(c) \textsc{Cartooner}}\hfill
\caption{\textbf{Comparison to naive coloring approach.} We alter the color of sky to yellow and tree to orange maple. (\textbf{a}) A pipeline of re-colorization (to an input image) $\rightarrow$ cartoonization. (\textbf{b}) A pipeline of cartoonization $\rightarrow$ re-colorization. (\textbf{c}) Our method.}
\label{fig:abl_color}
\end{figure}
}

\newcommand{\figCartoonMaking}{
\begin{figure}[t]
\newcommand{\h}{40mm}
\newcommand{\himg}{2mm}
\centering
\includegraphics[width=\h]{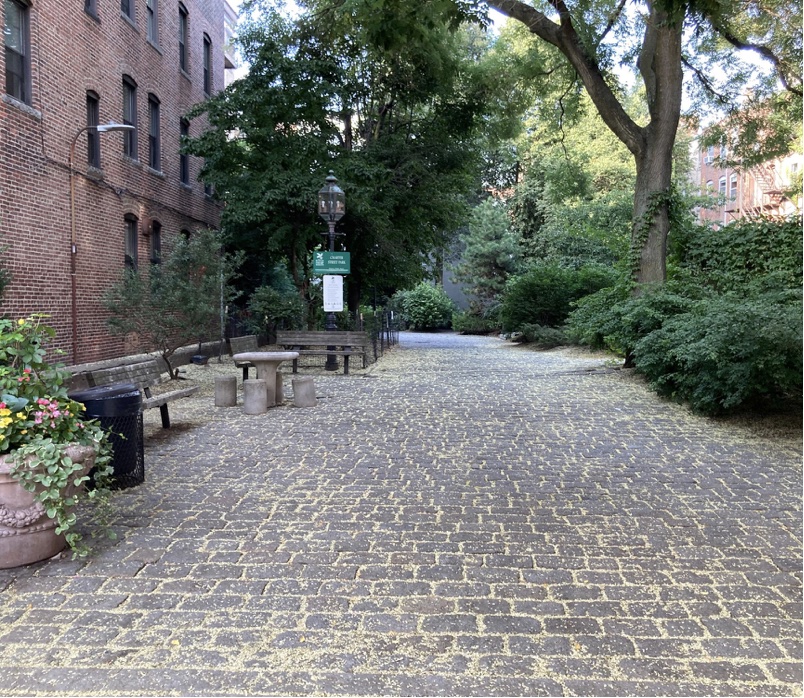}\hspace{\himg}
\includegraphics[width=\h]{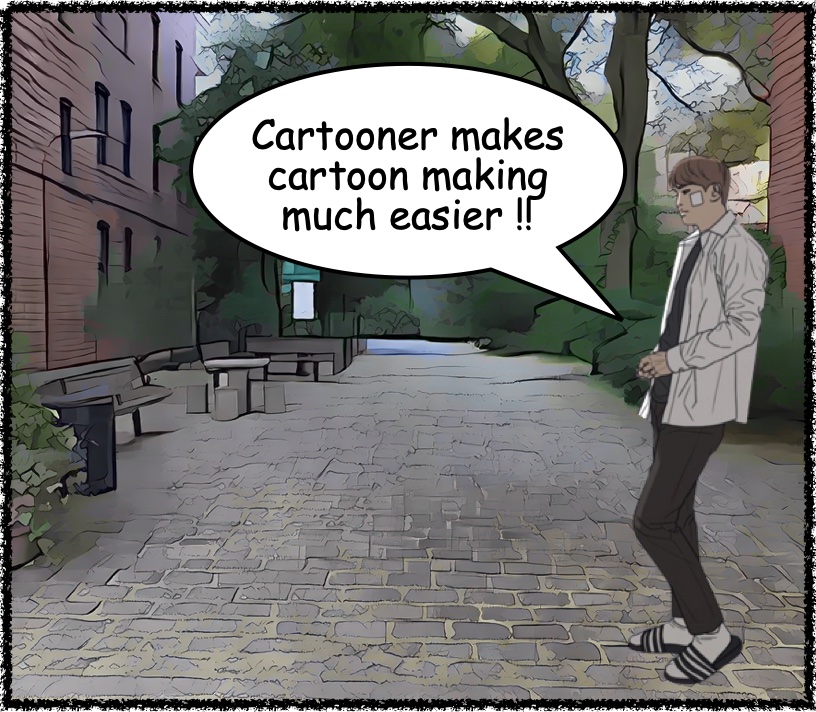}\hfill
\\
\makebox[\h][c]{(a) Source photo}\hspace{\himg}
\makebox[\h][c]{(b) Result}\hfill
\caption{\textbf{\textsc{Cartooner} in cartoon making.} The artists can employ \textsc{Cartooner} to their workflow to improve productivity.}
\label{fig:cartoon_making}
\end{figure}
}

\newcommand{\figInteractivity}{
\begin{figure*}[t]
\newcommand{\h}{43mm}
\newcommand{\himg}{0.0mm}
\centering
\makebox[\h][c]{Source photo}\hspace{\himg}
\makebox[\h][c]{Initial result}\hspace{\himg}
\makebox[\h][c]{$\rightarrow$ Change stroke \& abs.}\hspace{\himg}
\makebox[\h][c]{$\rightarrow$ Change color tone}\hspace{\himg}
\\
\includegraphics[width=\h]{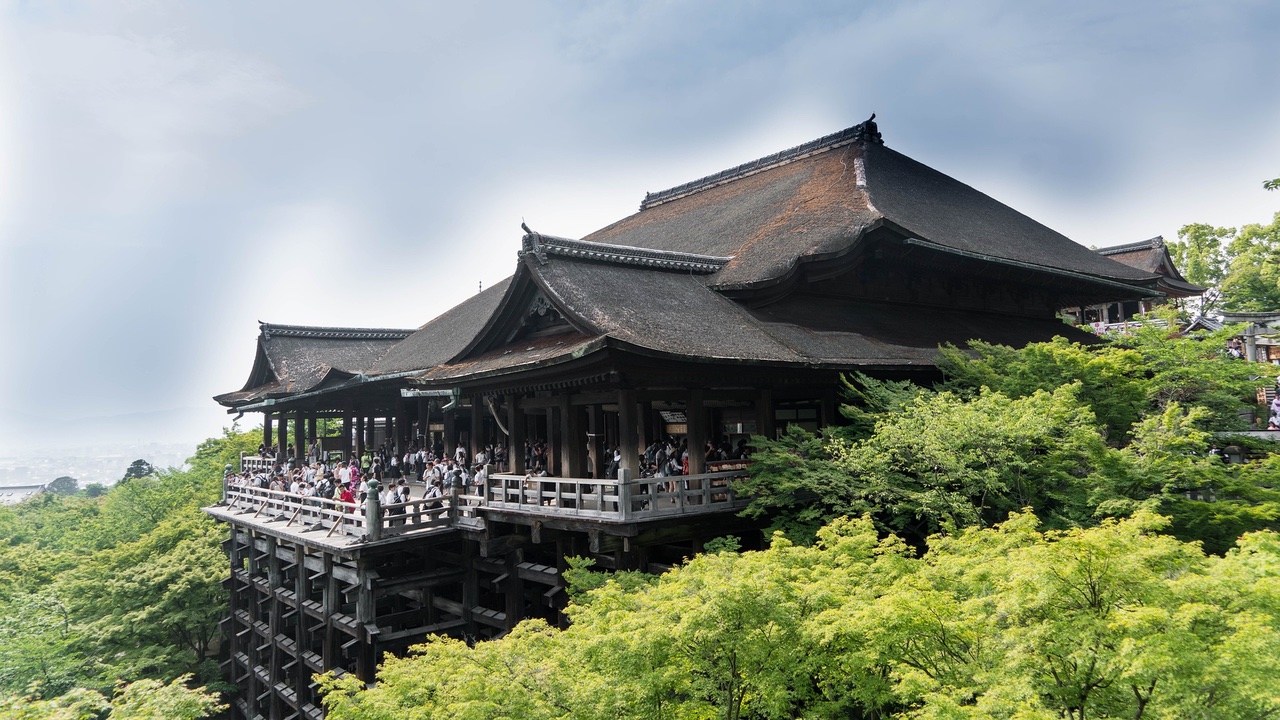}\hspace{\himg}
\includegraphics[width=\h]{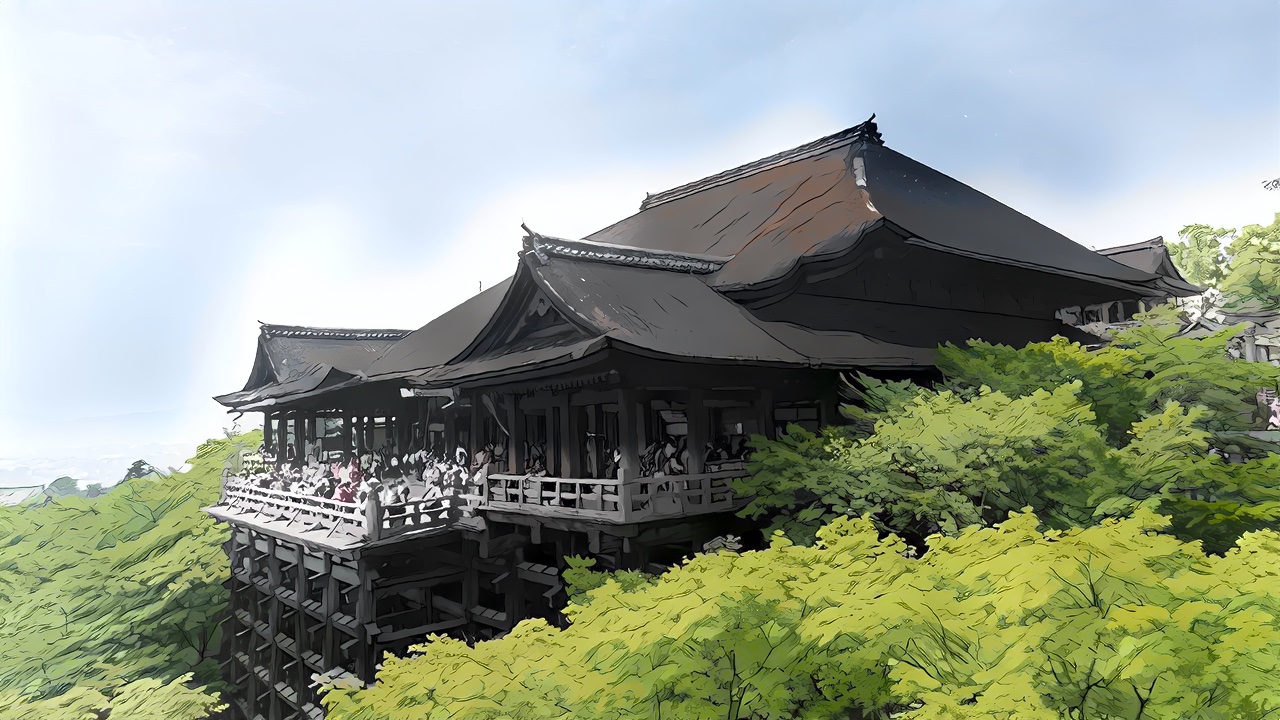}\hspace{\himg}
\includegraphics[width=\h]{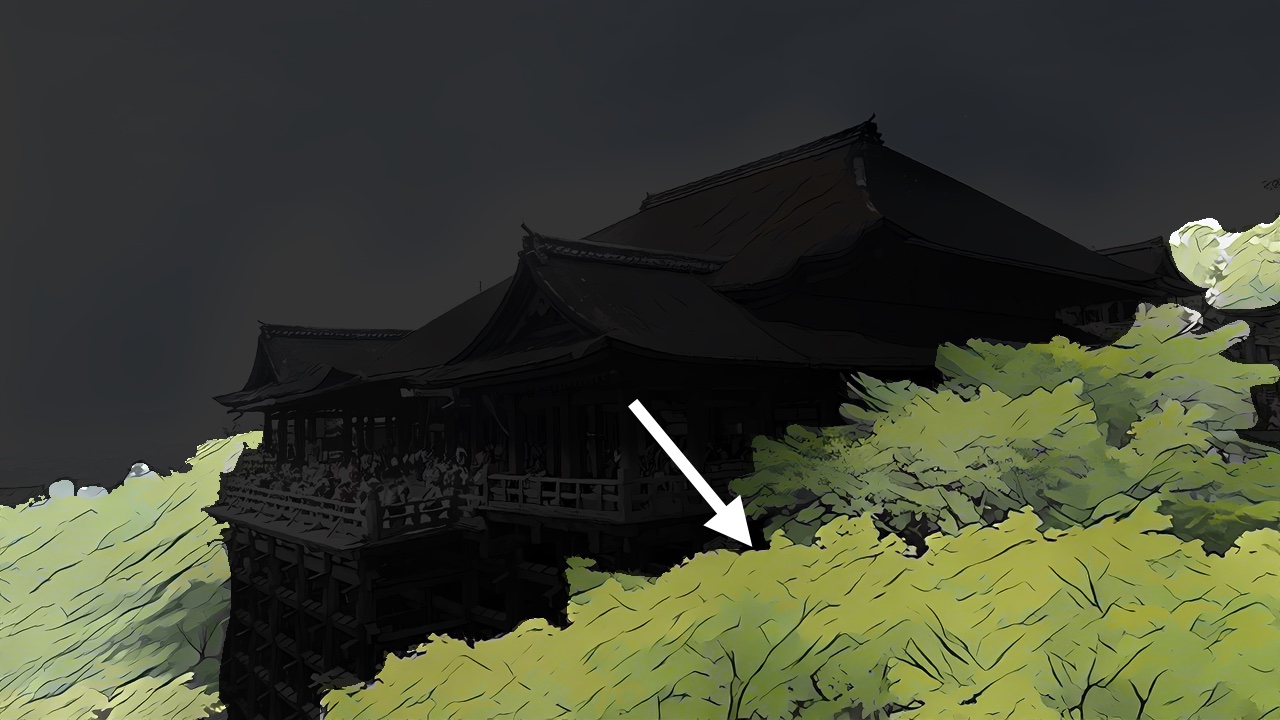}\hspace{\himg}
\includegraphics[width=\h]{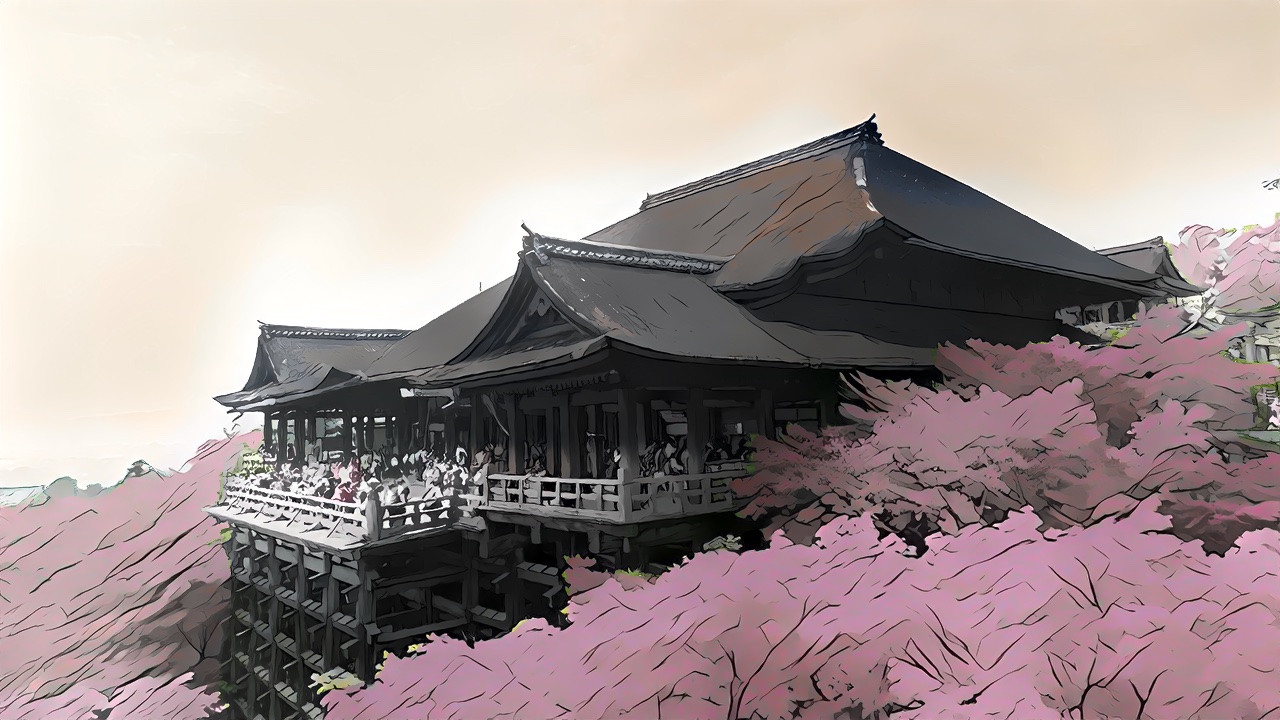}\hfill
\\
\makebox[\h][c]{Source photo}\hspace{\himg}
\makebox[\h][c]{Initial result}\hspace{\himg}
\makebox[\h][c]{$\rightarrow$ Change abstraction}\hspace{\himg}
\makebox[\h][c]{$\rightarrow$ Change color tone}\hspace{\himg}
\\
\includegraphics[width=\h]{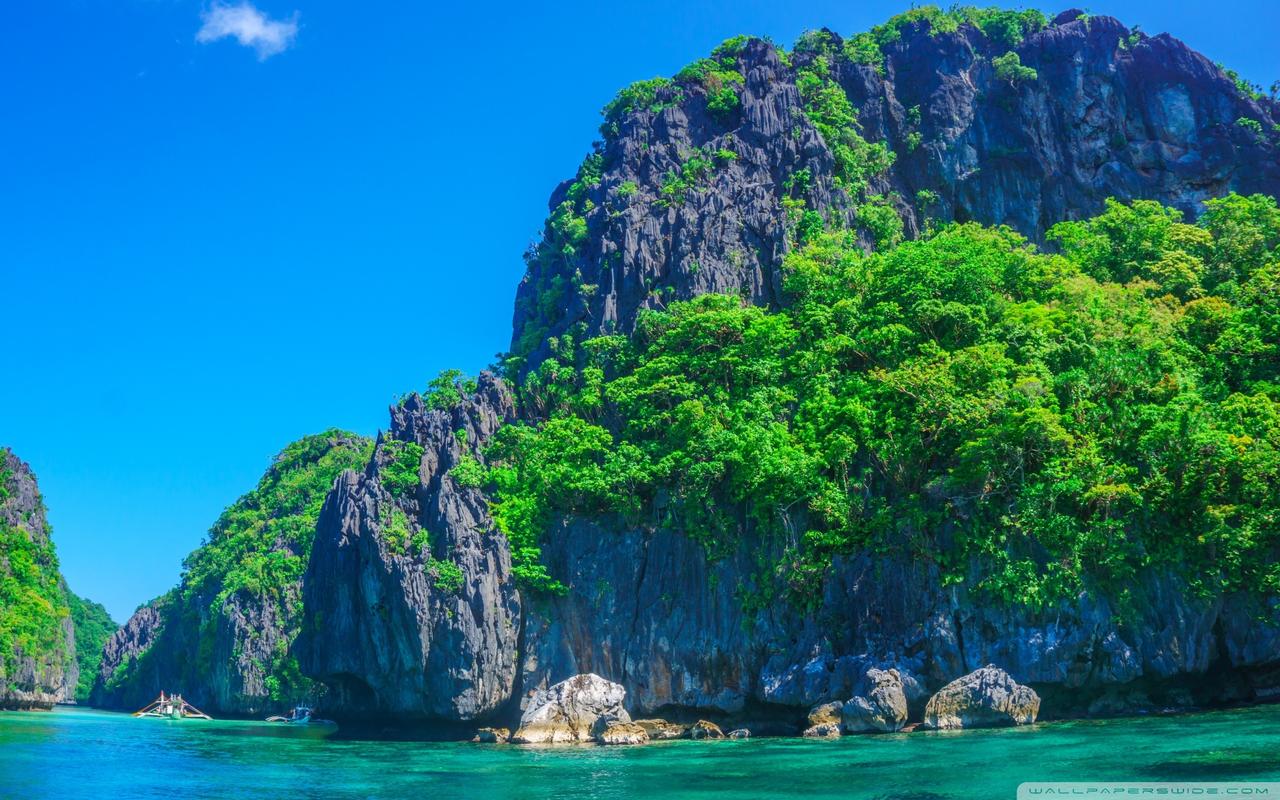}\hspace{\himg}
\includegraphics[width=\h]{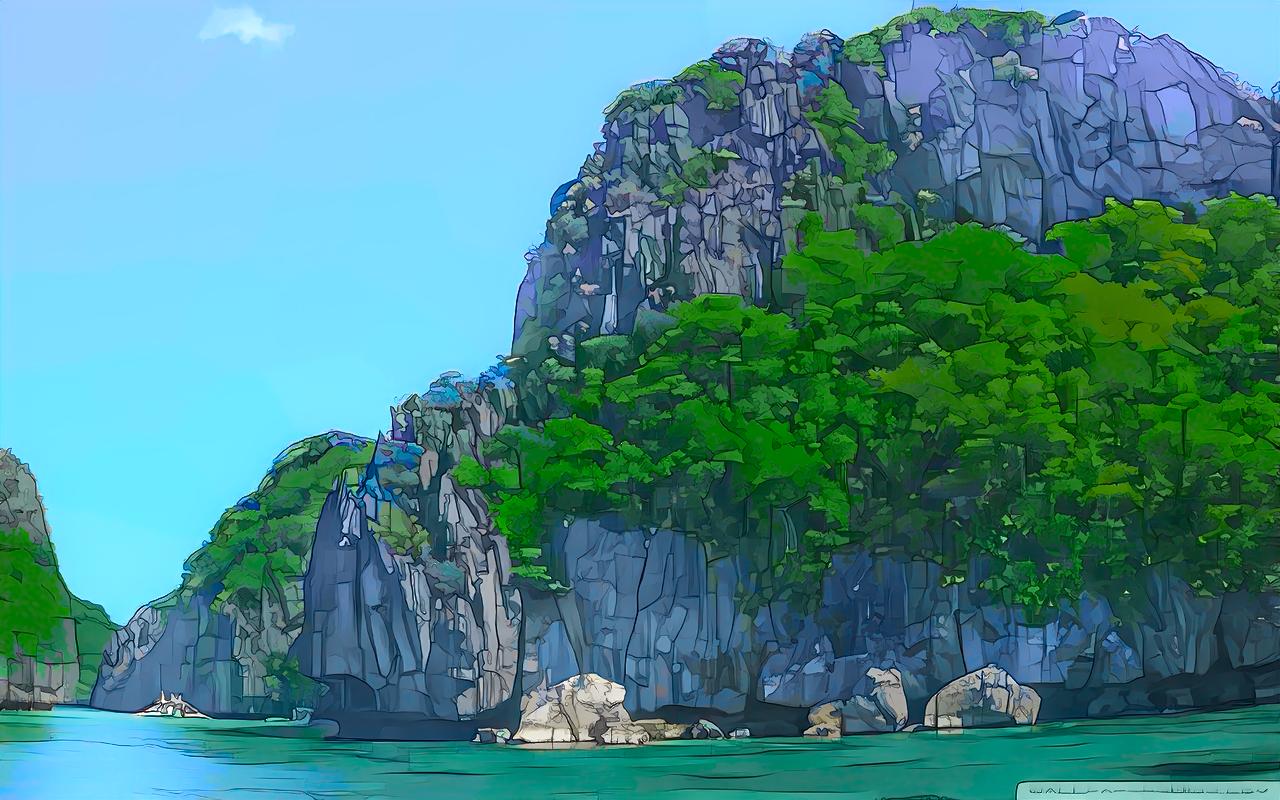}\hspace{\himg}
\includegraphics[width=\h]{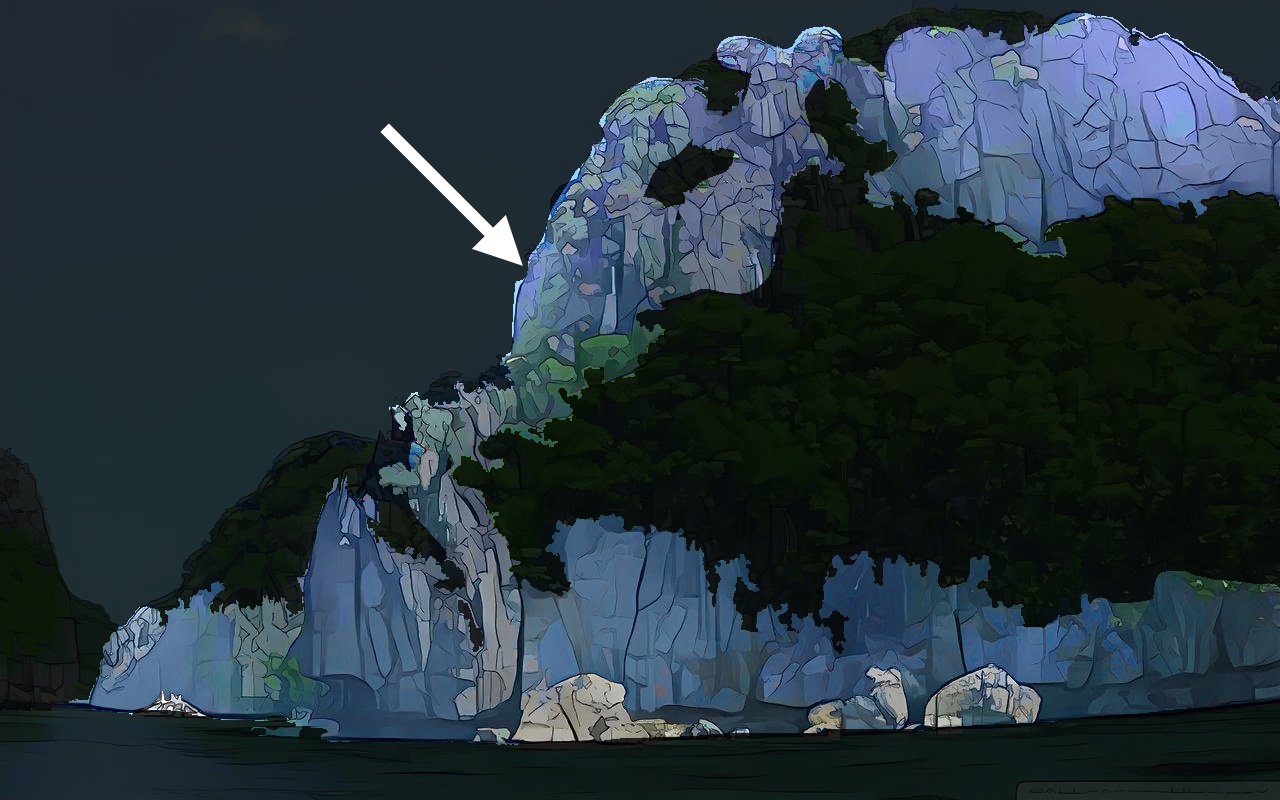}\hspace{\himg}
\includegraphics[width=\h]{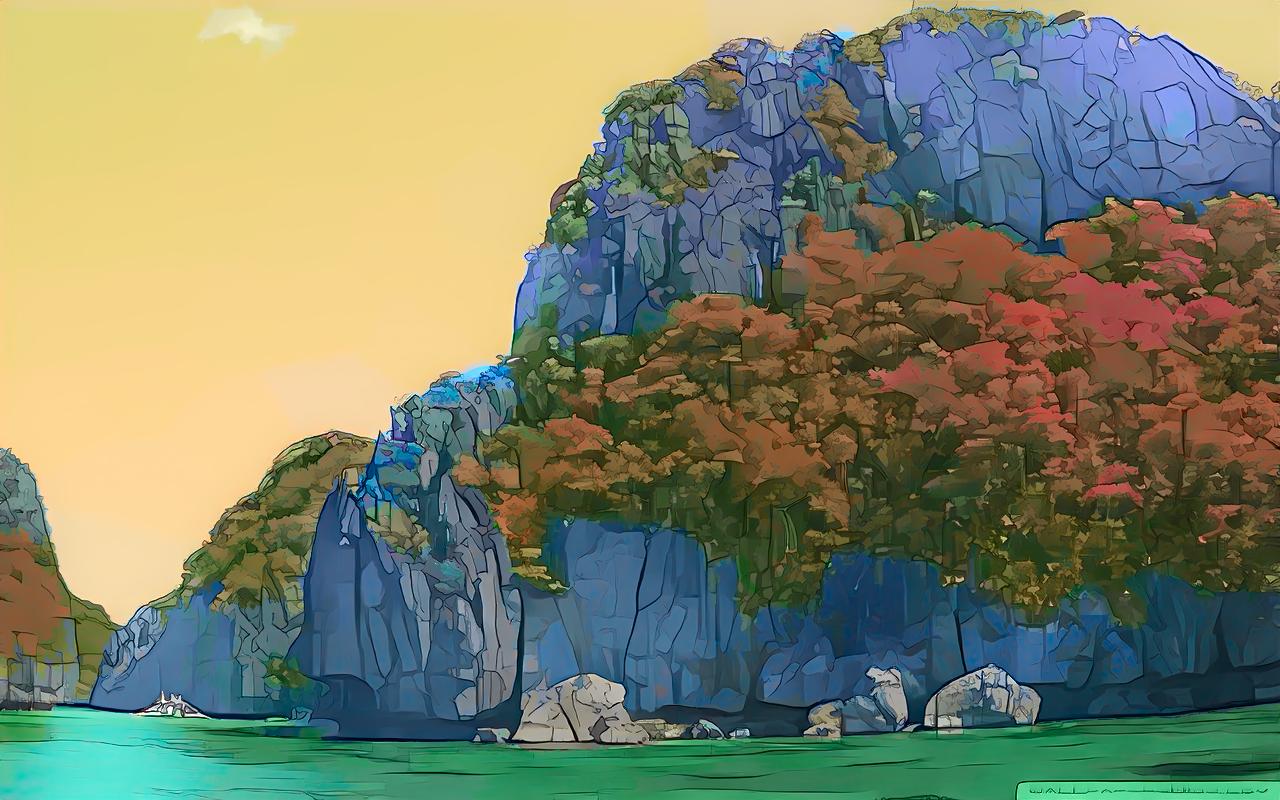}\hfill
\\
\makebox[\h][c]{Source photo}\hspace{\himg}
\makebox[\h][c]{Initial result}\hspace{\himg}
\makebox[\h][c]{$\rightarrow$ Change style}\hspace{\himg}
\makebox[\h][c]{$\rightarrow$ Change color tone}\hspace{\himg}
\\
\includegraphics[width=\h]{figures/interactivity/c1.jpg}\hspace{\himg}
\includegraphics[width=\h]{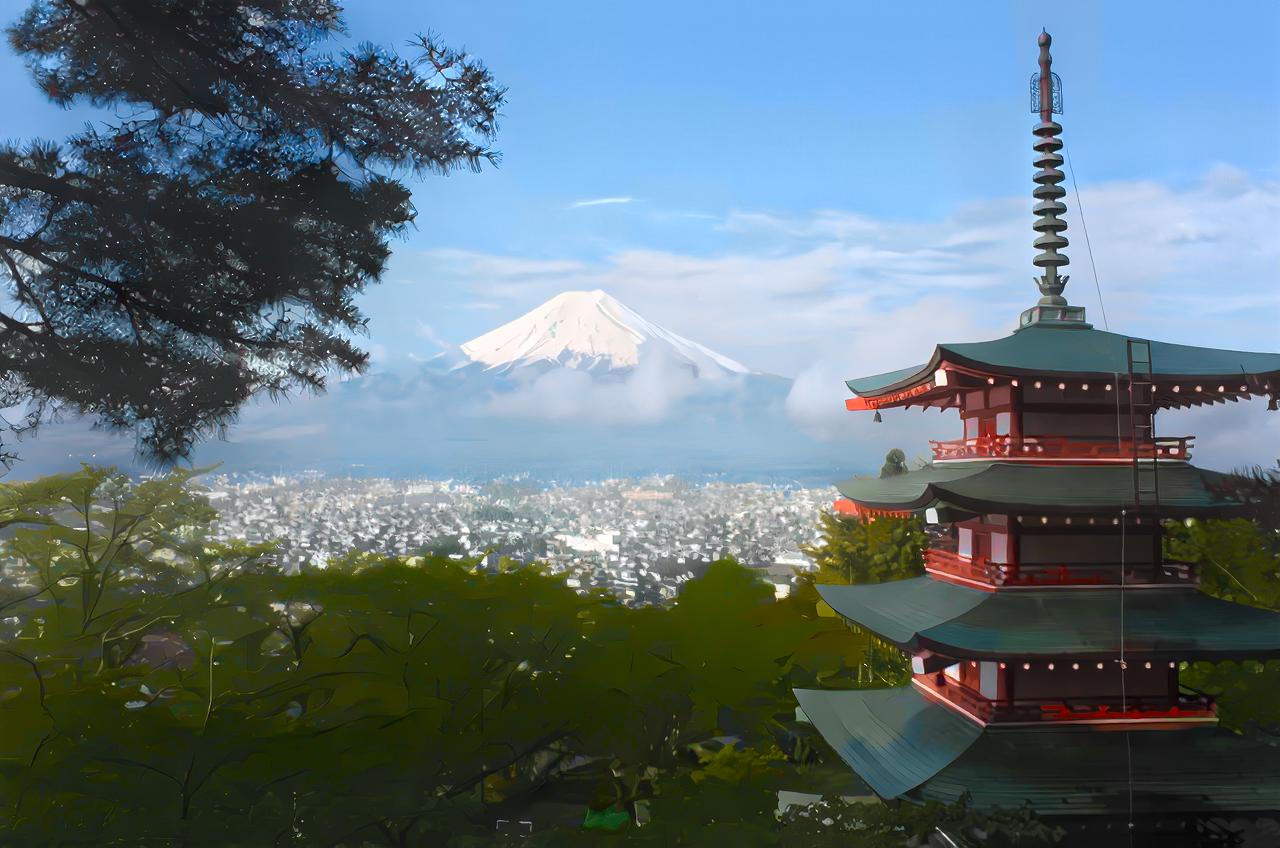}\hspace{\himg}
\includegraphics[width=\h]{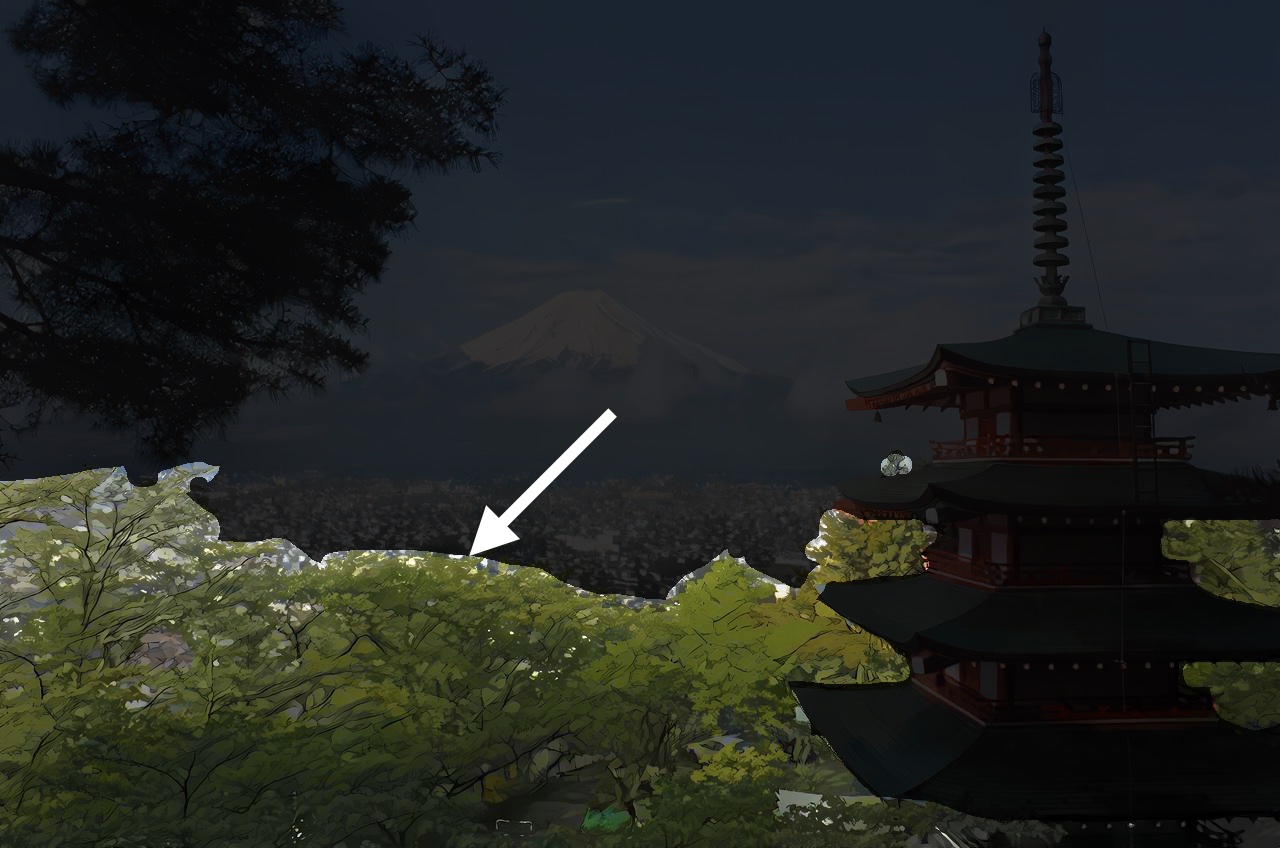}\hspace{\himg}
\includegraphics[width=\h]{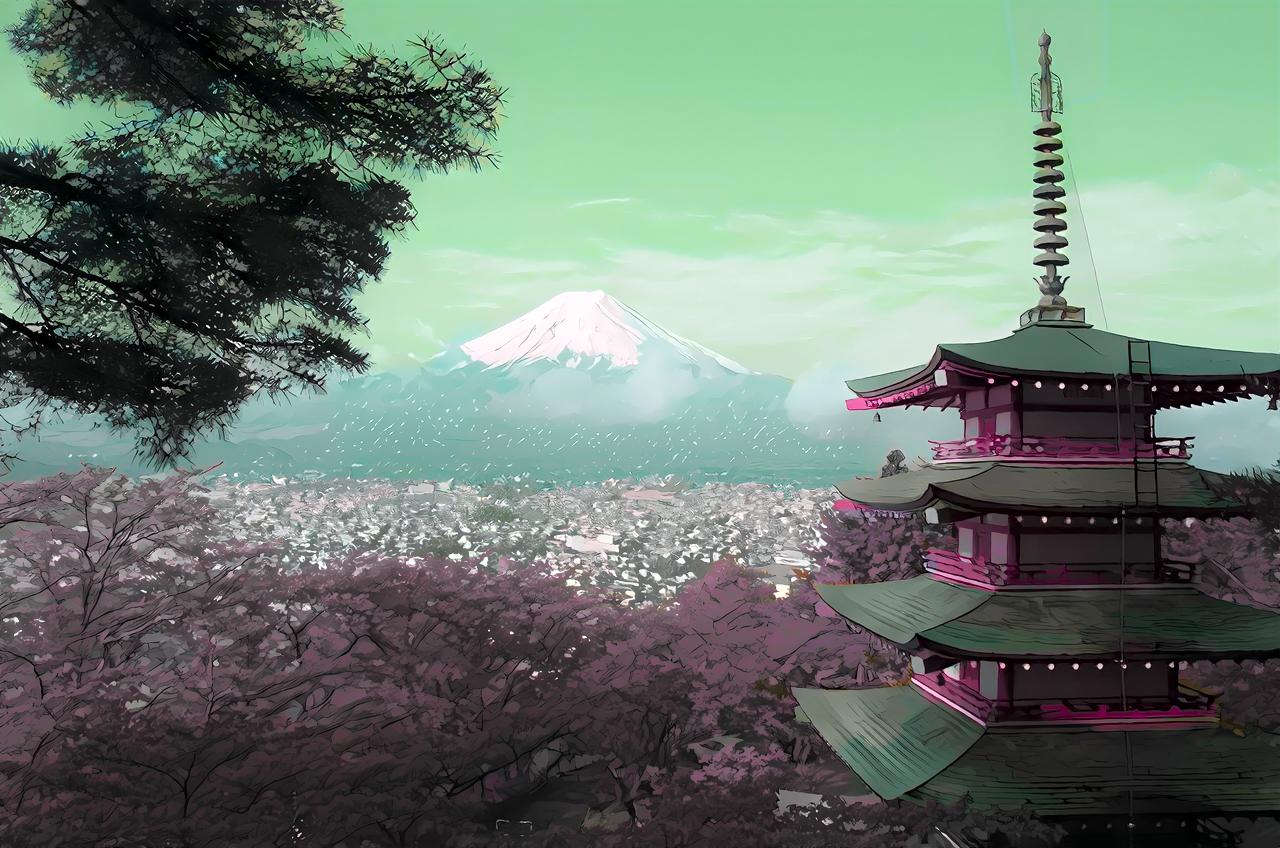}\hfill
\caption{\textbf{Results of interactive cartoonization.} In 3rd column, we highlight the altered region for better visualization.}
\vspace{-0.8em}
\label{fig:interactivity}
\end{figure*}
}

\newcommand{\figRefBased}{
\begin{figure}[t]
\centering
\includegraphics[width=\linewidth]{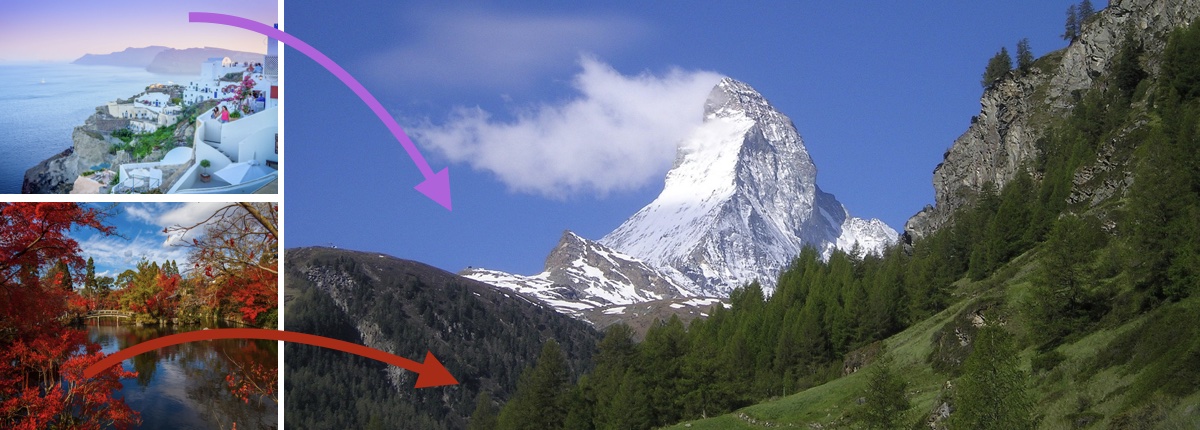}
\includegraphics[width=\linewidth, height=3.4cm]{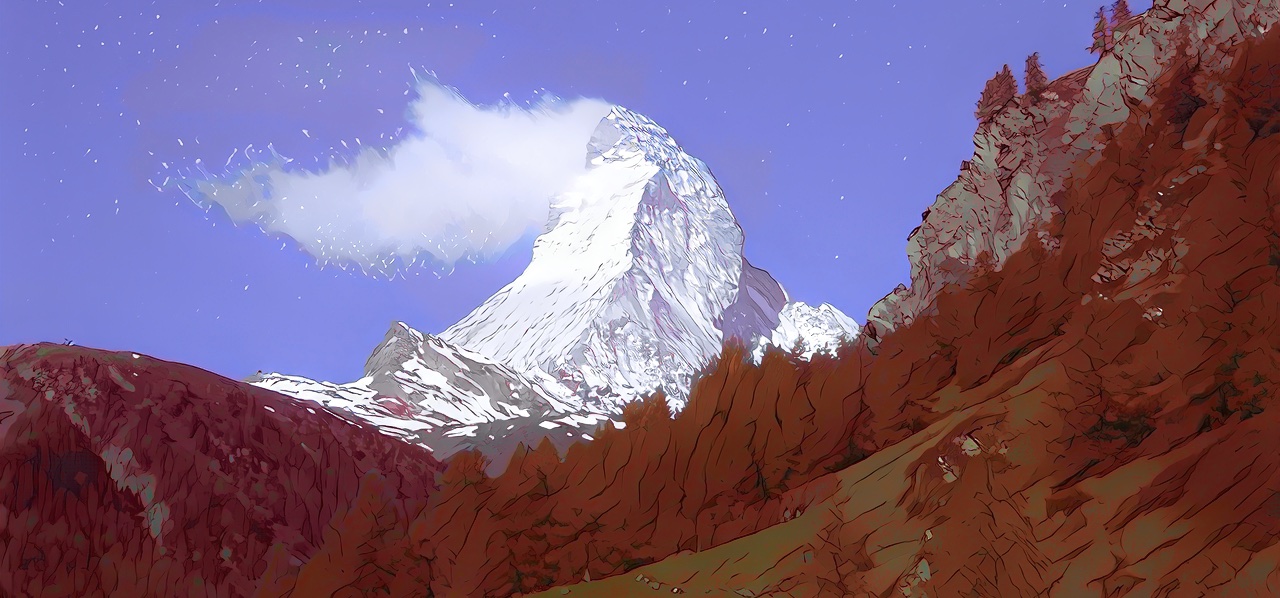}
\caption{\textbf{Reference-based color control.} (\textbf{Top}) Color-reference images and a source photo. (\textbf{Bottom}) Cartoonized output.}
\label{fig:ref_based}
\end{figure}
}

\newcommand{\tableCompQuant}{ 
\begin{table*}[t]
\parbox{.73\linewidth}{
\centering
\setlength\tabcolsep{3.5pt}
\vspace{0.02em}
\caption{\textbf{Quantitative comparison.} We compare with previous deep cartoonization methods using FID~\cite{heusel2017gans} and $\text{FID}_{\text{CLIP}}$~\cite{kynkaanniemi2022role} metrics. Lower score is better.}
\begin{tabular}{l | c c | c c || c c | c c}
\hline
\multirow{2}{*}{Method} & \multicolumn{2}{c|}{Hayao} & \multicolumn{2}{c||}{Shinkai} & \multicolumn{2}{c|}{FreeDraw} & \multicolumn{2}{c}{Barkhan}
\\\cline{2-9}
& $\text{FID}$ & $\text{FID}_{\text{CLIP}}$ & $\text{FID}$ & $\text{FID}_{\text{CLIP}}$ & $\text{FID}$ & $\text{FID}_{\text{CLIP}}$ & $\text{FID}$ & $\text{FID}_{\text{CLIP}}$
\\\hline\hline
CartoonGAN & 172.63 & 59.09 & 167.66 & 56.67 & 162.69 & 49.09 & 108.80 & 28.04
\\
AnimeGANv2 & 156.21 & 57.81 & 146.28 & 54.62 & 132.67 & 45.27 & 82.67 & 23.38
\\
WhiteboxGAN & 153.50 & 63.38 & 150.69 & 58.53 & 120.12 & 41.10 & 93.25 & 27.67
\\
\textsc{Cartooner} & \textbf{142.77} & \textbf{56.76} & \textbf{132.57} & \textbf{51.78} & \textbf{103.88} & \textbf{28.50} & \textbf{74.68} & \textbf{16.44}
\\\hline
\end{tabular}
\label{table:comp_quant}
}
\hfill
\parbox{.24\linewidth}{
\centering
\setlength\tabcolsep{4.5pt}
\caption{\textbf{User study.} Higher quality preference score is better.}
\begin{tabular}{l | c}
\hline
\multirow{2}{*}{Method} & \multirow{2}{*}{Preference} \\
\\\hline\hline
CartoonGAN & \;\;8.9\%
\\
AnimeGANv2 & 17.5\%
\\
WhiteboxGAN & 16.1\%
\\
\textsc{Cartooner} & \textbf{57.5\%}
\\\hline
\end{tabular}
\label{table:user_study}
}
\end{table*}
}

\newcommand{\figSharedUnit}{
\begin{figure}[t]
\newcommand{\hone}{50.5mm}
\newcommand{\htwo}{25.5mm}
\newcommand{\himg}{5.0mm}
\centering
\includegraphics[width=\hone]{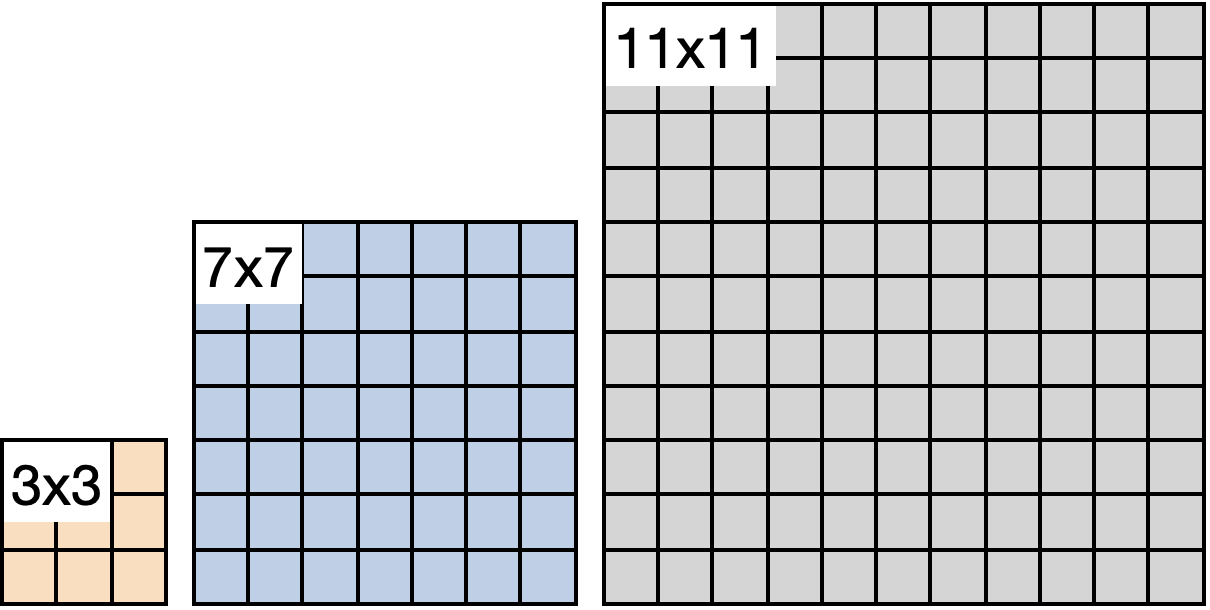}\hspace{\himg}
\includegraphics[width=\htwo]{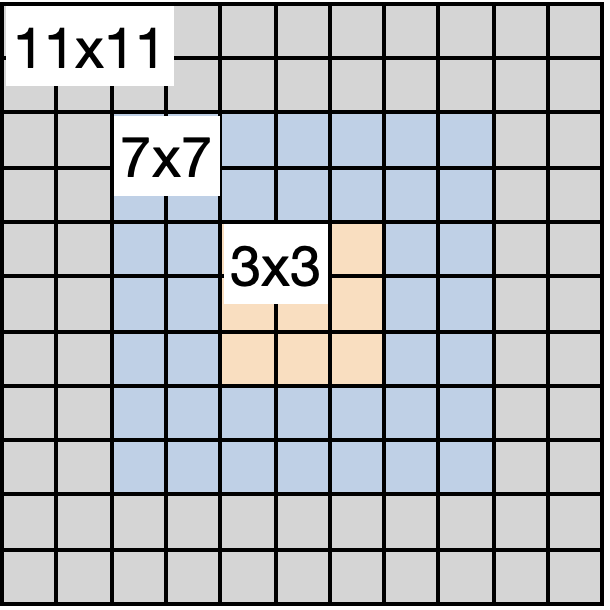}\hfill
\\
\makebox[\hone][c]{(a) Unshared conv}\hspace{\himg}
\makebox[\htwo][c]{(b) Shared conv}\hfill
\caption{\textbf{Schematic overview of the shared conv kernel.} (\textbf{a}) Unshared (vanilla) conv kernel scheme. (\textbf{b}) Our shared conv kernel scheme, which was applied to constitute a multi-branch module in the abstraction control unit. To do that, we employed the conv with the largest kernel size only. All the other conv kernels were set to be a subset of the largest one.}
\label{fig:shared_unit}
\end{figure}
}

\newcommand{\figAblPretrain}{
\begin{figure}[t]
\newcommand{\h}{27mm}
\newcommand{\himg}{0.0mm}
\newcommand{\height}{23.6mm}
\centering
\includegraphics[width=\h, height=\height]{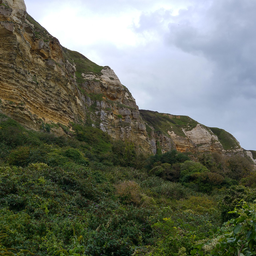}\hspace{\himg}
\includegraphics[width=\h, height=\height]{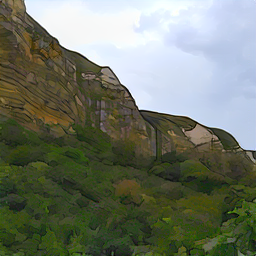}\hspace{\himg}
\includegraphics[width=\h, height=\height]{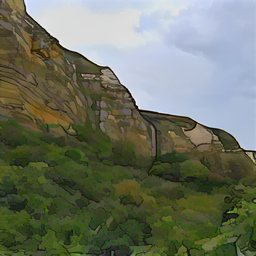}\hfill
\\
\includegraphics[width=\h, height=\height]{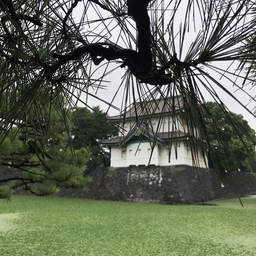}\hspace{\himg}
\includegraphics[width=\h, height=\height]{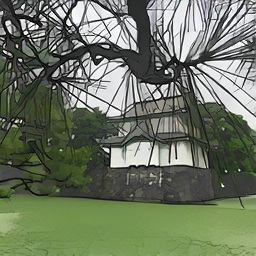}\hspace{\himg}
\includegraphics[width=\h, height=\height]{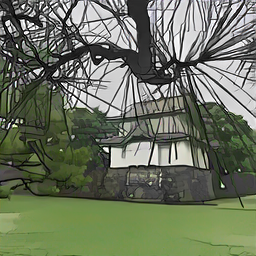}\hfill
\\
\makebox[\h][c]{(a) Photo}\hspace{\himg}
\makebox[\h][c]{(b) w/ Pre-train}\hspace{\himg}
\makebox[\h][c]{(c) w/o Pre-train}\hspace{\himg}
\caption{\textbf{Is pre-training necessary?} Previous deep cartoonization methods~\cite{chen2018cartoongan,chen2019animegan,wang2020learning} rely on pre-training strategy. However, \textsc{Cartooner} does not require tedious pre-training at all; both cases show a very similar performance.}
\label{fig:abl_pretrain}
\end{figure}
}

\newcommand{\figUI}{
\begin{figure*}[t]
\centering
\includegraphics[width=\linewidth]{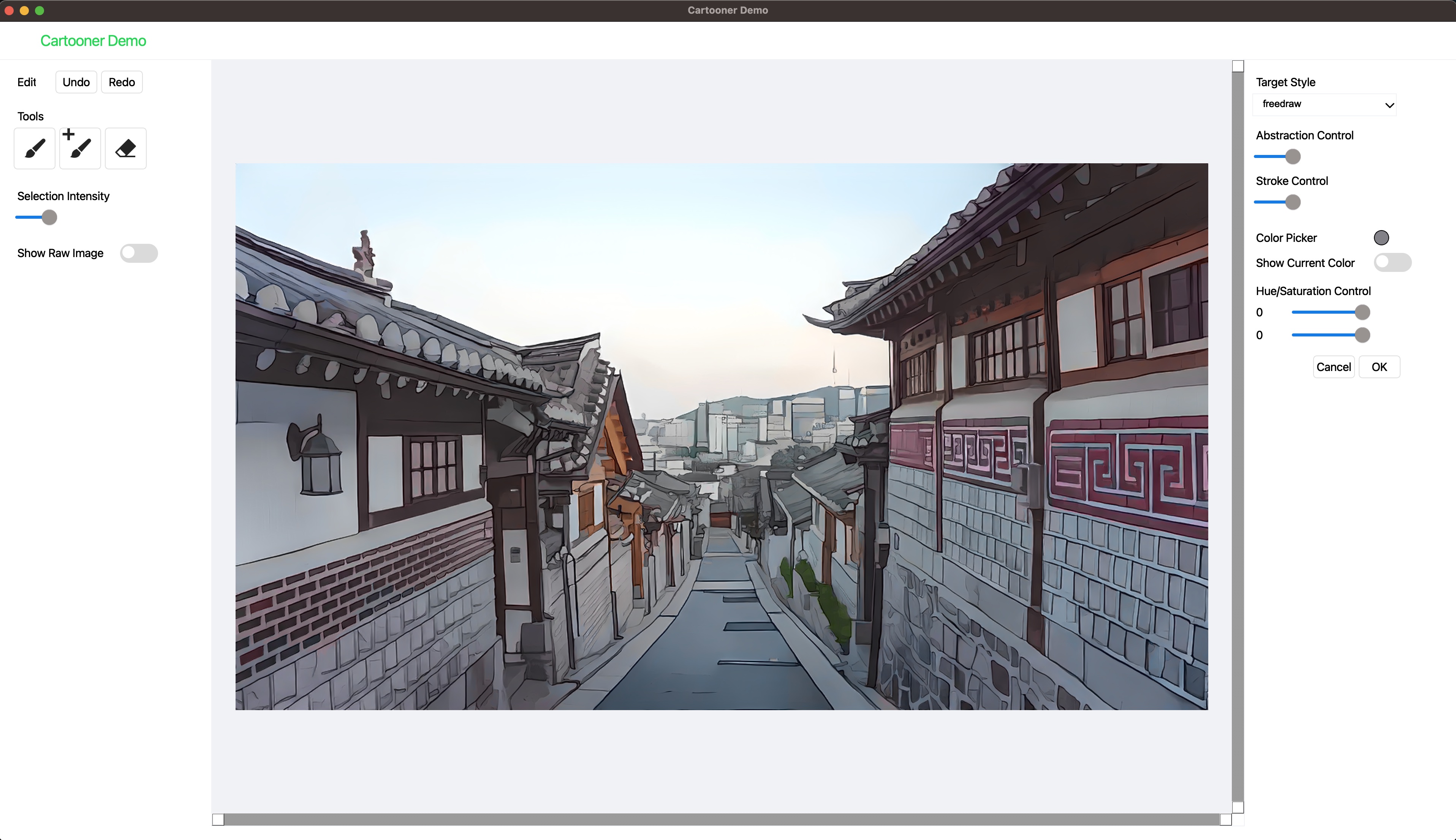}
\caption{\textbf{Interactive UI of \textsc{Cartooner}.}}
\label{fig:ui}
\end{figure*}
}

\newcommand{\figAblAbsShared}{
\begin{figure}[t]
\centering
\includegraphics[width=\linewidth]{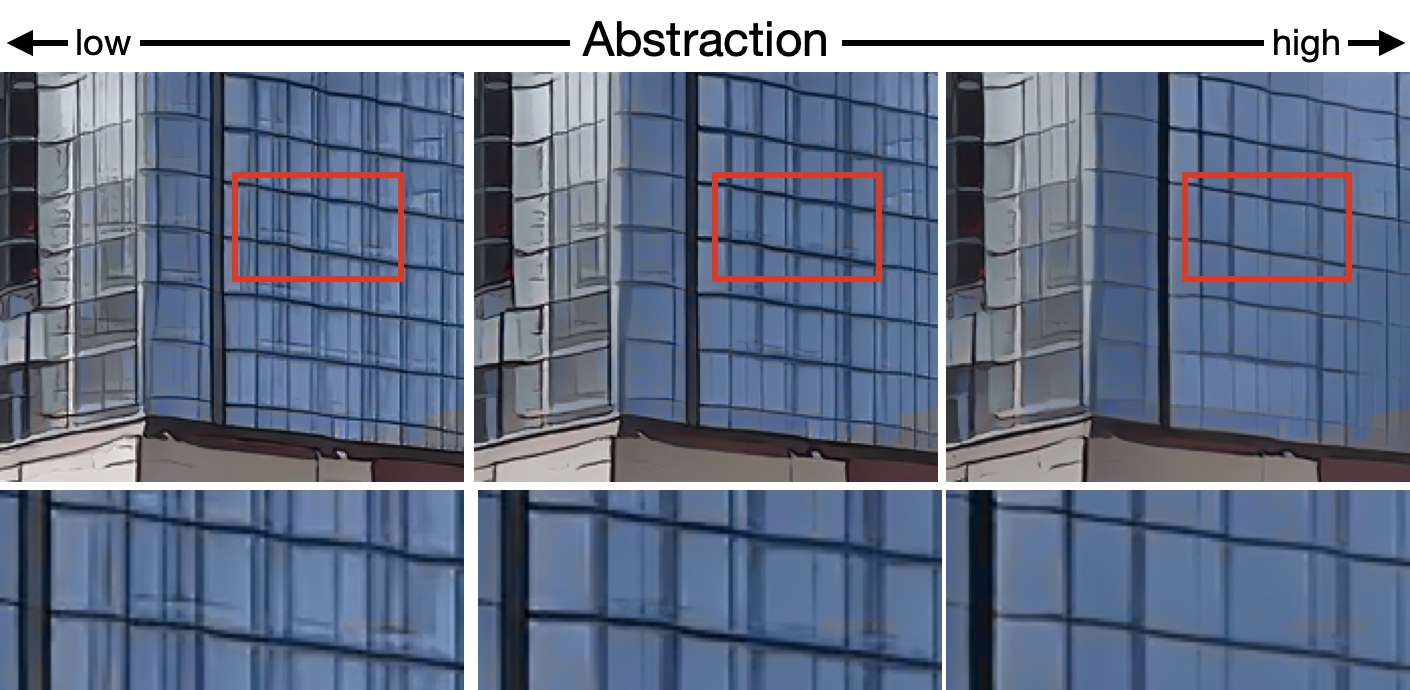}
\\
\makebox[\linewidth][c]{(a) Unshared conv kernel}
\\\vspace{0.45em}
\includegraphics[width=\linewidth]{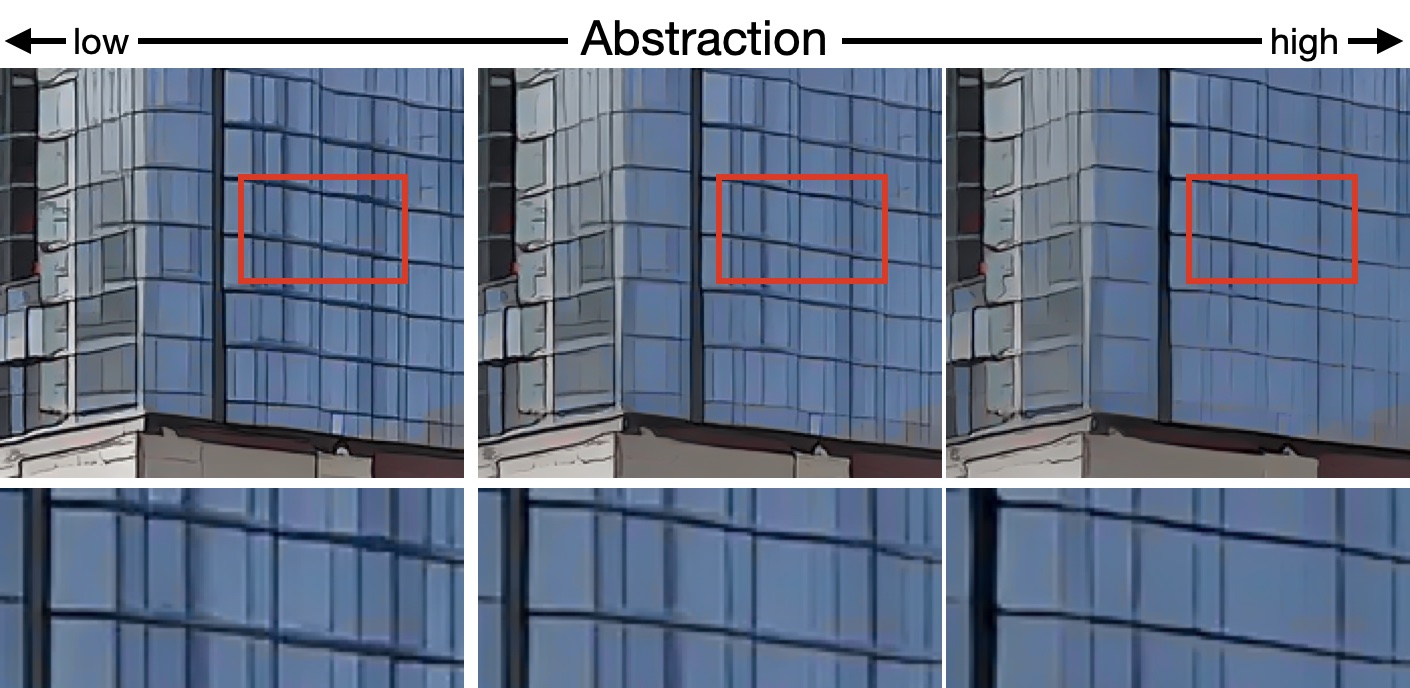}
\\
\makebox[\linewidth][c]{(b) Shared conv kernel}
\\\vspace{0.15em}
\caption{\textbf{Study on the shared kernel in the abstraction control unit.} (\textbf{a}) When the unit does not share the conv kernels, the abstraction levels are not smoothly changed since each conv branch may learn different representations. (\textbf{b}) When the unit shares the conv kernels, the abstraction levels are gradually transited thanks to the shared representations.}
\label{fig:abl_abs_shared}
\end{figure}
}

\newcommand{\figAblAbsKsize}{
\begin{figure}[t]
\centering
\includegraphics[width=\linewidth]{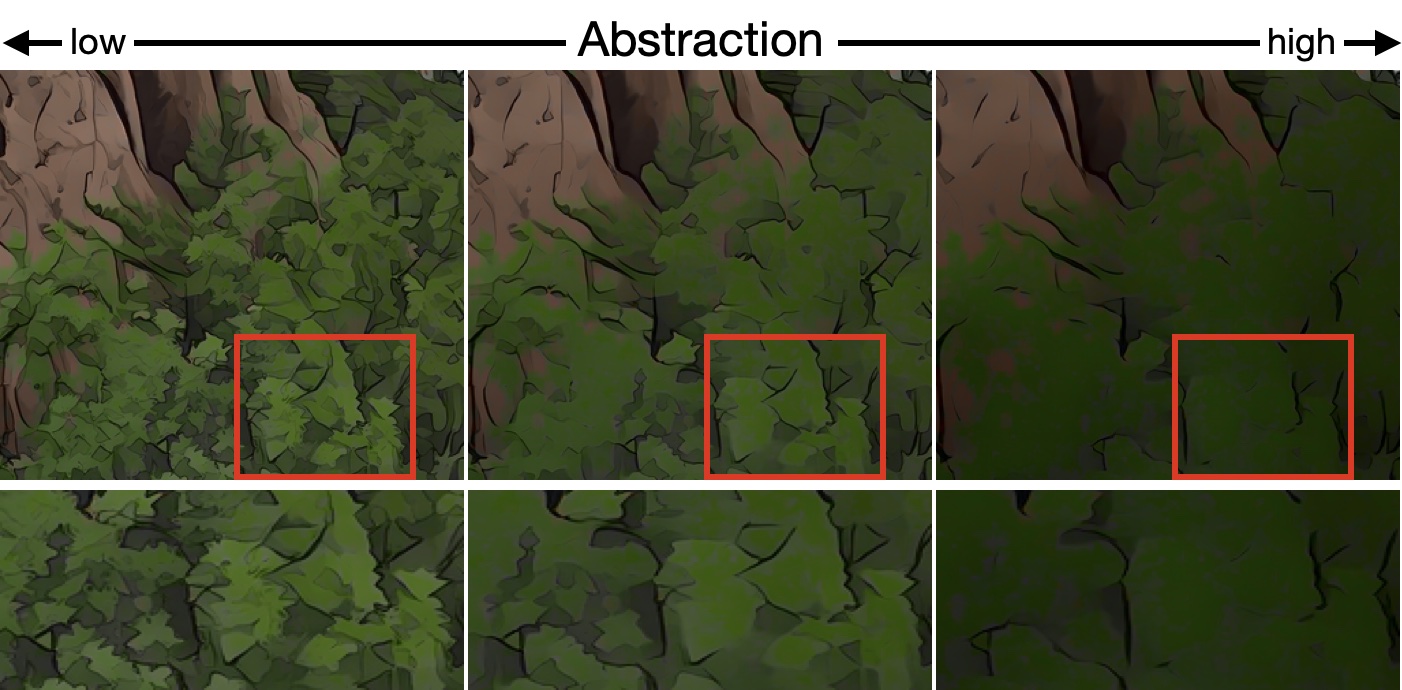}
\\
\makebox[\linewidth][c]{(a) kernel sizes: \{3, 11, 19, 27, 35\}}
\\\vspace{0.3em}
\includegraphics[width=\linewidth]{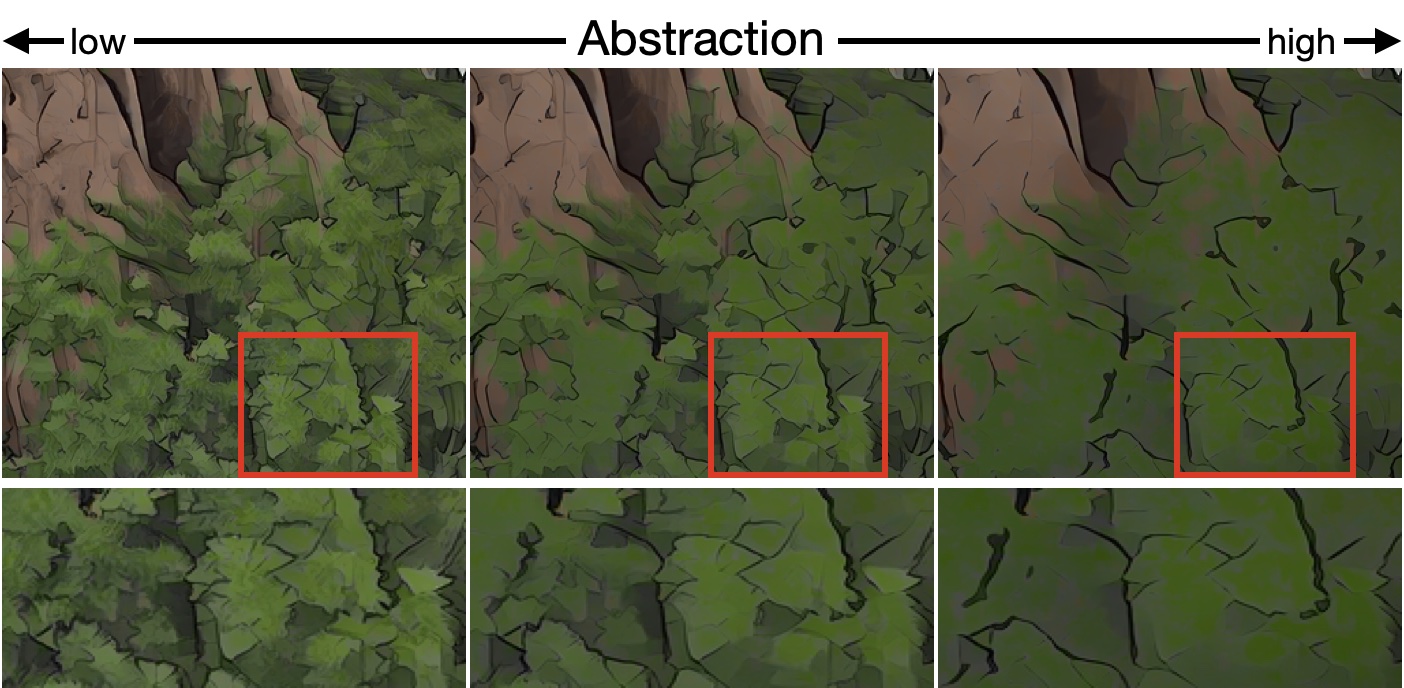}
\\
\makebox[\linewidth][c]{(b) kernel size: \{3, 7, 11, 15, 19\}}
\\
\caption{\textbf{Study on the kernel sizes in the abstraction control unit.} (\textbf{a}) When we extremely increase kernel sizes, the model tends to generate blurred output since the conv kernel refers to too many flat regions. (\textbf{b}) We found that this kernel size setting enables the best abstraction control. Decreasing kernel size also produces good abstraction, but is saturated on the highest level.}
\label{fig:abl_abs_ksize}
\end{figure}
}

\newcommand{\figResultPreAnalysisSuppl}{
\begin{figure}[t]
\centering
\includegraphics[width=\linewidth]{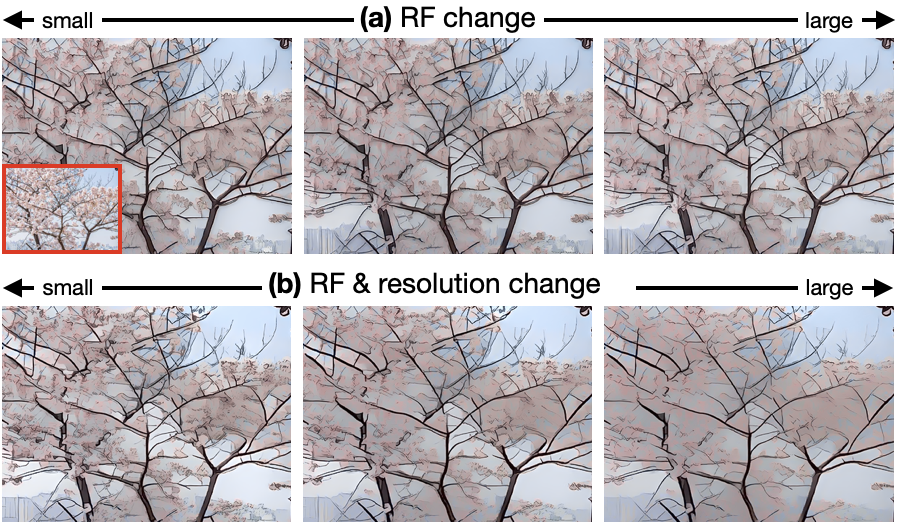}
\caption{\textbf{What affects the abstraction level?} (\textbf{a}) When we only increase receptive field (RF) of the generator, the abstraction is not well changed. (\textbf{b}) When we increase both RF of the generator and the resolution of target cartoon images, the abstraction is adequately changed (this result (b) is also in the main text).}
\label{fig:result_preanalysis_suppl}
\end{figure}
}

\newcommand{\figInteractivitySuppl}{
\begin{figure*}[t]
\newcommand{\h}{43mm}
\newcommand{\himg}{0.0mm}
\centering
\makebox[\h][c]{Source photo}\hspace{\himg}
\makebox[\h][c]{Initial result}\hspace{\himg}
\makebox[\h][c]{$\rightarrow$ Change style}\hspace{\himg}
\makebox[\h][c]{$\rightarrow$ Change color tone}\hspace{\himg}
\\
\includegraphics[width=\h]{figures/interactivity/h1.jpg}\hspace{\himg}
\includegraphics[width=\h]{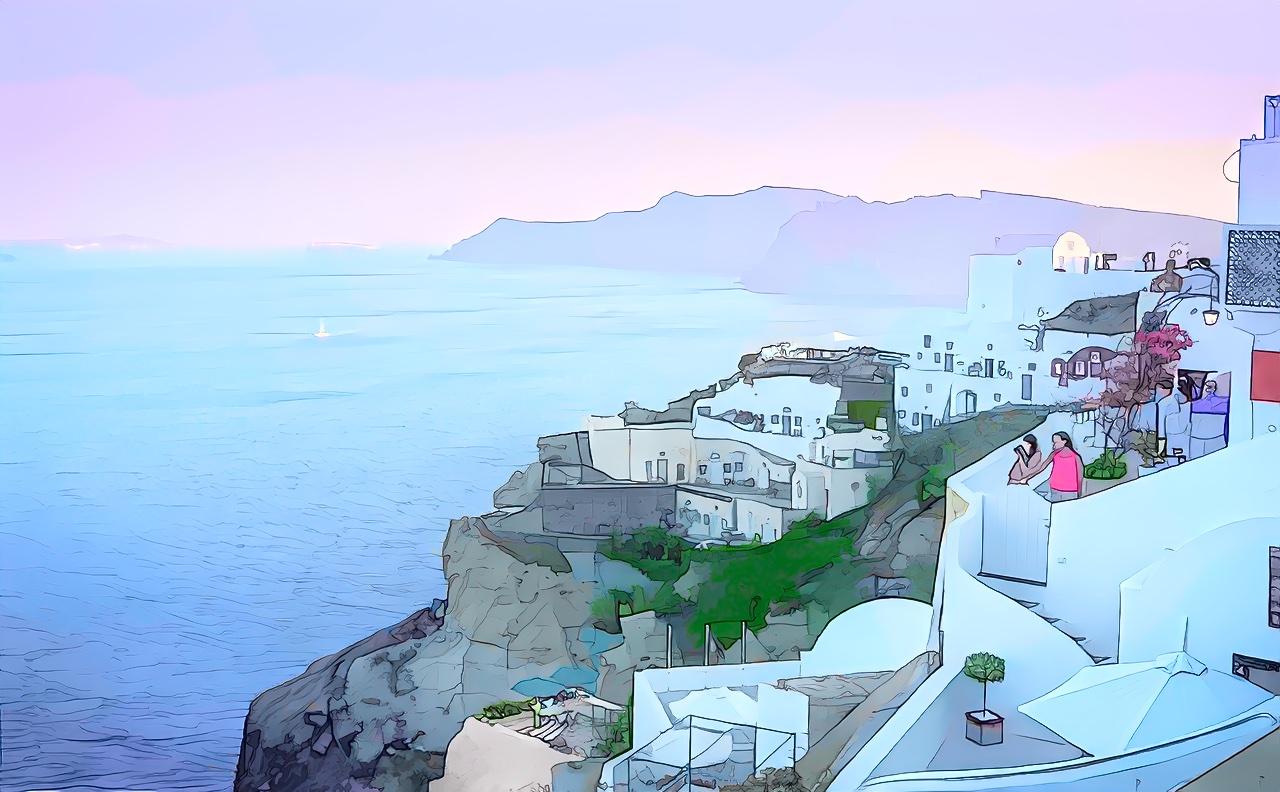}\hspace{\himg}
\includegraphics[width=\h]{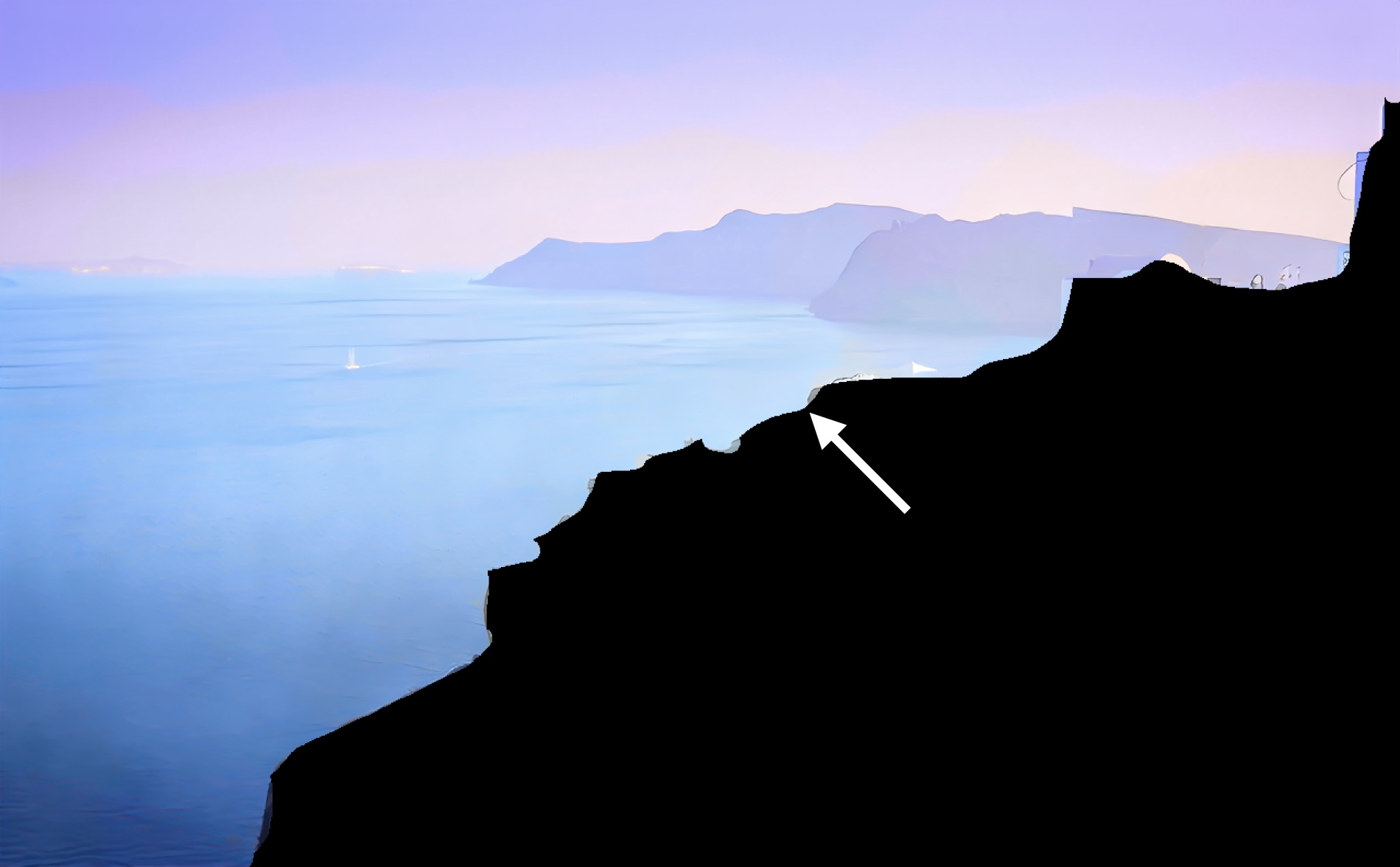}\hspace{\himg}
\includegraphics[width=\h]{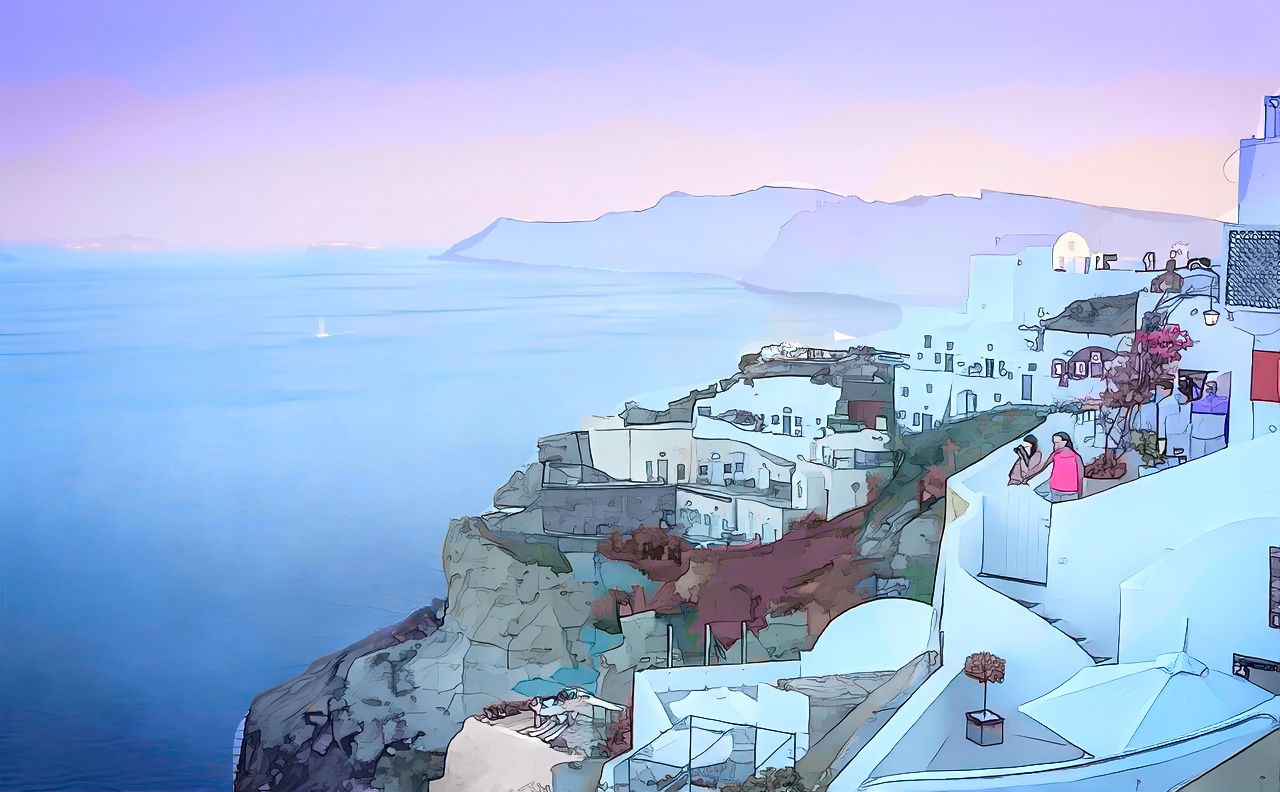}\hfill
\\
\makebox[\h][c]{Source photo}\hspace{\himg}
\makebox[\h][c]{Initial result}\hspace{\himg}
\makebox[\h][c]{$\rightarrow$ Change style \& stroke}\hspace{\himg}
\makebox[\h][c]{$\rightarrow$ Change color tone}\hspace{\himg}
\\
\includegraphics[width=\h]{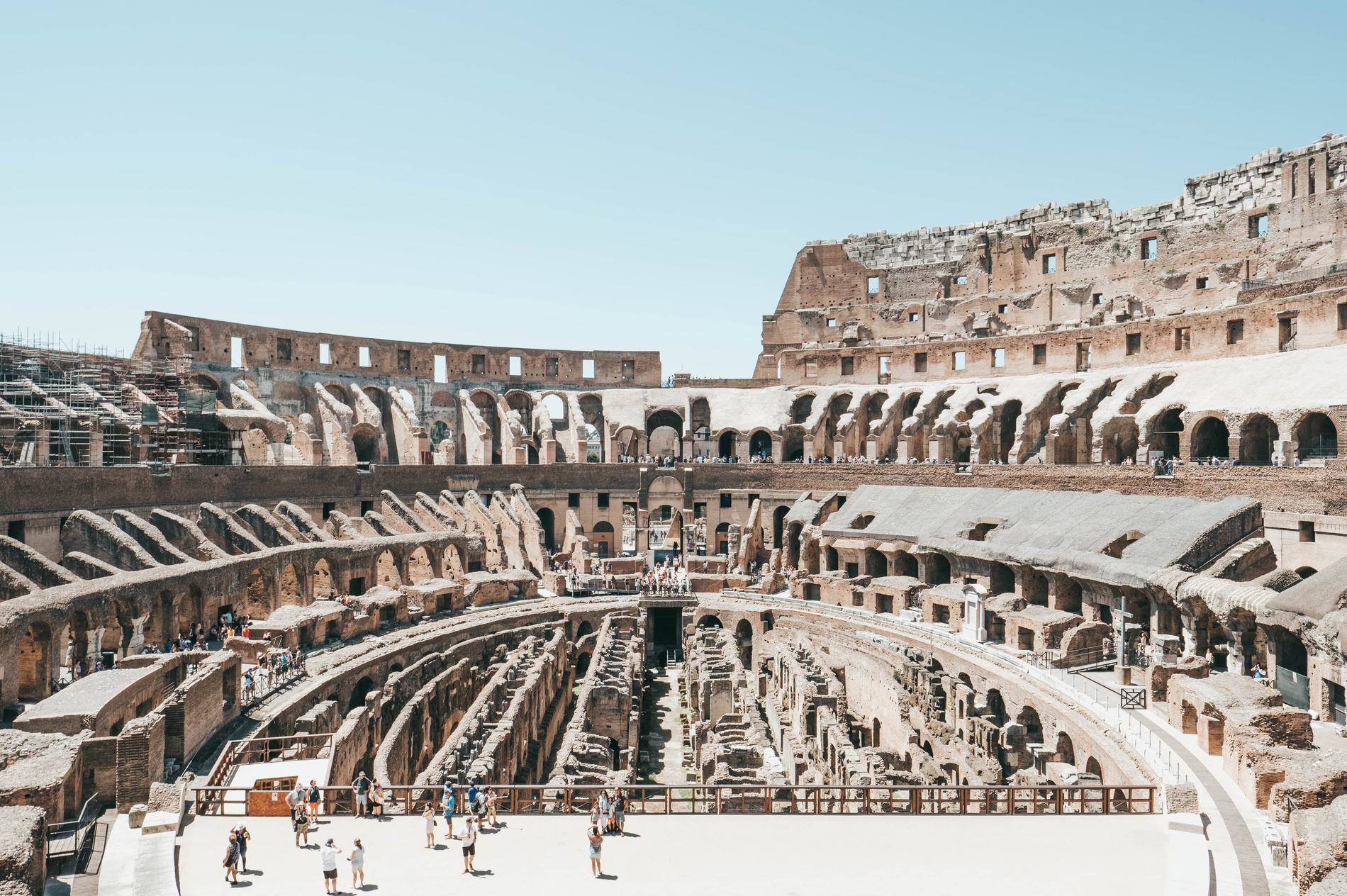}\hspace{\himg}
\includegraphics[width=\h]{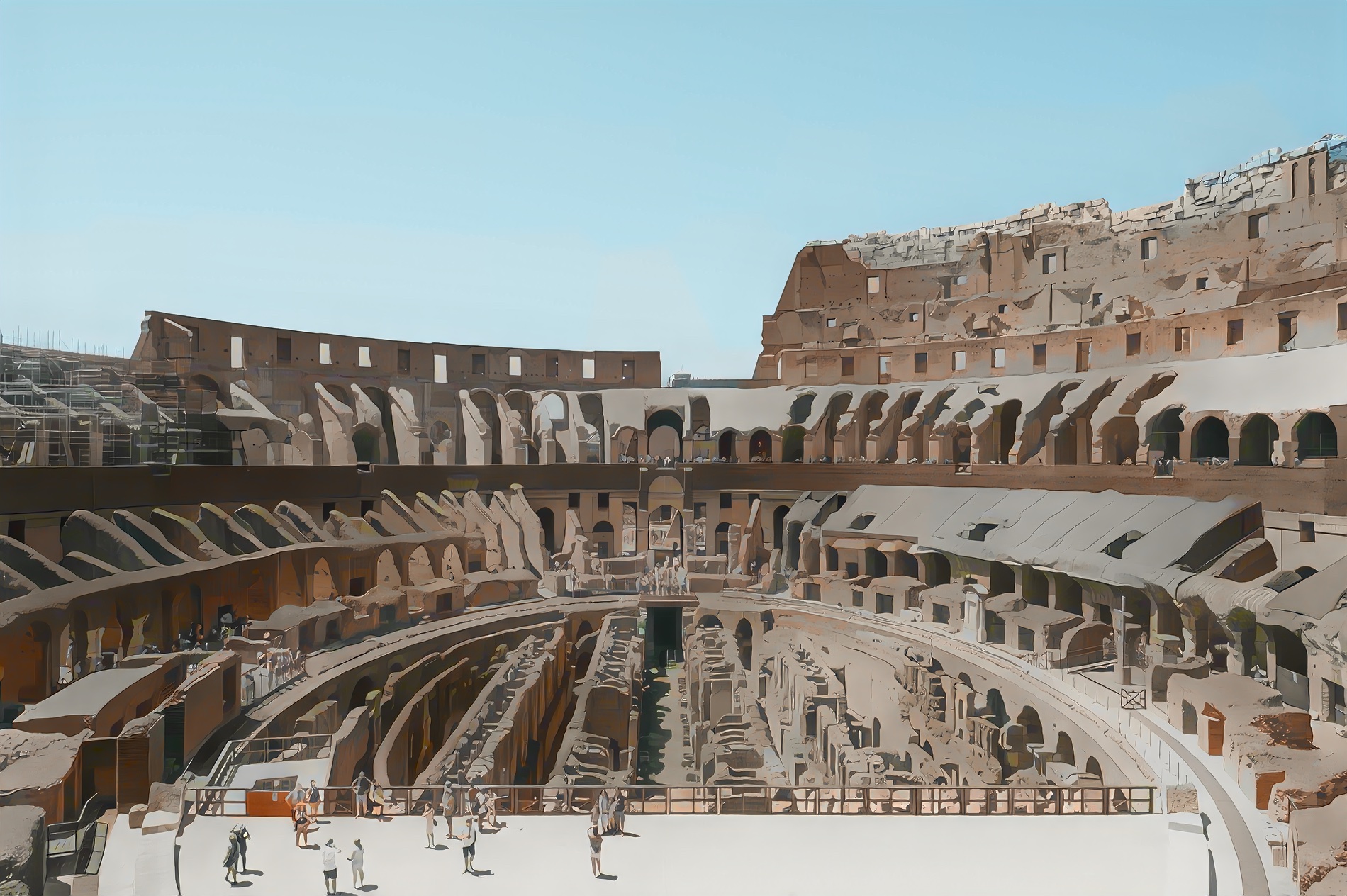}\hspace{\himg}
\includegraphics[width=\h]{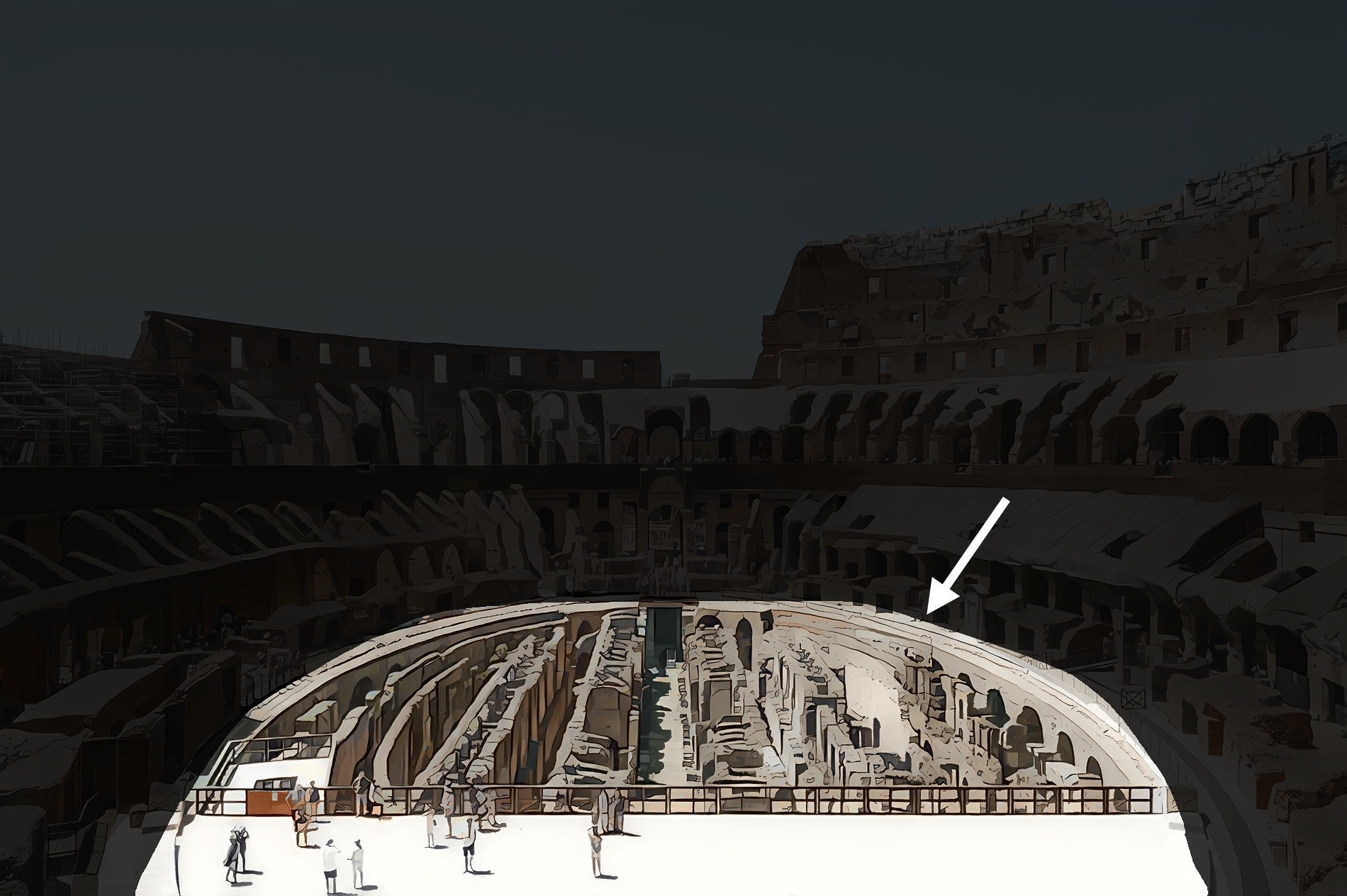}\hspace{\himg}
\includegraphics[width=\h]{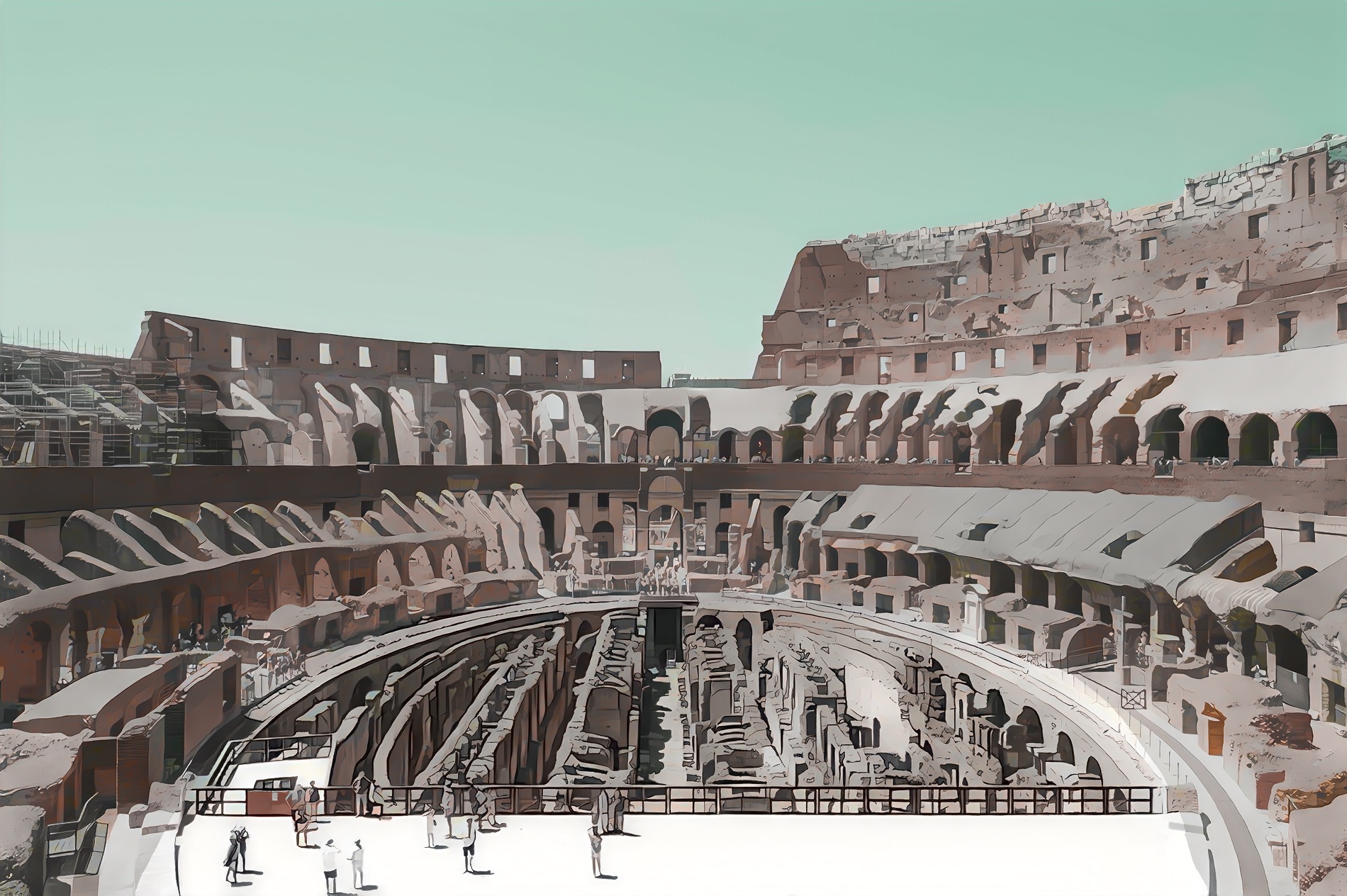}\hfill
\\
\makebox[\h][c]{Source photo}\hspace{\himg}
\makebox[\h][c]{Initial result}\hspace{\himg}
\makebox[\h][c]{$\rightarrow$ Change abstraction}\hspace{\himg}
\makebox[\h][c]{$\rightarrow$ Change color tone}\hspace{\himg}
\\
\includegraphics[width=\h]{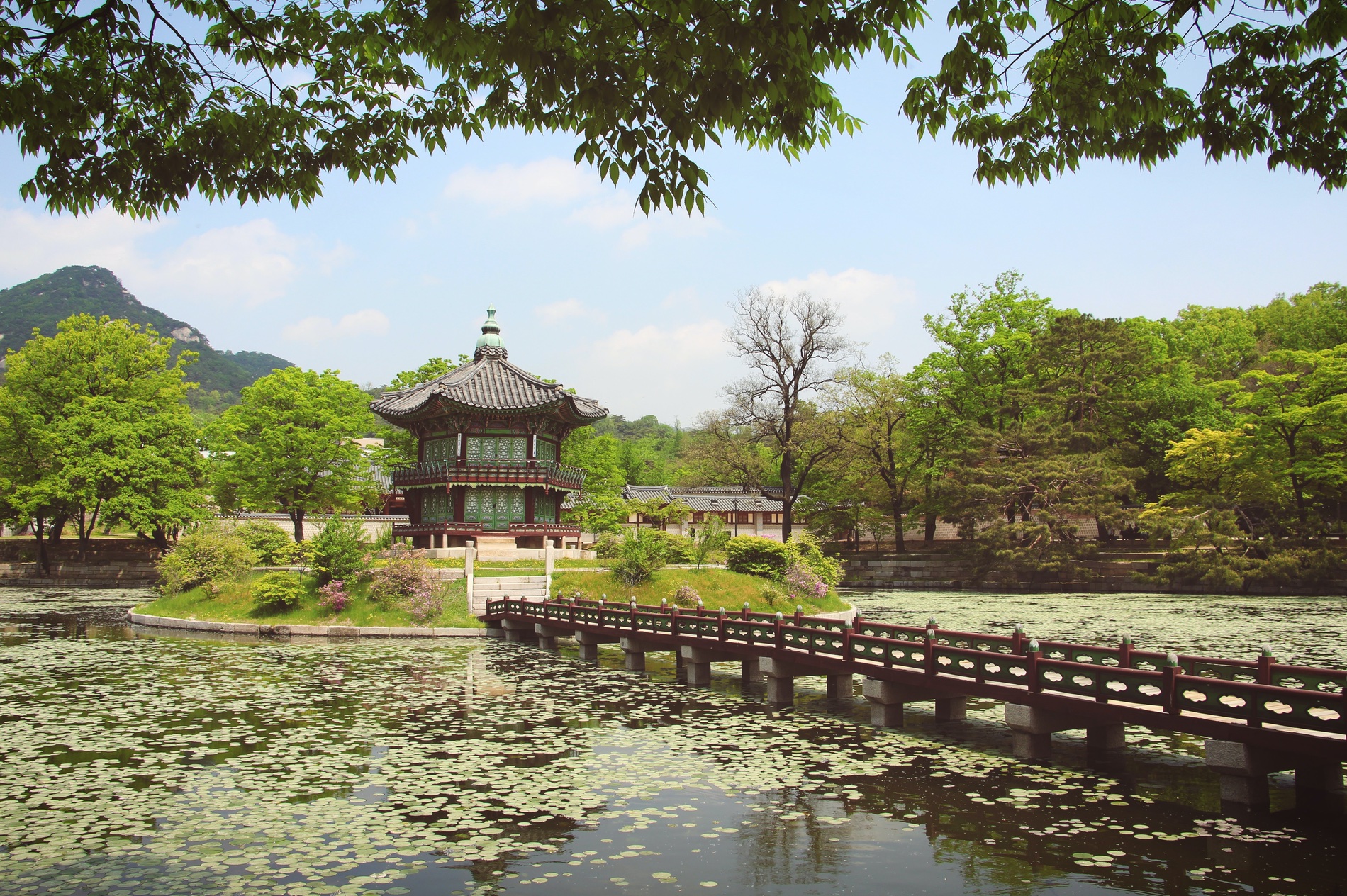}\hspace{\himg}
\includegraphics[width=\h]{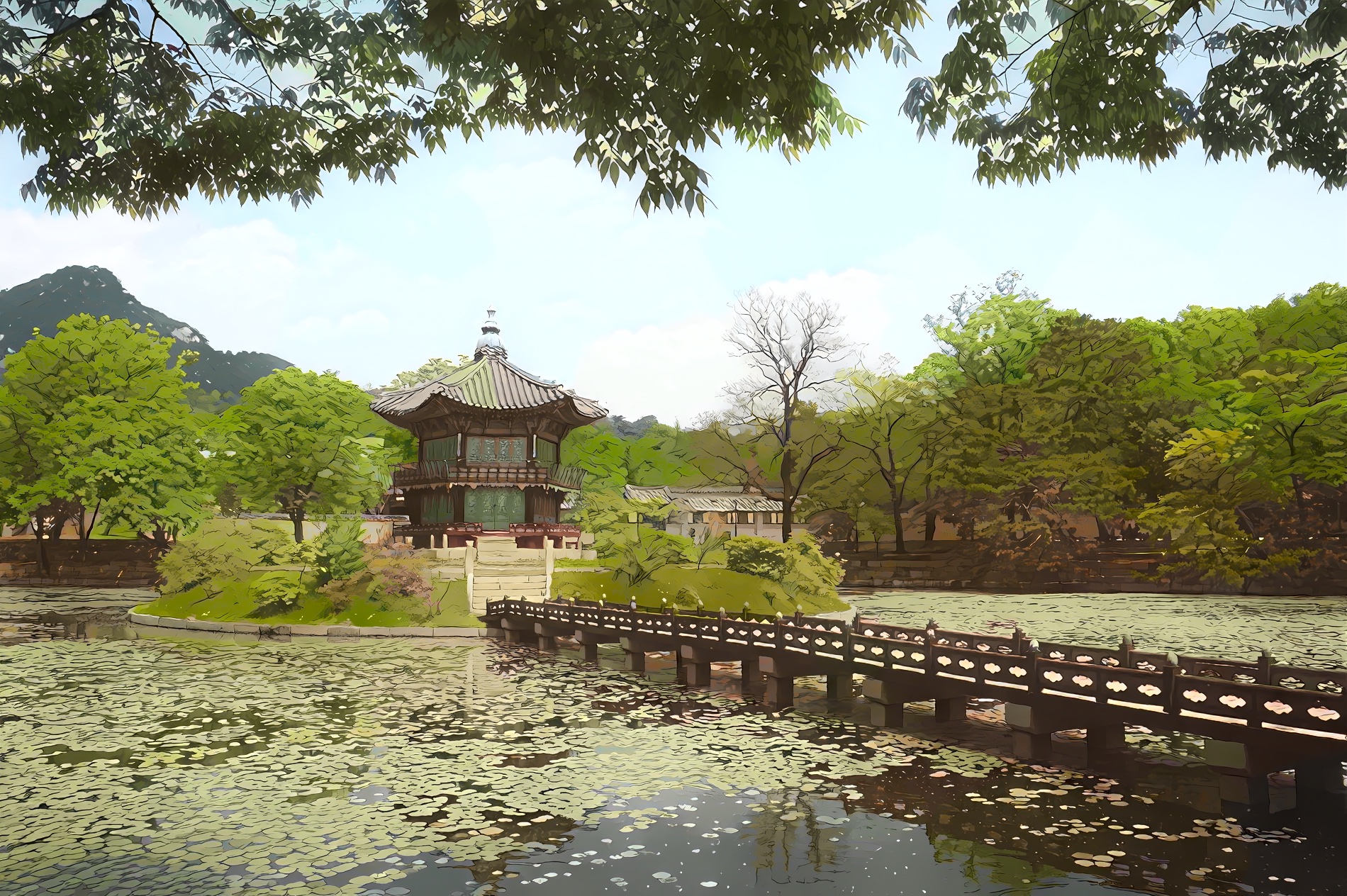}\hspace{\himg}
\includegraphics[width=\h]{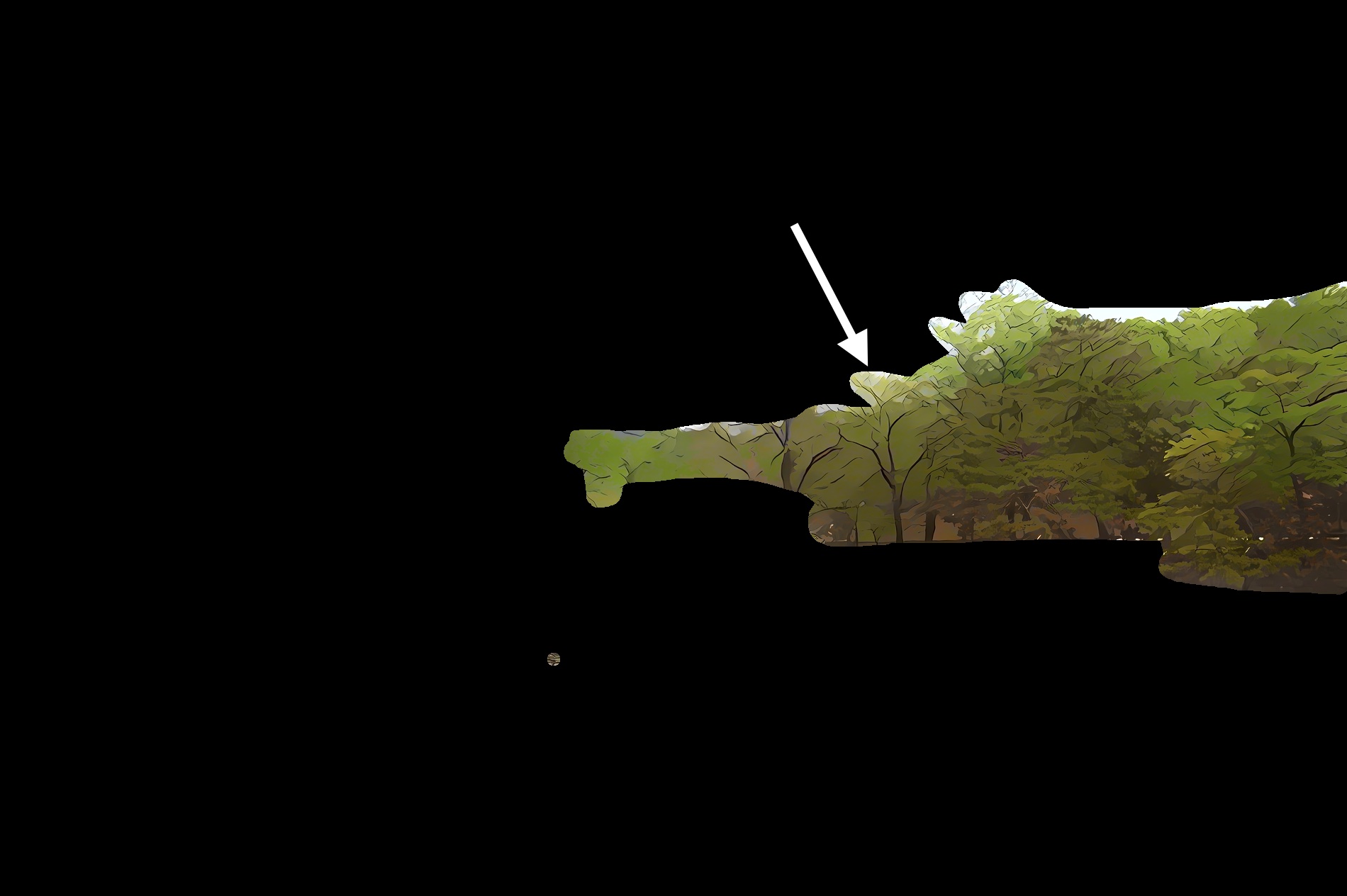}\hspace{\himg}
\includegraphics[width=\h]{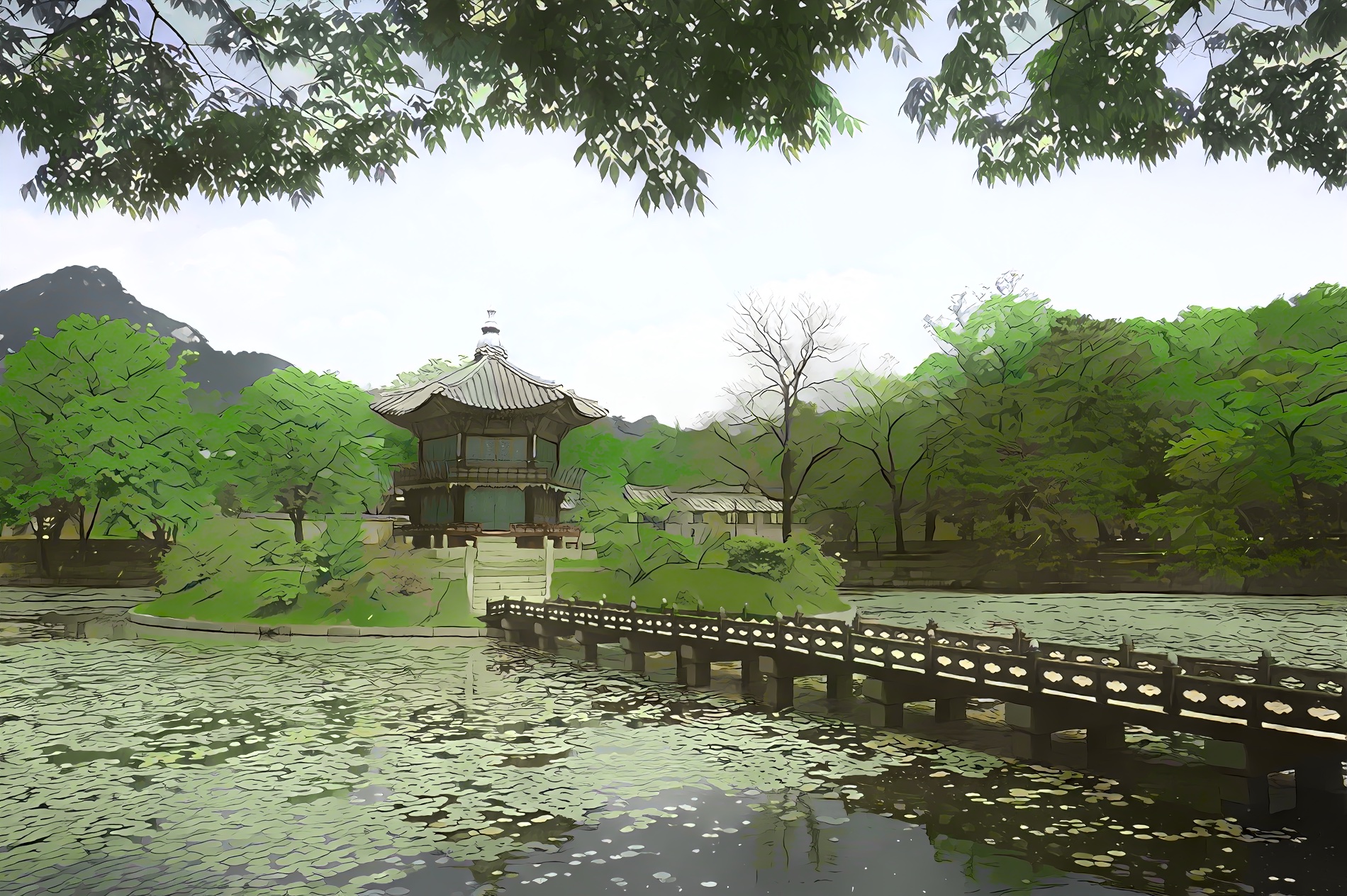}\hfill
\\
\makebox[\h][c]{Source photo}\hspace{\himg}
\makebox[\h][c]{Initial result}\hspace{\himg}
\makebox[\h][c]{$\rightarrow$ Change stroke \& abs.}\hspace{\himg}
\makebox[\h][c]{$\rightarrow$ Change color tone}\hspace{\himg}
\\
\includegraphics[width=\h]{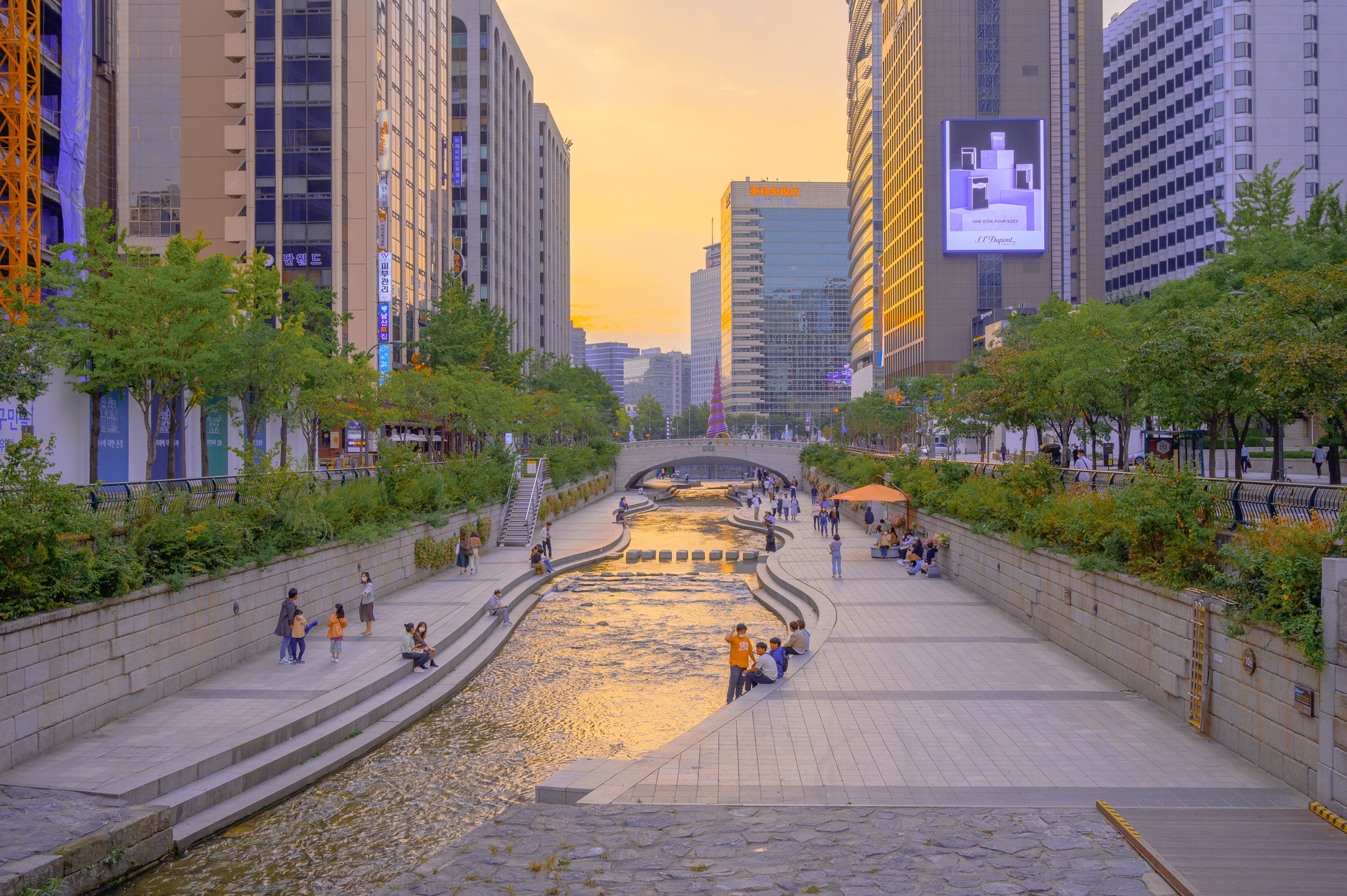}\hspace{\himg}
\includegraphics[width=\h]{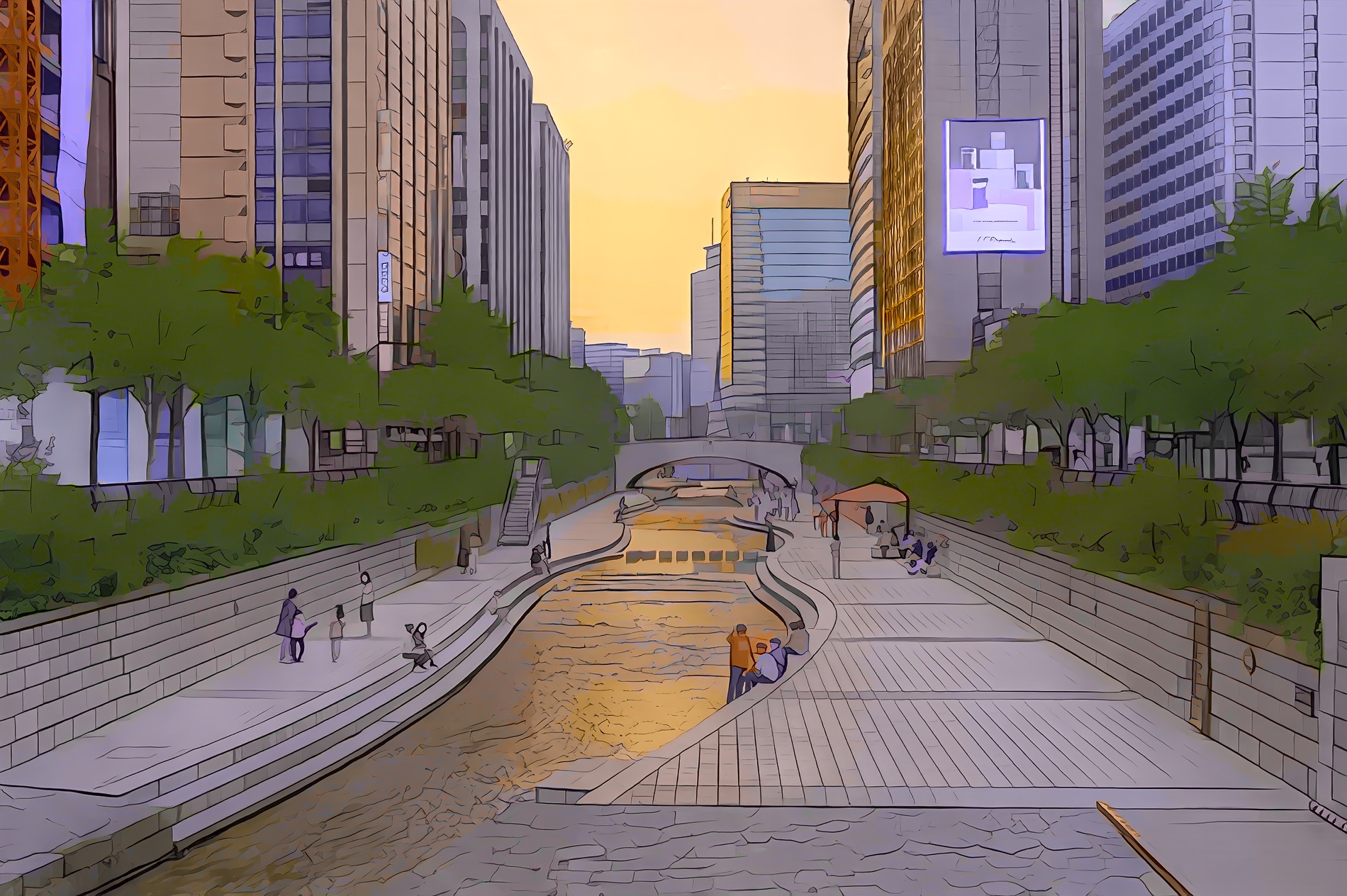}\hspace{\himg}
\includegraphics[width=\h]{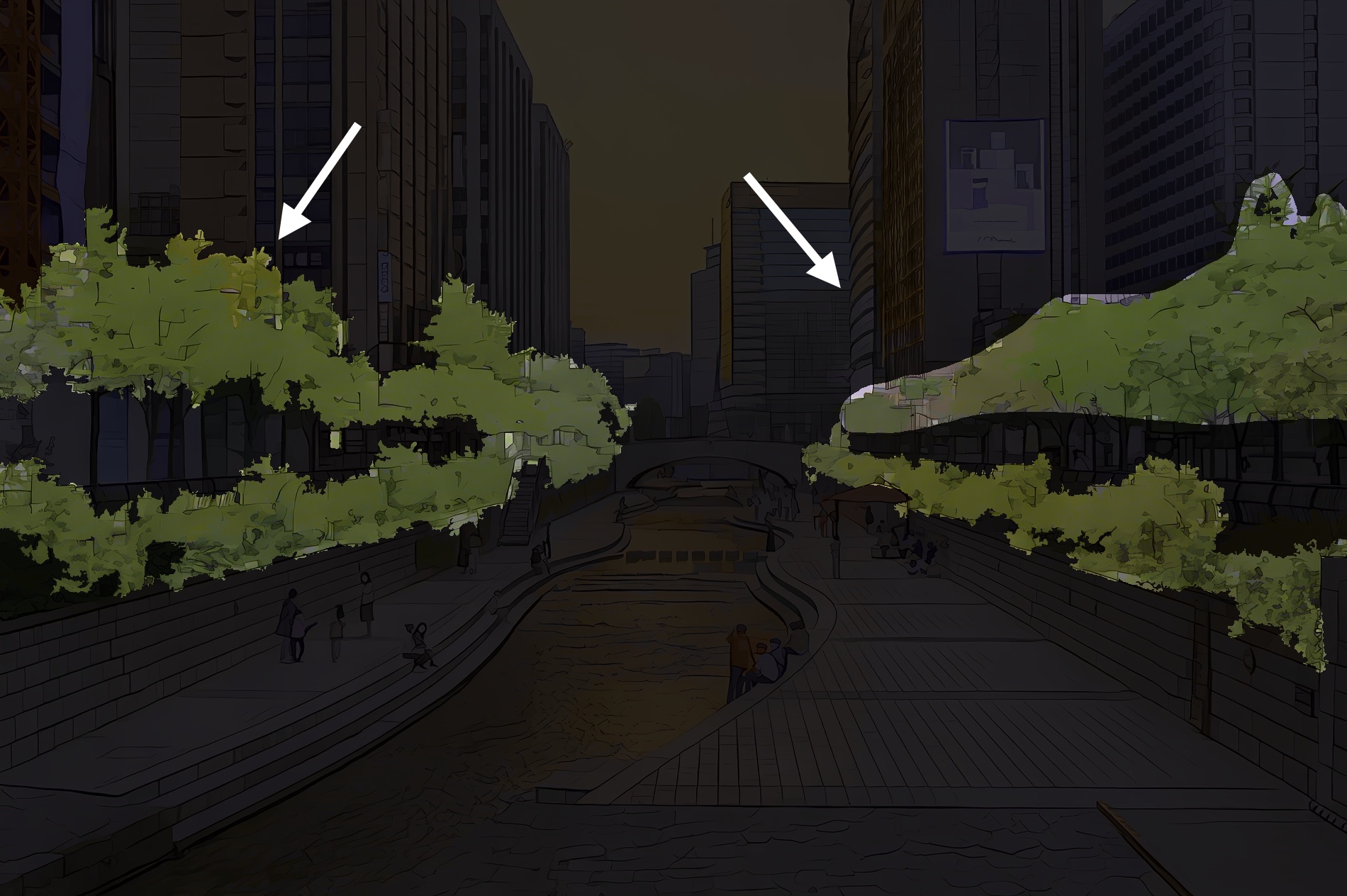}\hspace{\himg}
\includegraphics[width=\h]{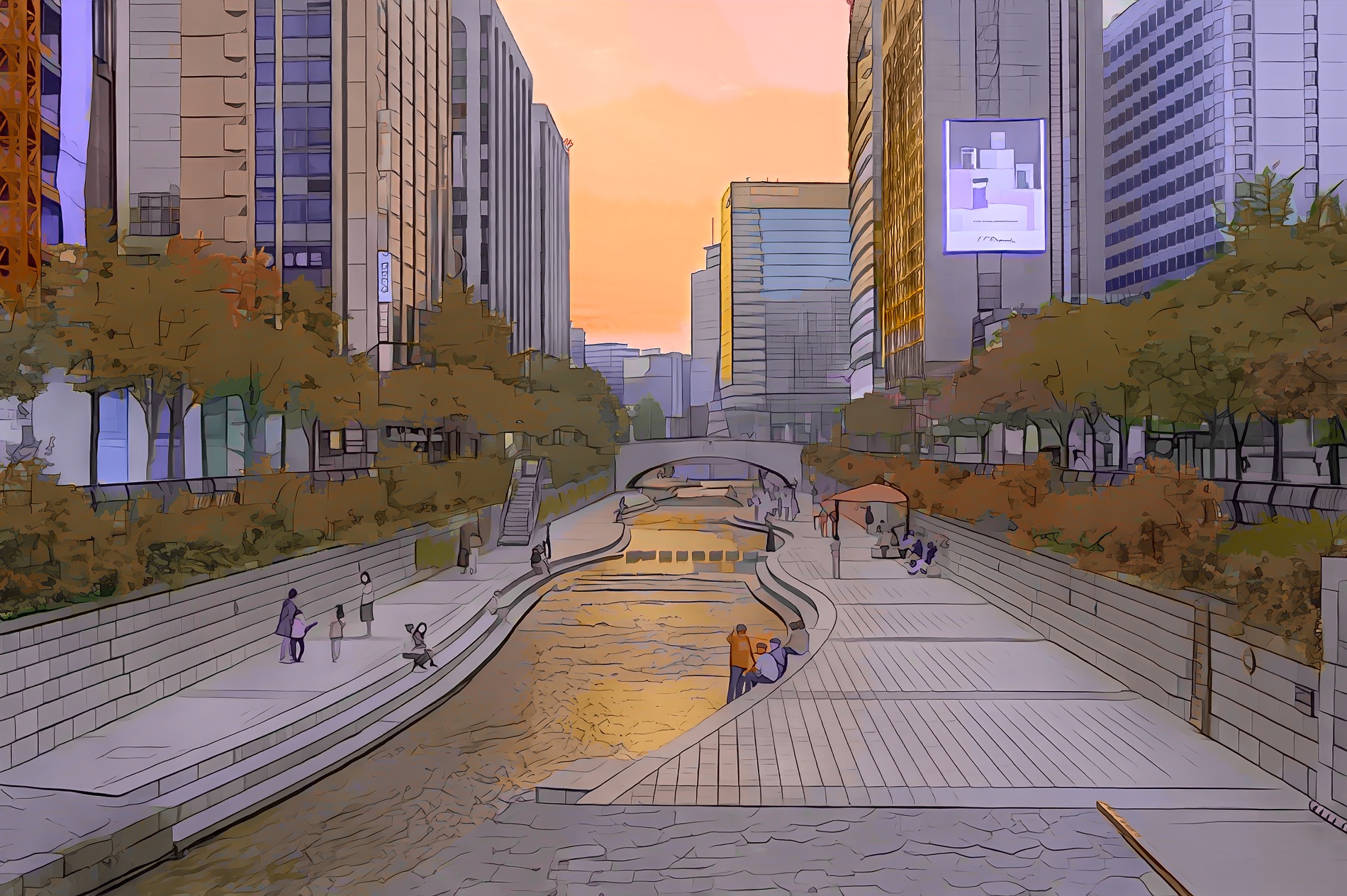}\hfill
\\
\makebox[\h][c]{Source photo}\hspace{\himg}
\makebox[\h][c]{Initial result}\hspace{\himg}
\makebox[\h][c]{$\rightarrow$ Change abstraction}\hspace{\himg}
\makebox[\h][c]{$\rightarrow$ Change color tone}\hspace{\himg}
\\
\includegraphics[width=\h]{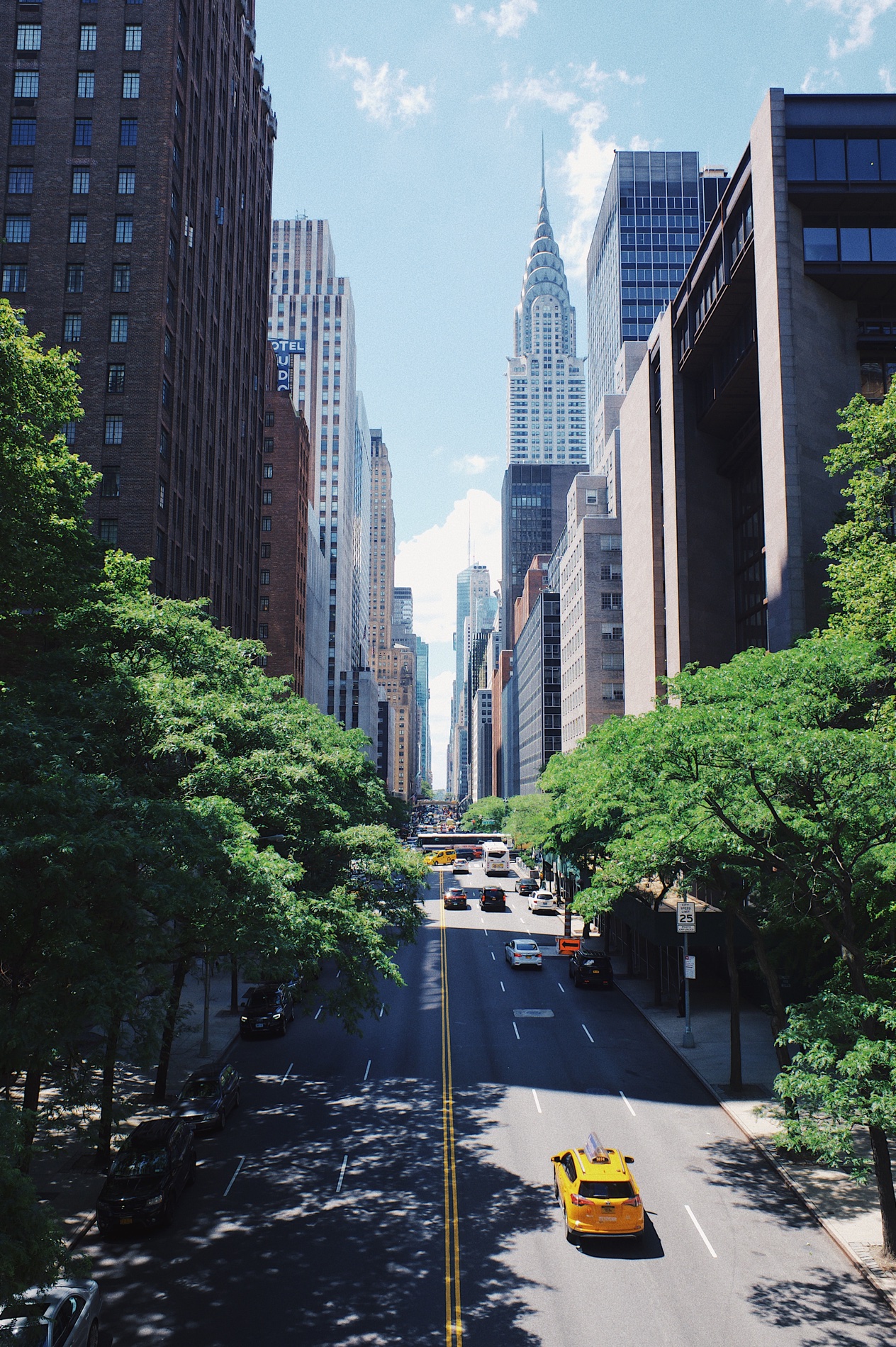}\hspace{\himg}
\includegraphics[width=\h]{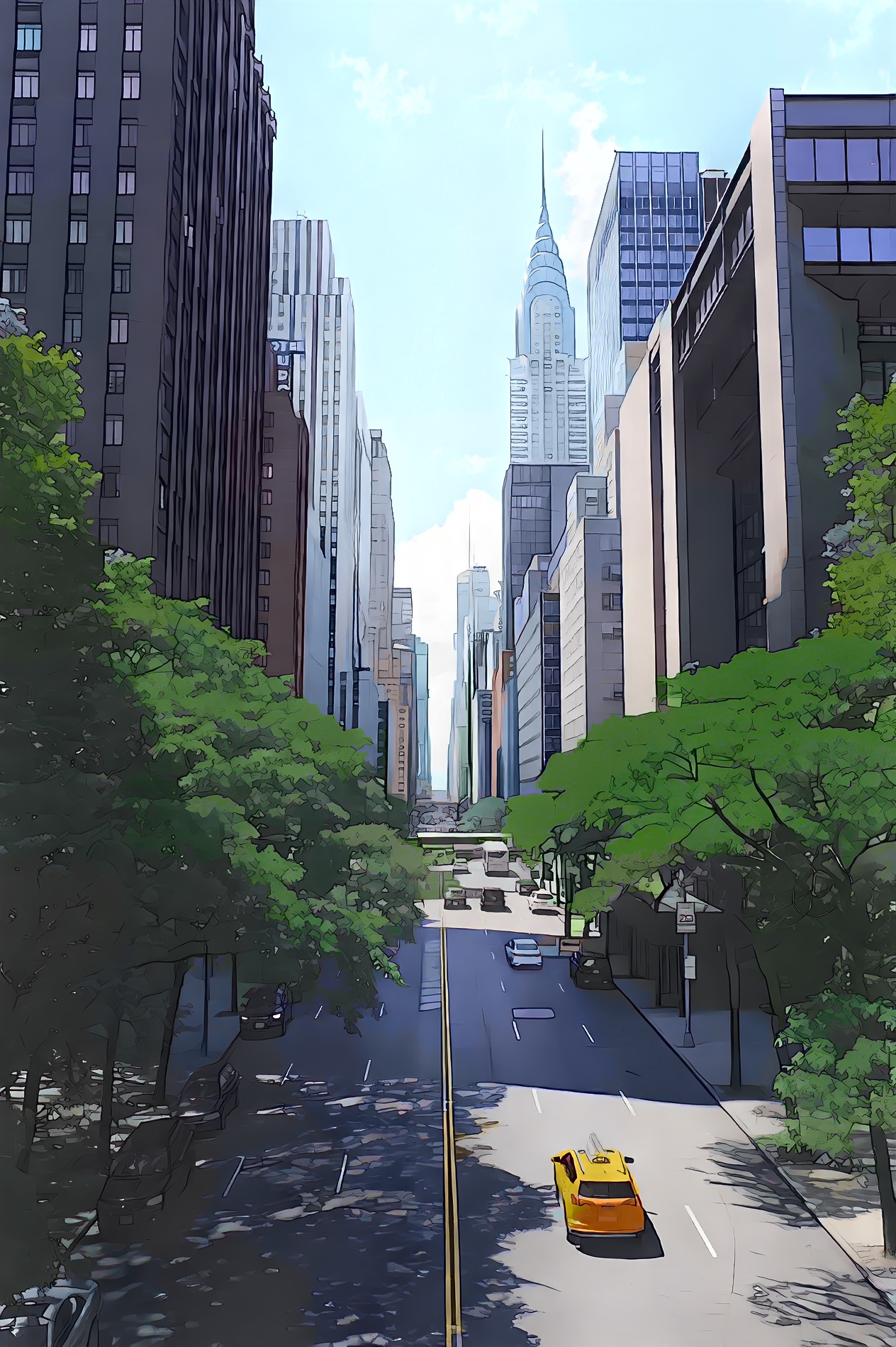}\hspace{\himg}
\includegraphics[width=\h]{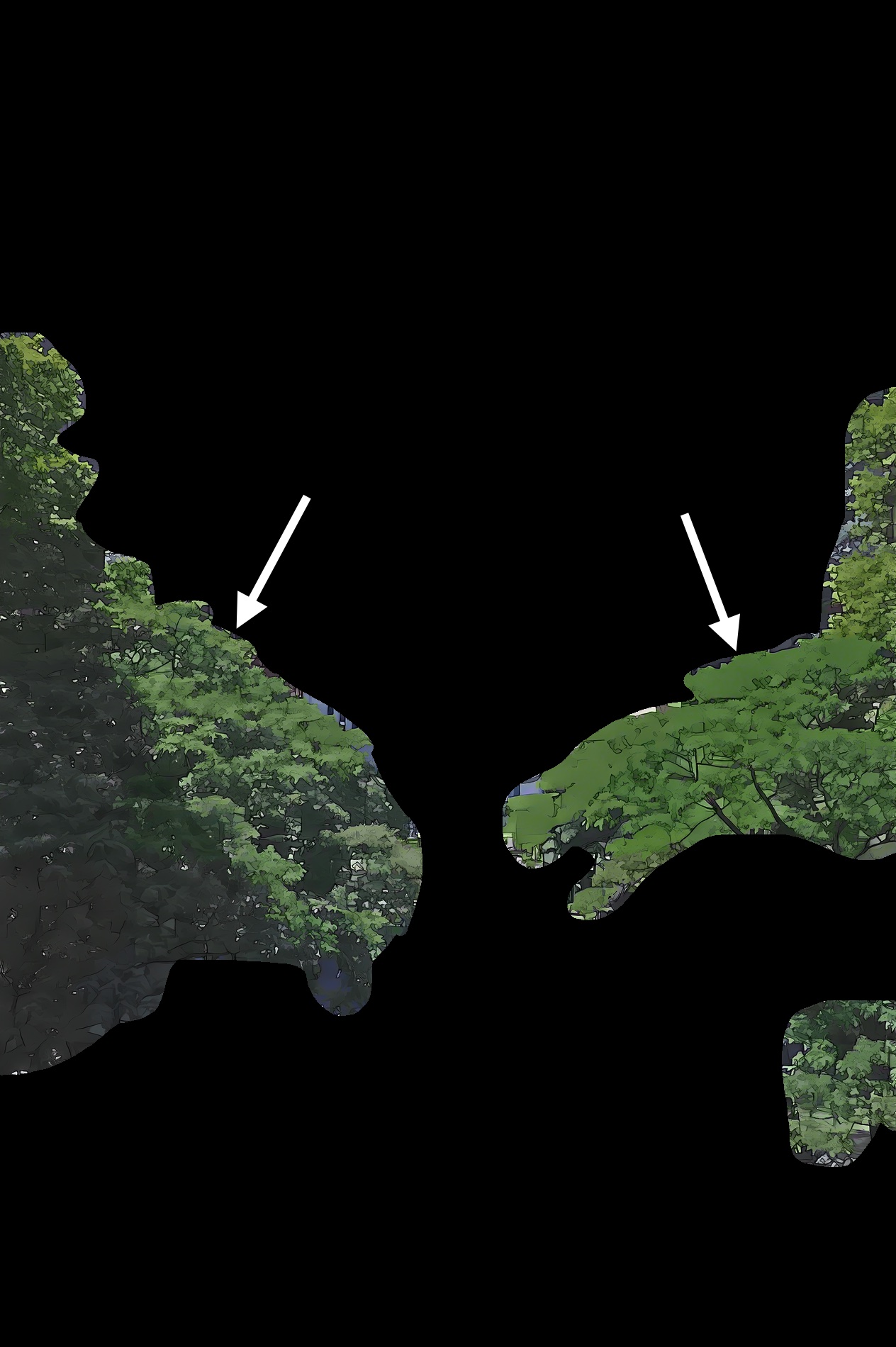}\hspace{\himg}
\includegraphics[width=\h]{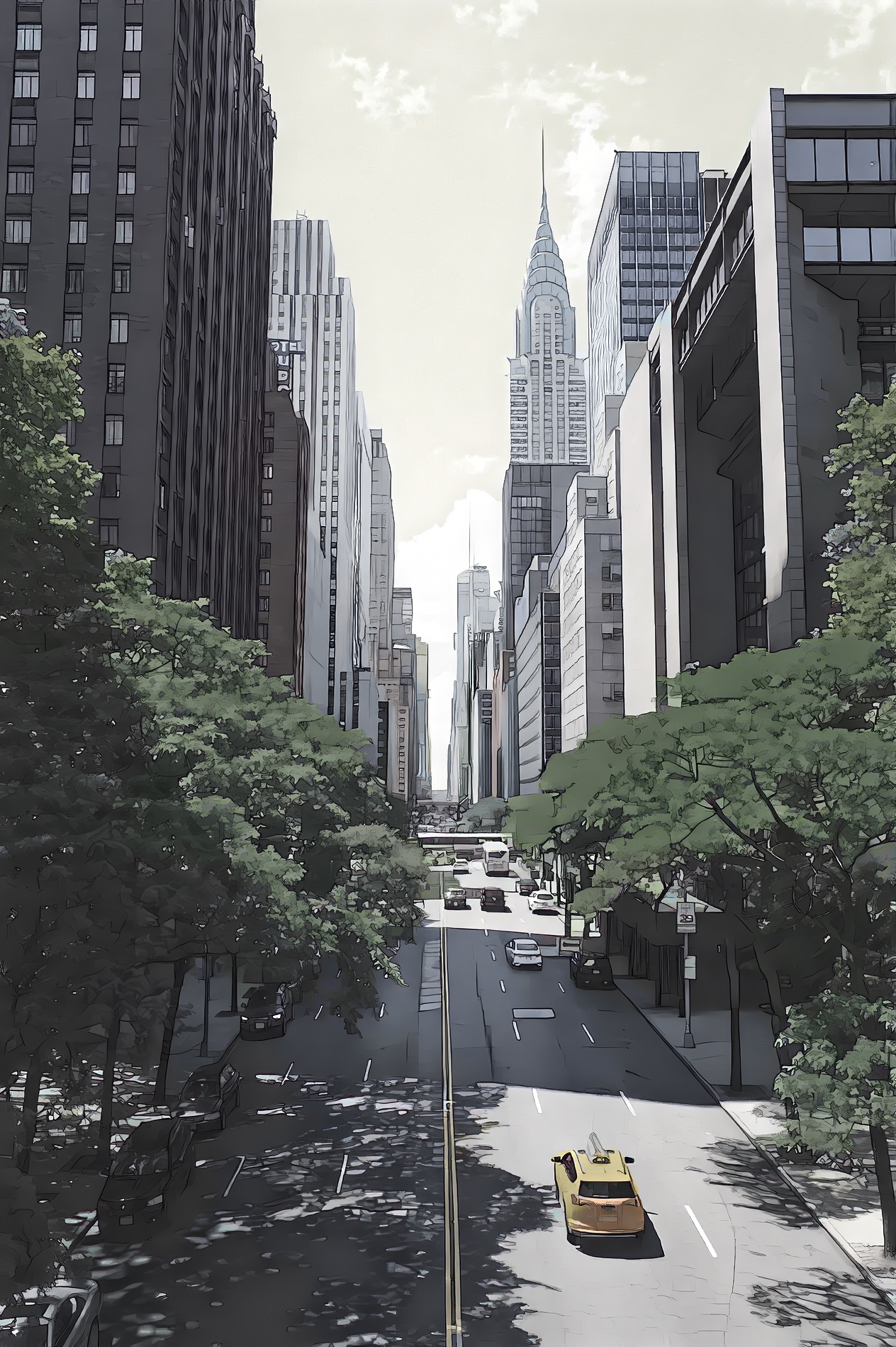}\hfill
\caption{\textbf{Results of interactive cartoonization.} In 3rd column, we highlight the altered region for better visualization.}
\label{fig:interactivity_suppl}
\end{figure*}
}

\newcommand{\figCompSupplOne}{
\begin{figure*}[t]
\newcommand{\h}{80mm}
\newcommand{\w}{2mm}
\newcommand{\himg}{5.0mm}
\centering
\makebox[\h][c]{(a) Source photo}\hspace{\himg}
\makebox[\h][c]{(b) Target cartoon}\hfill
\\
\includegraphics[width=\h]{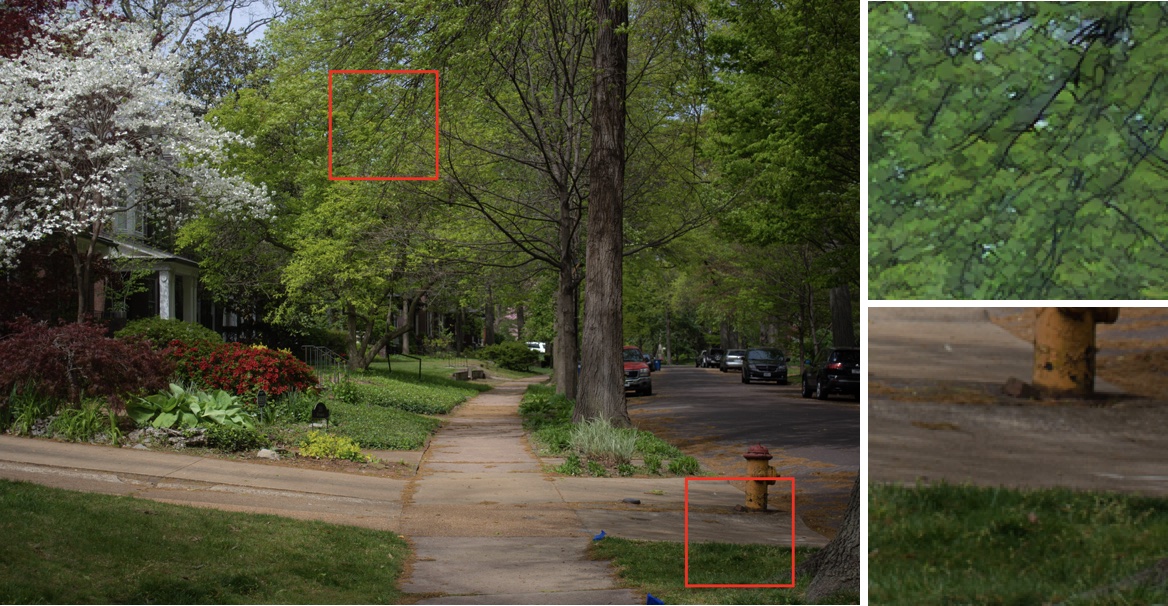}\hspace{\himg}
\includegraphics[width=\h]{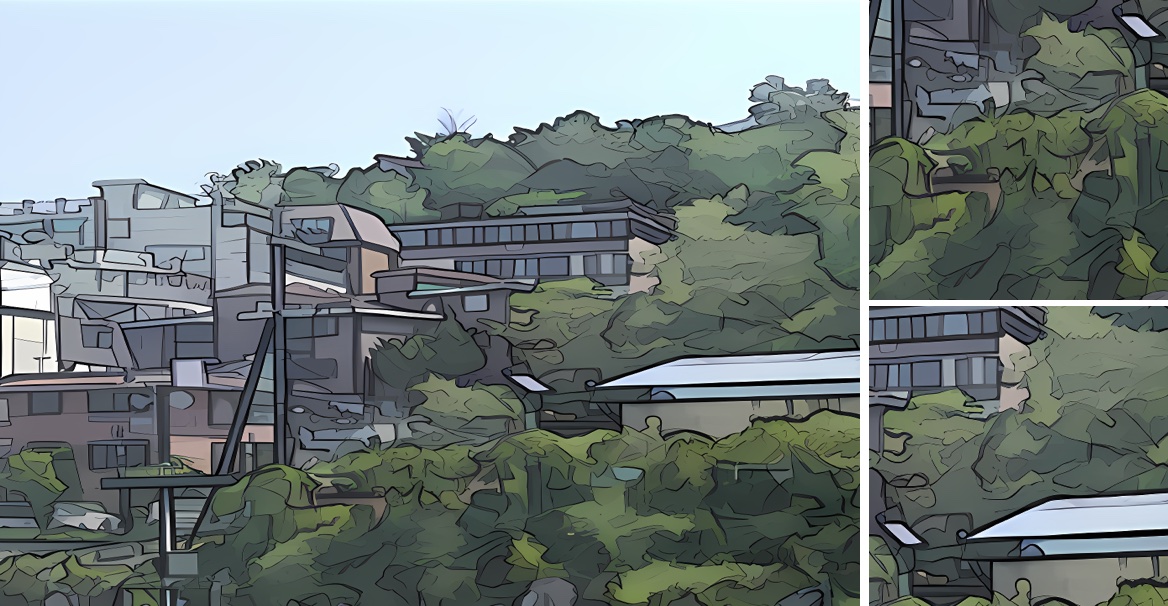}\hfill
\\\vspace{\w}
\makebox[\h][c]{(c) CartoonGAN~\cite{chen2018cartoongan}}\hspace{\himg}
\makebox[\h][c]{(d) AnimeGANv2~\cite{chen2019animegan}}\hfill
\\
\includegraphics[width=\h]{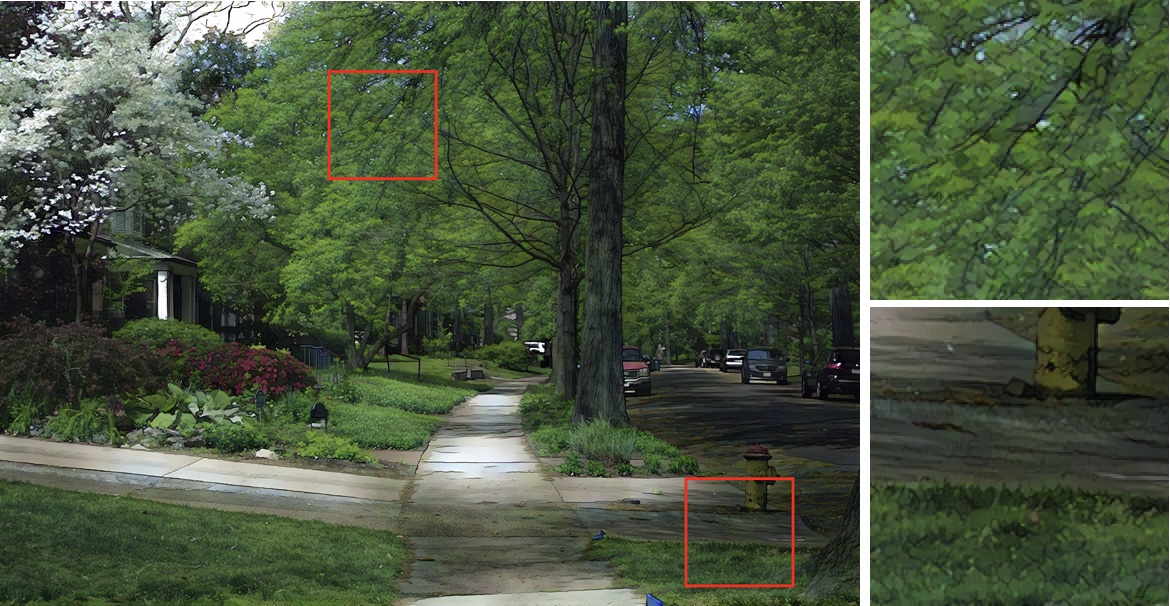}\hspace{\himg}
\includegraphics[width=\h]{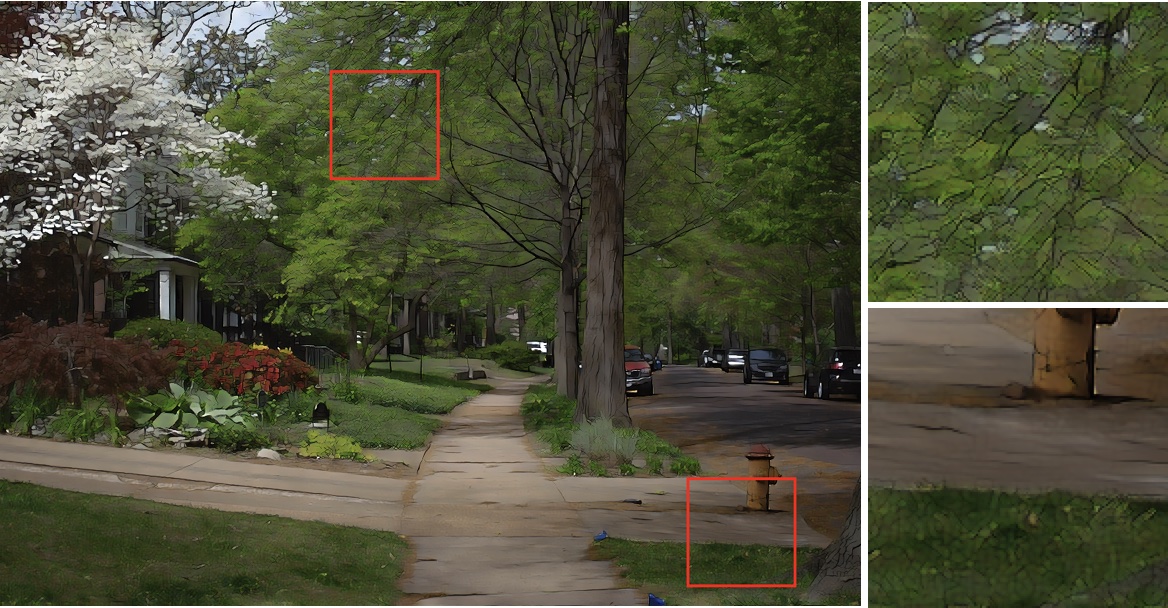}\hfill
\\\vspace{\w}
\makebox[\h][c]{(e) WhiteboxGAN~\cite{wang2020learning}}\hspace{\himg}
\makebox[\h][c]{(f) \textsc{Cartooner} (ours)}\hfill
\\
\includegraphics[width=\h]{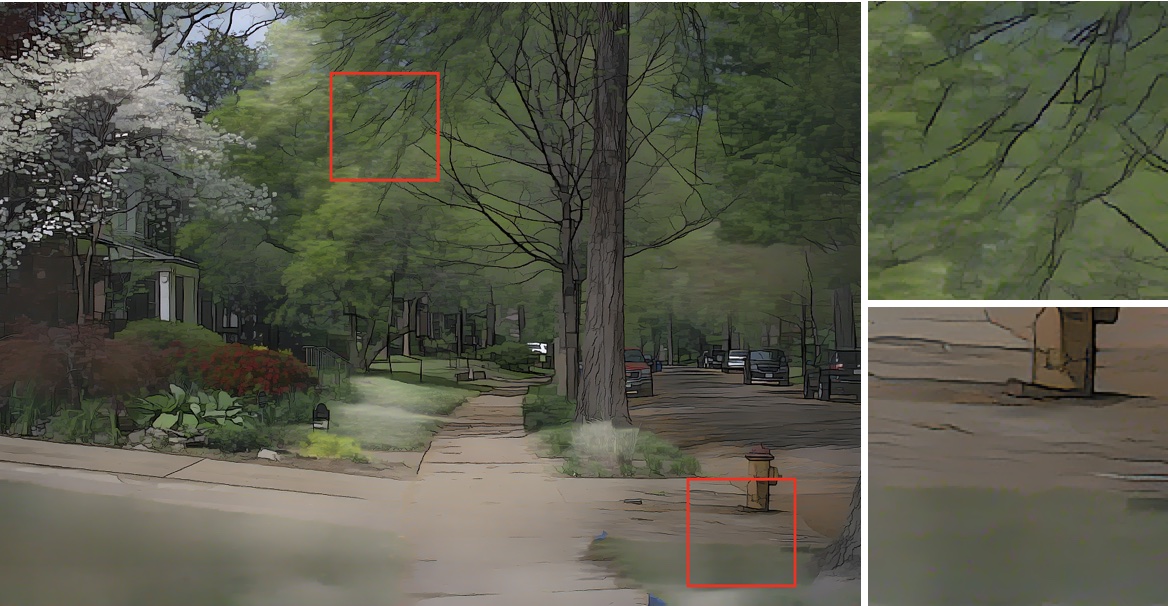}\hspace{\himg}
\includegraphics[width=\h]{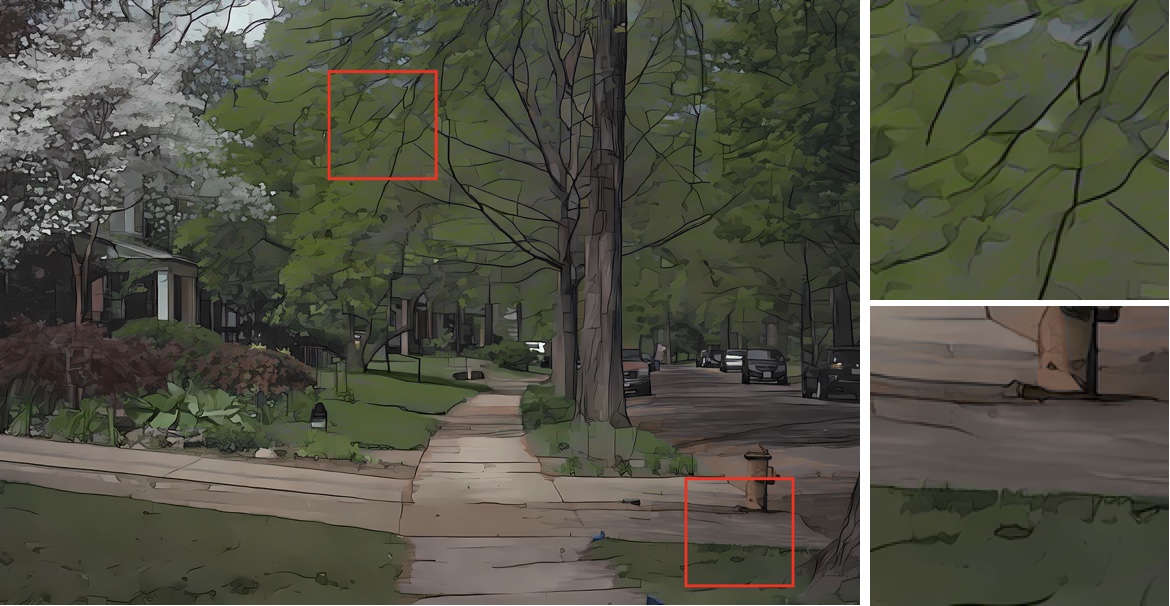}\hfill
\\
\caption{\textbf{Visual comparison.} Images along with source photos indicate exemplar images of the target cartoon. Best viewed in zoom.}
\label{fig:comp_suppl_one}
\end{figure*}
}

\newcommand{\figCompSupplTwo}{
\begin{figure*}[t]
\newcommand{\h}{80mm}
\newcommand{\w}{2mm}
\newcommand{\himg}{5.0mm}
\centering
\makebox[\h][c]{(a) Source photo}\hspace{\himg}
\makebox[\h][c]{(b) Target cartoon}\hfill
\\
\includegraphics[width=\h]{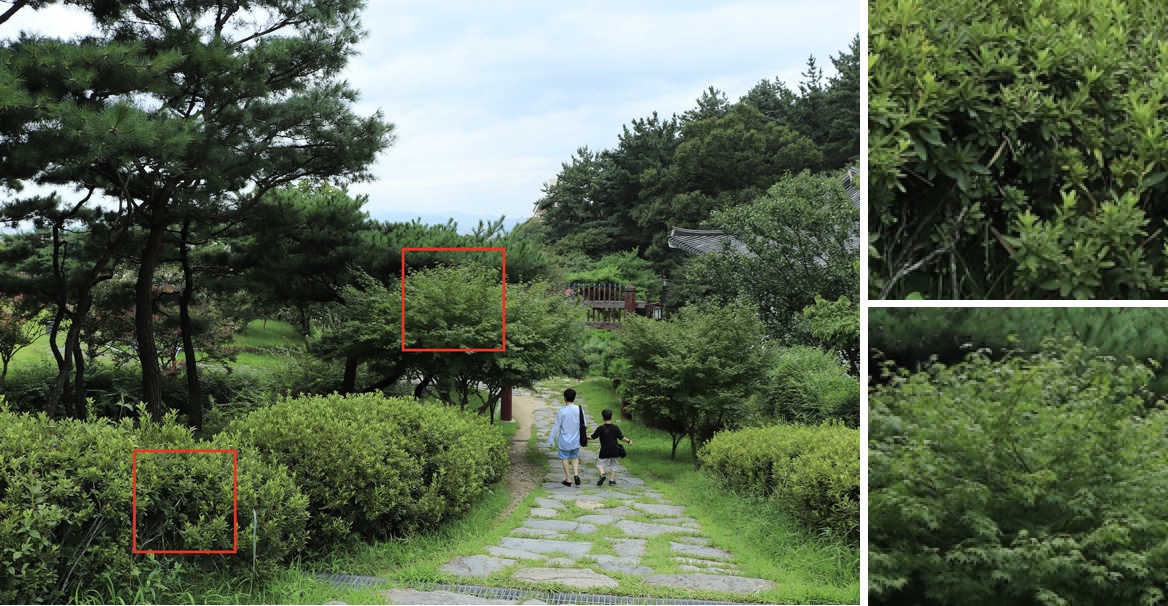}\hspace{\himg}
\includegraphics[width=\h]{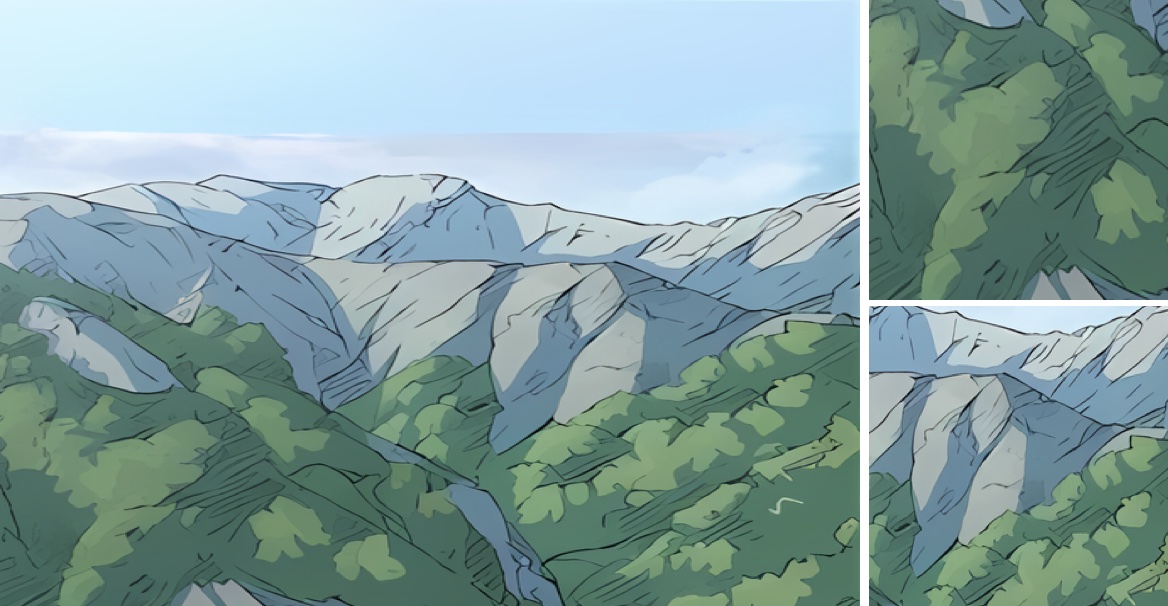}\hfill
\\\vspace{\w}
\makebox[\h][c]{(c) CartoonGAN~\cite{chen2018cartoongan}}\hspace{\himg}
\makebox[\h][c]{(d) AnimeGANv2~\cite{chen2019animegan}}\hfill
\\
\includegraphics[width=\h]{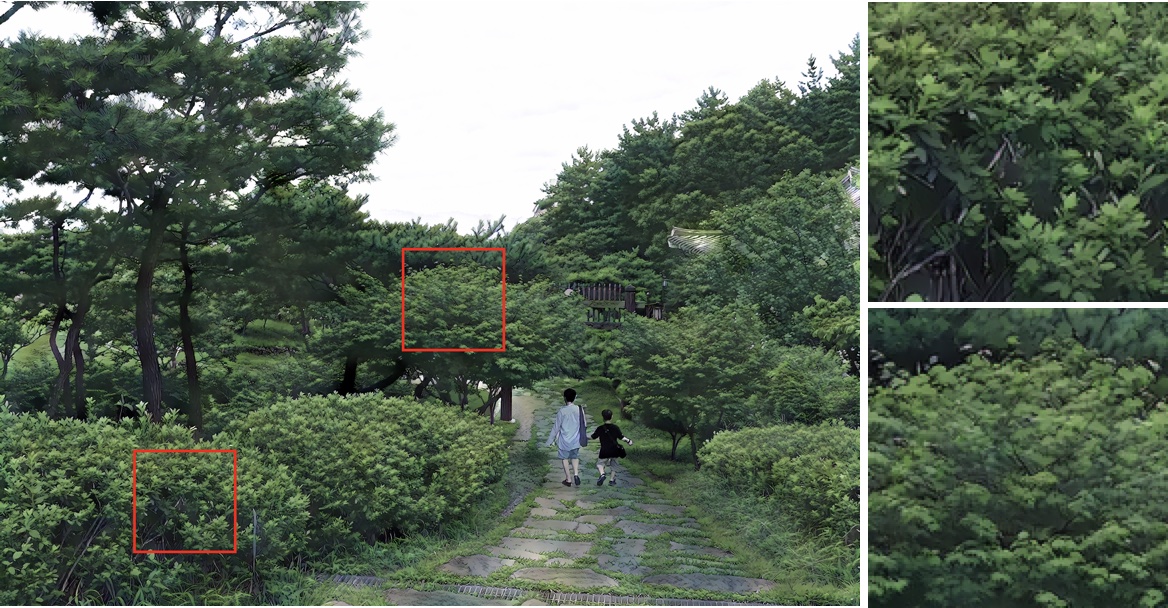}\hspace{\himg}
\includegraphics[width=\h]{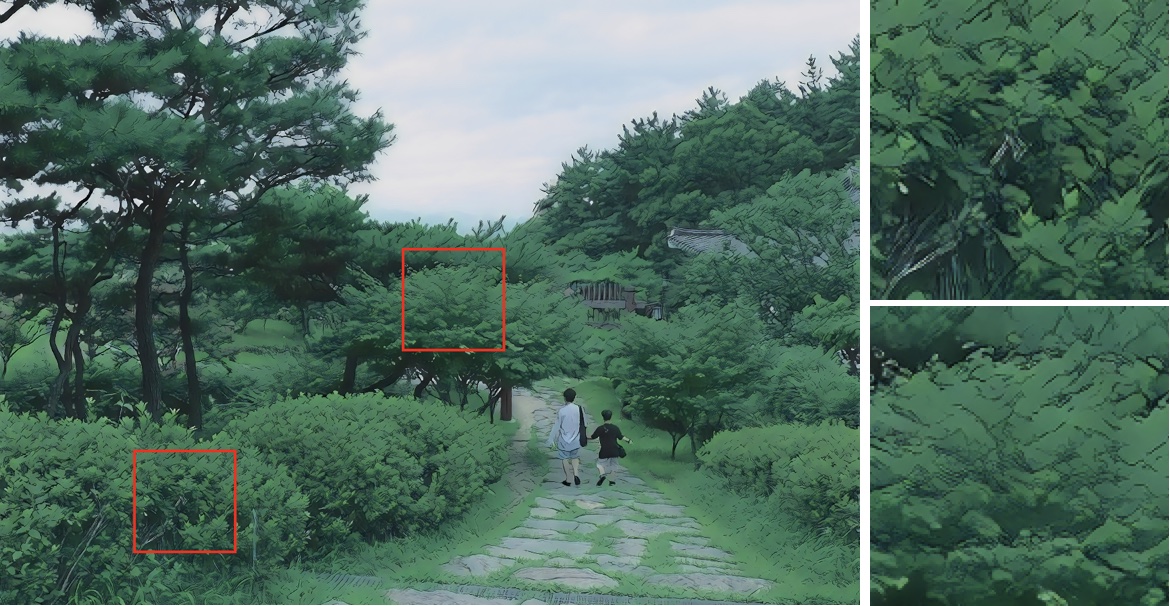}\hfill
\\\vspace{\w}
\makebox[\h][c]{(e) WhiteboxGAN~\cite{wang2020learning}}\hspace{\himg}
\makebox[\h][c]{(f) \textsc{Cartooner} (ours)}\hfill
\\
\includegraphics[width=\h]{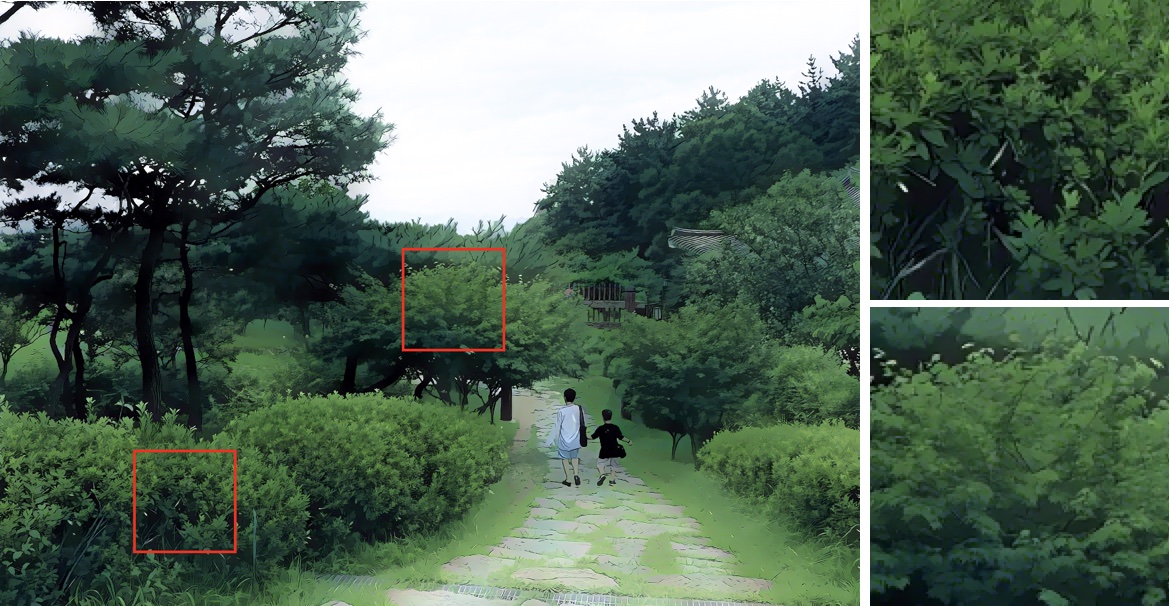}\hspace{\himg}
\includegraphics[width=\h]{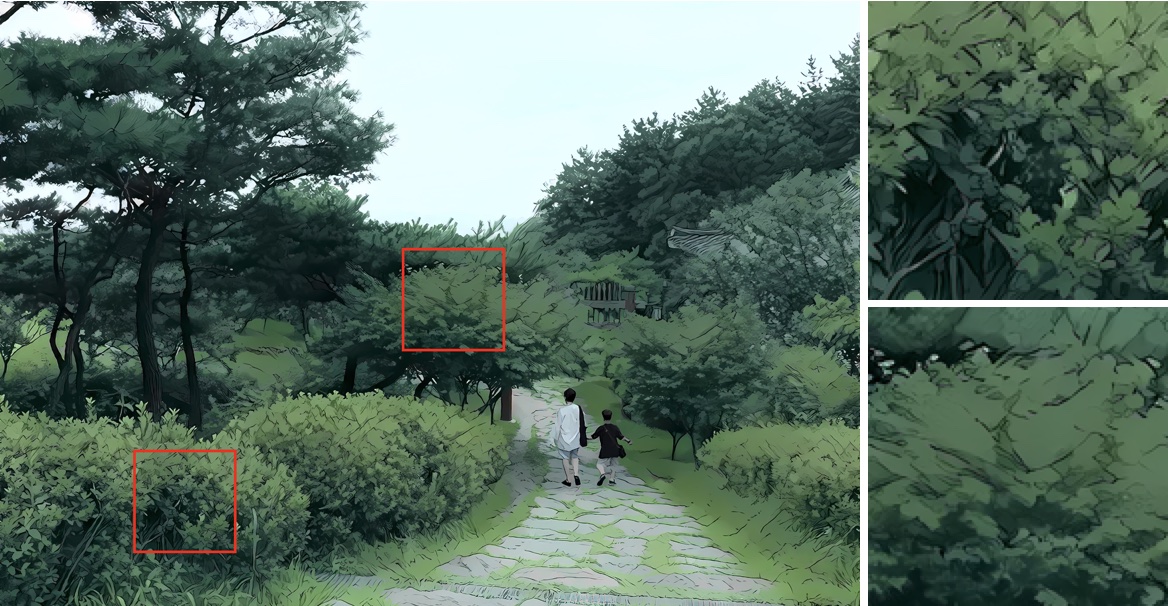}\hfill
\\
\caption{\textbf{Visual comparison.} Images along with source photos indicate exemplar images of the target cartoon. Best viewed in zoom.}
\label{fig:comp_suppl_two}
\end{figure*}
}

\newcommand{\figCompSupplThree}{
\begin{figure*}[t]
\newcommand{\h}{80mm}
\newcommand{\w}{2mm}
\newcommand{\himg}{5.0mm}
\centering
\makebox[\h][c]{(a) Source photo}\hspace{\himg}
\makebox[\h][c]{(b) Target cartoon}\hfill
\\
\includegraphics[width=\h]{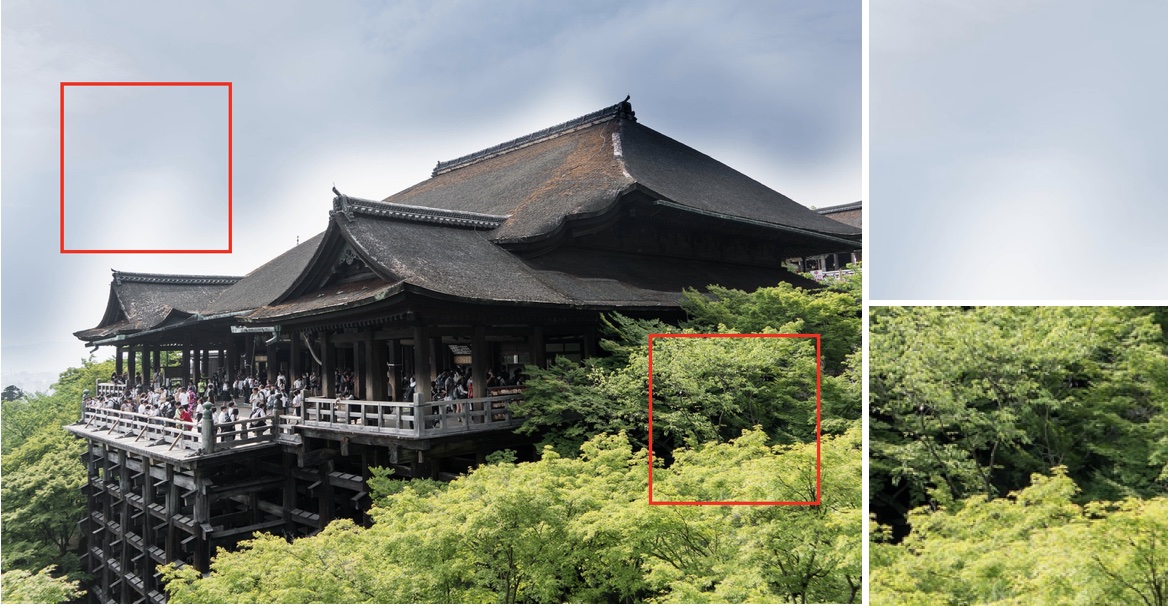}\hspace{\himg}
\includegraphics[width=\h]{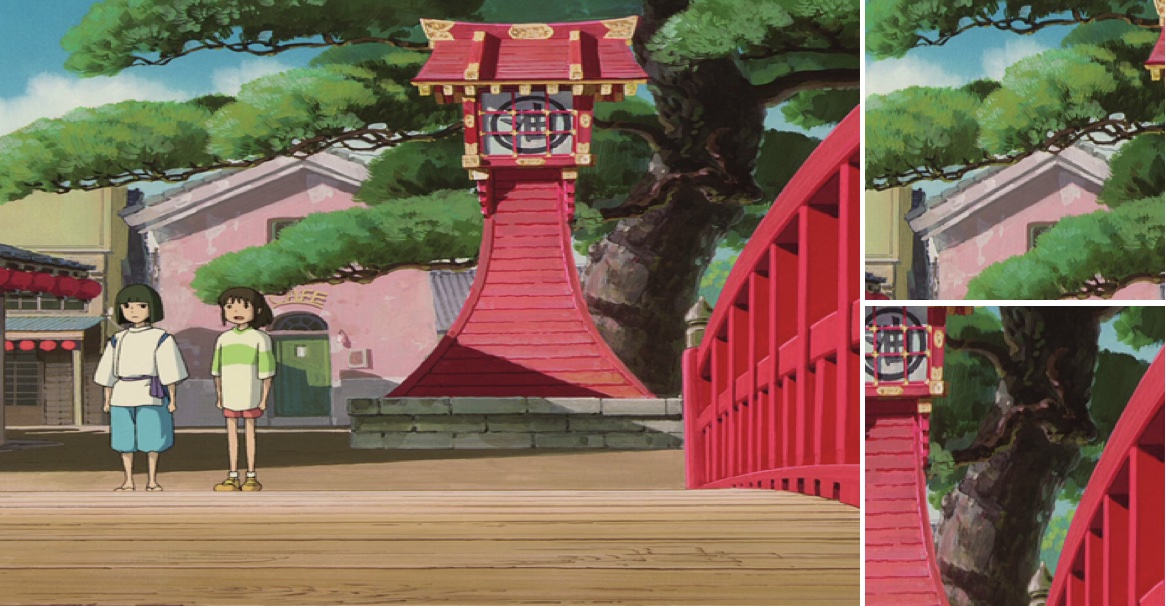}\hfill
\\\vspace{\w}
\makebox[\h][c]{(c) CartoonGAN~\cite{chen2018cartoongan}}\hspace{\himg}
\makebox[\h][c]{(d) AnimeGANv2~\cite{chen2019animegan}}\hfill
\\
\includegraphics[width=\h]{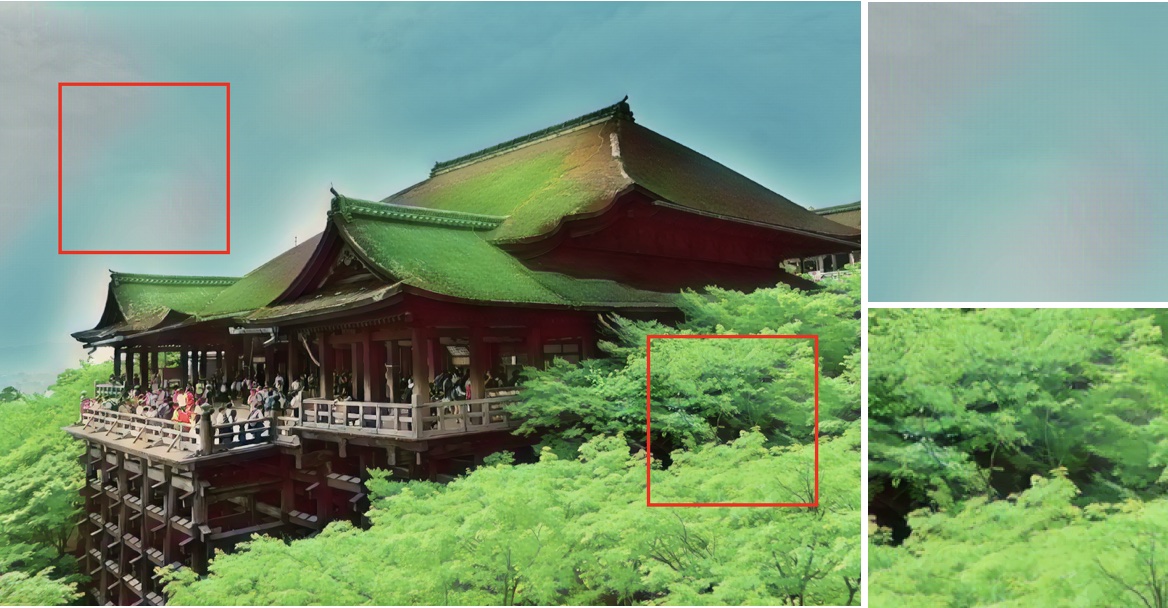}\hspace{\himg}
\includegraphics[width=\h]{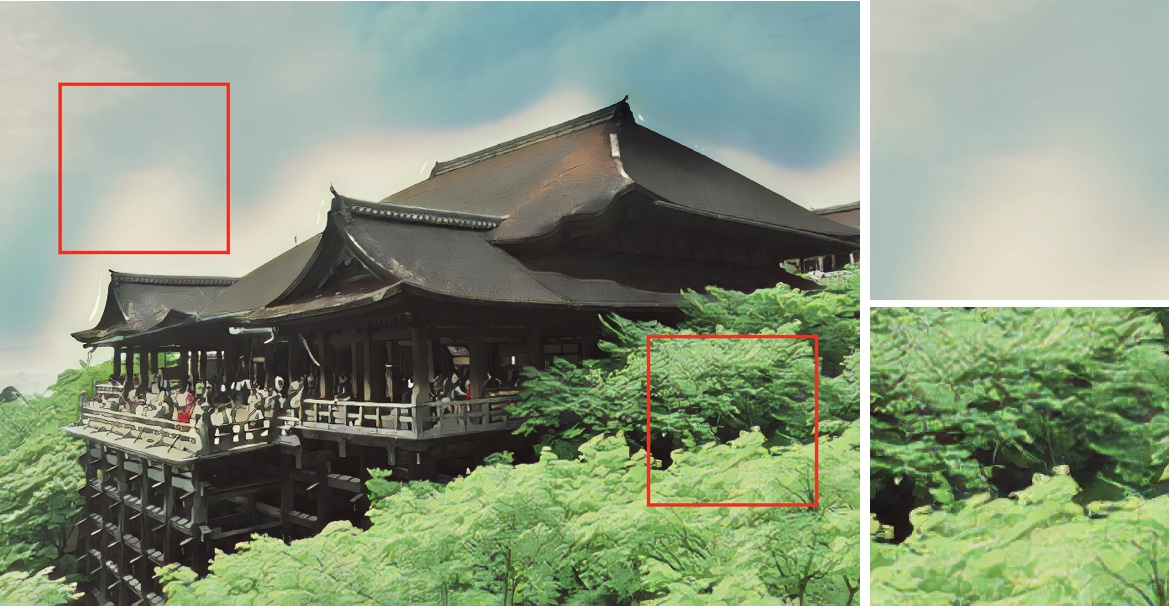}\hfill
\\\vspace{\w}
\makebox[\h][c]{(e) WhiteboxGAN~\cite{wang2020learning}}\hspace{\himg}
\makebox[\h][c]{(f) \textsc{Cartooner} (ours)}\hfill
\\
\includegraphics[width=\h]{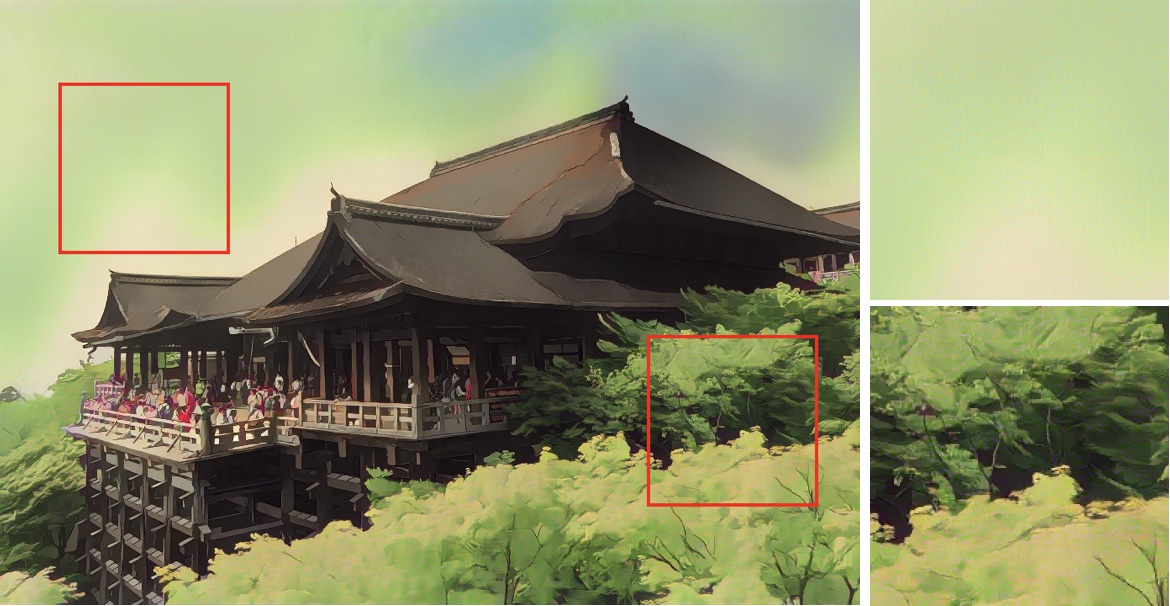}\hspace{\himg}
\includegraphics[width=\h]{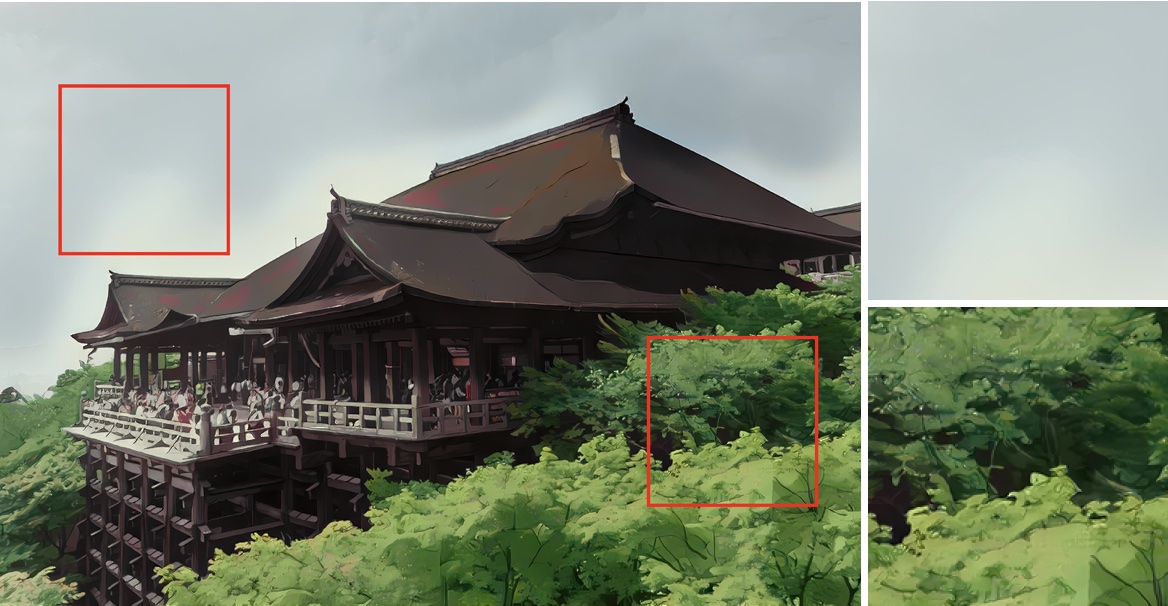}\hfill
\\
\caption{\textbf{Visual comparison.} Images along with source photos indicate exemplar images of the target cartoon. Best viewed in zoom.}
\label{fig:comp_suppl_three}
\end{figure*}
}

\newcommand{\figCompSupplFour}{
\begin{figure*}[t]
\newcommand{\h}{80mm}
\newcommand{\w}{2mm}
\newcommand{\himg}{5.0mm}
\centering
\makebox[\h][c]{(a) Source photo}\hspace{\himg}
\makebox[\h][c]{(b) Target cartoon}\hfill
\\
\includegraphics[width=\h]{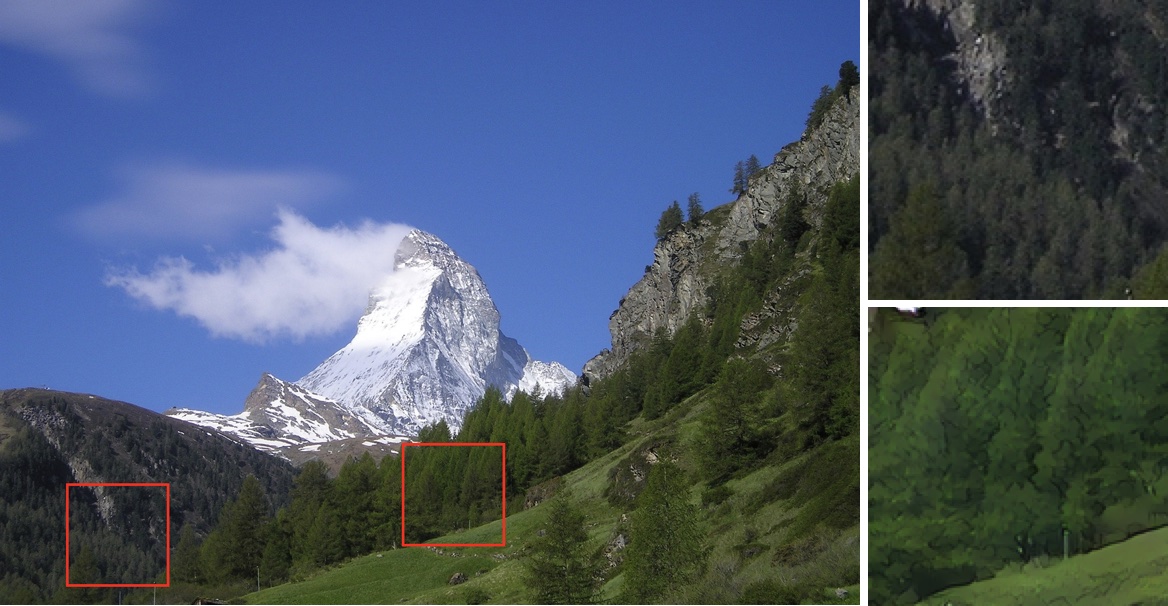}\hspace{\himg}
\includegraphics[width=\h]{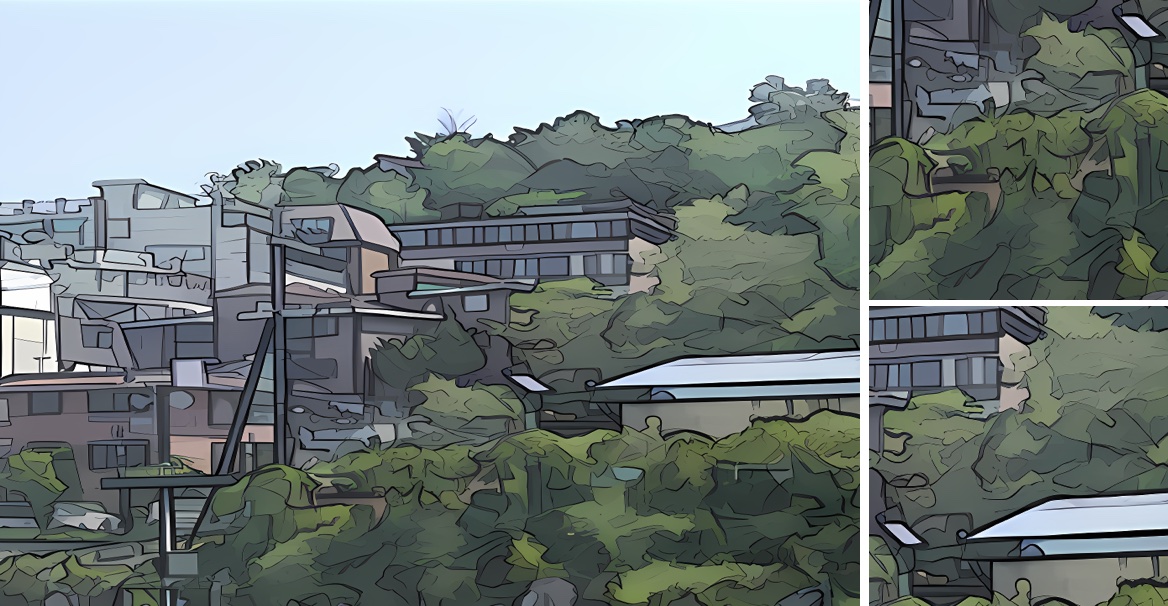}\hfill
\\\vspace{\w}
\makebox[\h][c]{(c) CartoonGAN~\cite{chen2018cartoongan}}\hspace{\himg}
\makebox[\h][c]{(d) AnimeGANv2~\cite{chen2019animegan}}\hfill
\\
\includegraphics[width=\h]{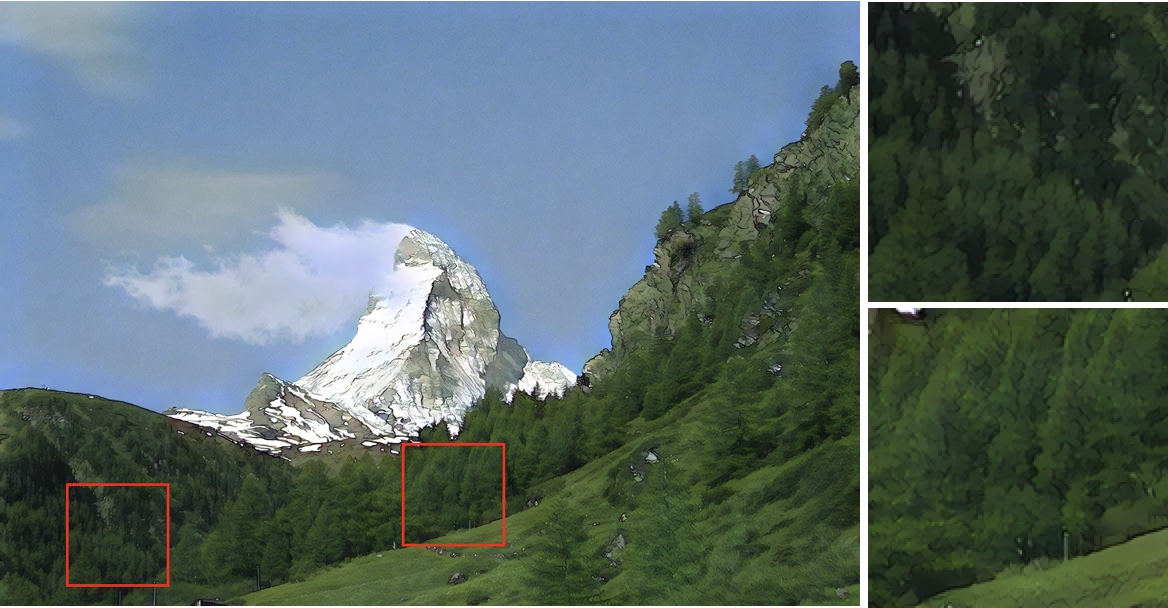}\hspace{\himg}
\includegraphics[width=\h]{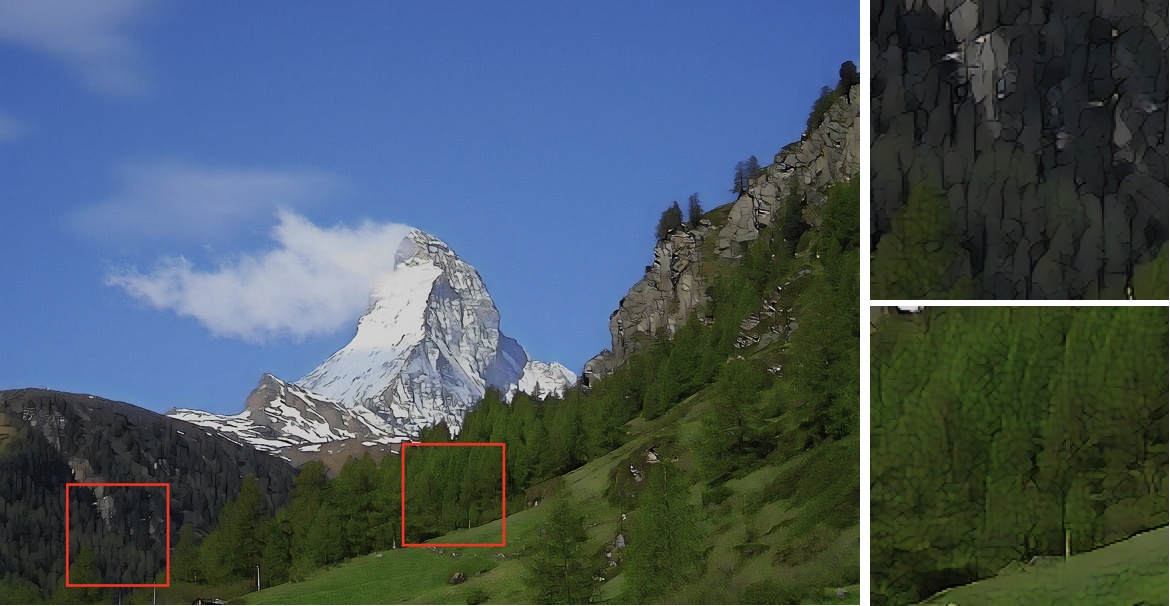}\hfill
\\\vspace{\w}
\makebox[\h][c]{(e) WhiteboxGAN~\cite{wang2020learning}}\hspace{\himg}
\makebox[\h][c]{(f) \textsc{Cartooner} (ours)}\hfill
\\
\includegraphics[width=\h]{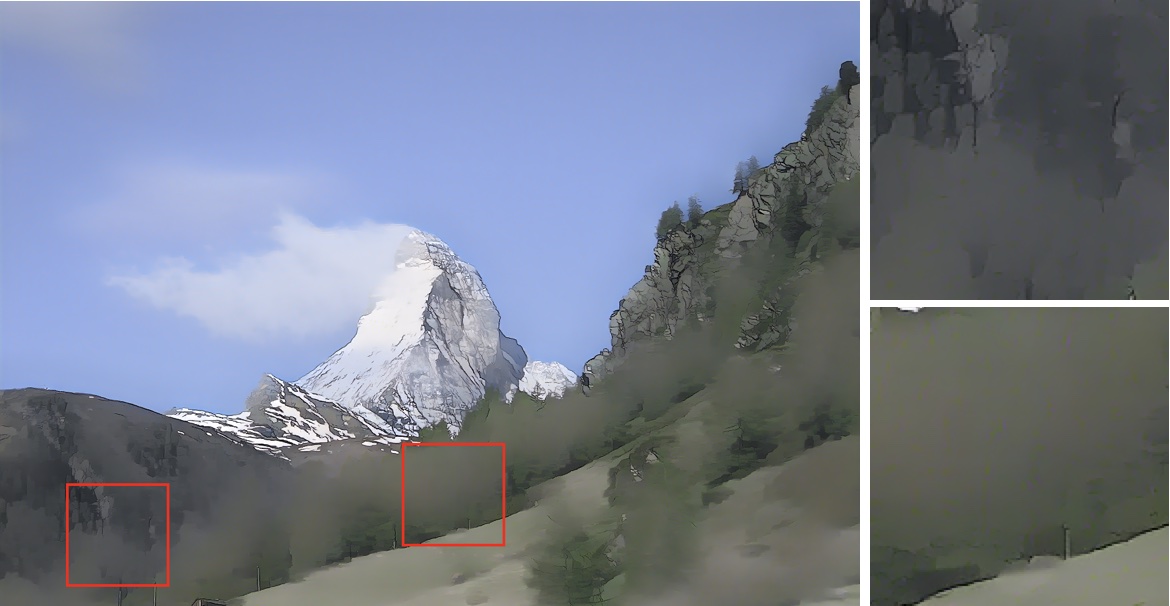}\hspace{\himg}
\includegraphics[width=\h]{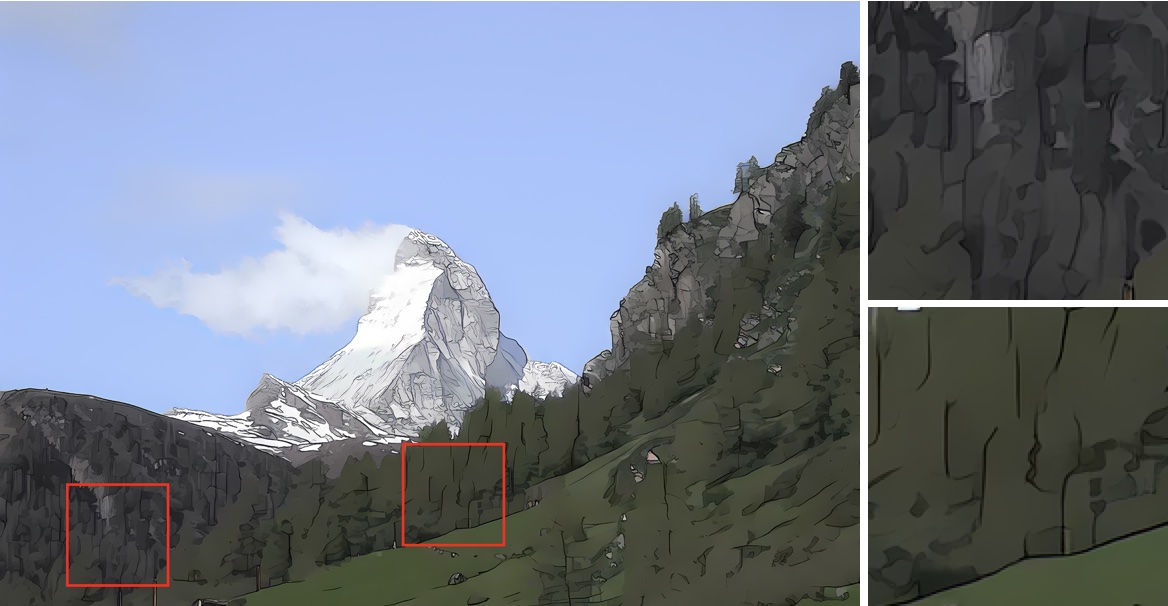}\hfill
\\
\caption{\textbf{Visual comparison.} Images along with source photos indicate exemplar images of the target cartoon. Best viewed in zoom.}
\label{fig:comp_suppl_four}
\end{figure*}
}

\newcommand{\tableNetworkArchitecture}{ 
\begin{table}[t]
\centering
\setlength\tabcolsep{11pt}
\caption{\textbf{Network architecture.} }
\begin{tabular}{c|c|c}
\hline
\multirow{2}{*}{Layer} & \multirow{2}{*}{Output size} & \multirow{2}{*}{Building block}
\\
& &
\\\hline\hline
\multicolumn{3}{c}{Shared Encoder ($E_{shared}$)}
\\\hline
conv1 & 256$\times$256 & Conv(3, 32, 7, 1, 3)
\\\hline
\multirow{2}{*}{conv2} & \multirow{2}{*}{128$\times$128} & Conv(32, 32, 3, 2, 1)
\\
& & Conv(32, 64, 3, 1, 1)
\\\hline
\multirow{2}{*}{conv3} & \multirow{2}{*}{64$\times$64} & Conv(64, 64, 3, 2, 1)
\\
& & Conv(64, 128, 3, 1, 1)
\\\hline
conv4 & 64$\times$64 & 3 $\times$ ResNeXt(128, 128)
\\\hline\hline
\multicolumn{3}{c}{Texture Decoder ($D_{texture}$)}
\\\hline
tex0 & 64$\times$64 & \textit{texture controller}
\\\hline
\multirow{3}{*}{tex1} & \multirow{3}{*}{128$\times$128} & Conv(128, 128, 3, 1, 1)
\\
& & Conv(128, 64, 3, 1, 1)
\\
& & Upsample(2)
\\\hline
\multirow{3}{*}{tex2} & \multirow{3}{*}{256$\times$256} & Conv(64, 64, 3, 1, 1)
\\
& & Conv(64, 32, 3, 1, 1)
\\
& & Upsample(2)
\\\hline
\multirow{2}{*}{tex3} & \multirow{2}{*}{256$\times$256} & Conv(32, 32, 3, 1, 1)
\\
& & Conv(32, 1, 7, 1, 3)
\\\hline\hline
\multicolumn{3}{c}{Color Decoder ($D_{color}$)}
\\\hline
col0 & 64$\times$64 &  ResNeXt(128, 128)
\\\hline
\multirow{3}{*}{col1} & \multirow{3}{*}{128$\times$128} & Conv(128, 128, 3, 1, 1)
\\
& & Conv(128, 64, 3, 1, 1)
\\
& & Upsample(2)
\\\hline
\multirow{3}{*}{col2} & \multirow{3}{*}{256$\times$256} & Conv(64+3, 64, 3, 1, 1)
\\
& & Conv(64, 32, 3, 1, 1)
\\
& & Upsample(2)
\\\hline
\multirow{2}{*}{col3} & \multirow{2}{*}{256$\times$256} & Conv(32+3, 32, 3, 1, 1)
\\
& & Conv(32, 2, 7, 1, 3)
\\\hline
\end{tabular}
\label{table:network_architecture}
\end{table}
}

\newcommand{\figVsSD}{
\begin{figure*}[t]
\newcommand{\h}{85mm}
\newcommand{\himg}{1.0mm}
\centering
\makebox[\h][c]{(a) Source Photo}\hspace{\himg}
\makebox[\h][c]{(b) \texttt{Cartooner} (ours)}\hfill
\\
\includegraphics[width=\h]{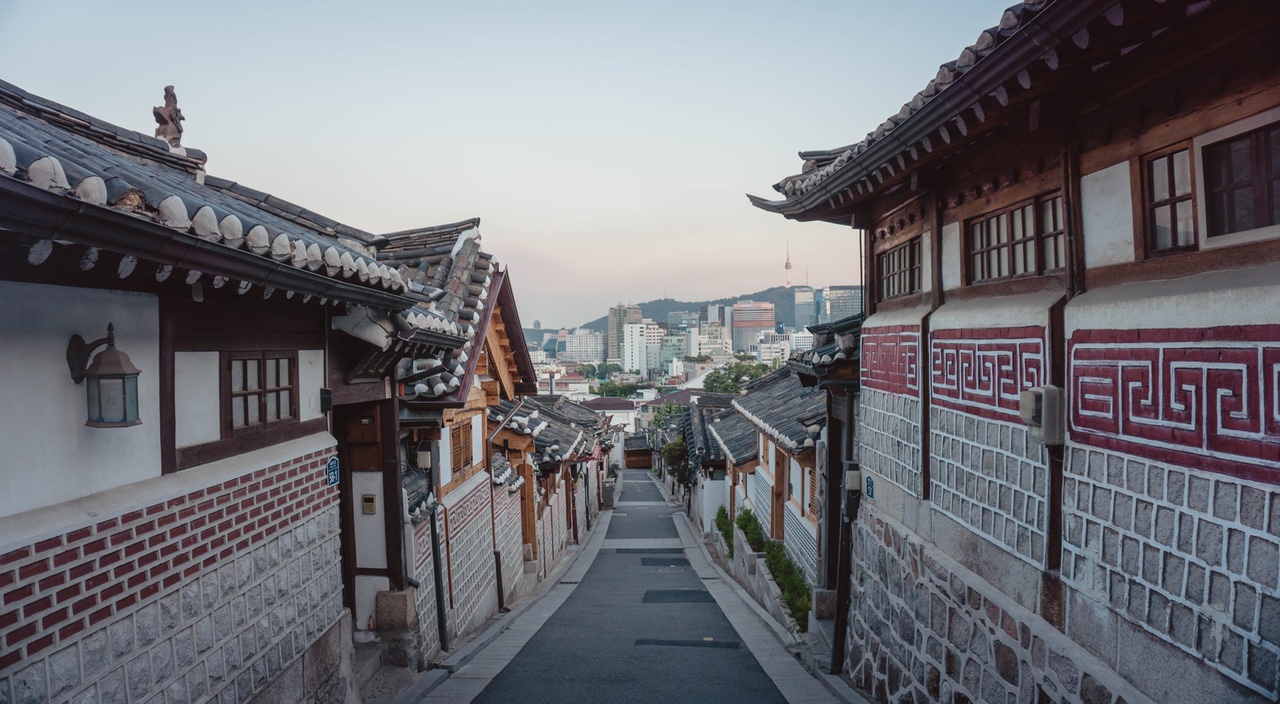}\hspace{\himg}
\includegraphics[width=\h]{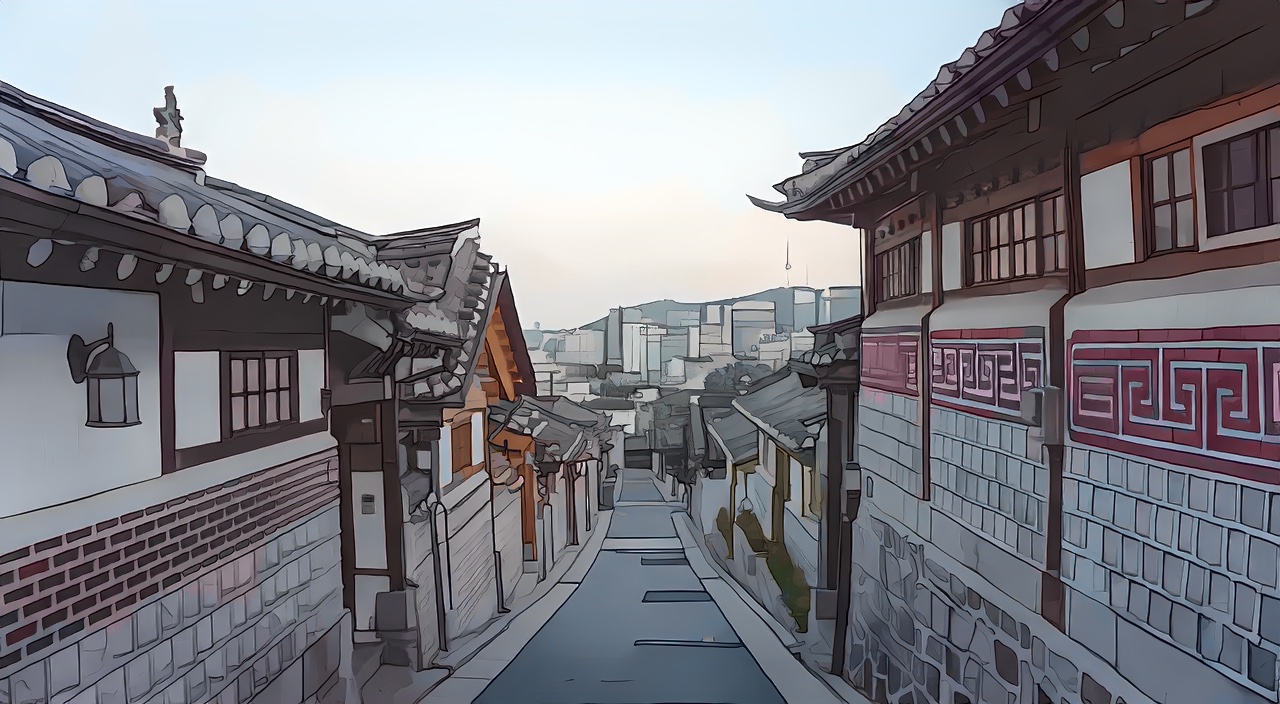}\hfill
\\
\makebox[\h][c]{(c) SD~\cite{rombach2022high} + SDEdit~\cite{meng2021sdedit} ($s=0.6$)}\hspace{\himg}
\makebox[\h][c]{(d) SD~\cite{rombach2022high} + SDEdit~\cite{meng2021sdedit} ($s=0.4$)}\hfill
\\
\includegraphics[width=\h]{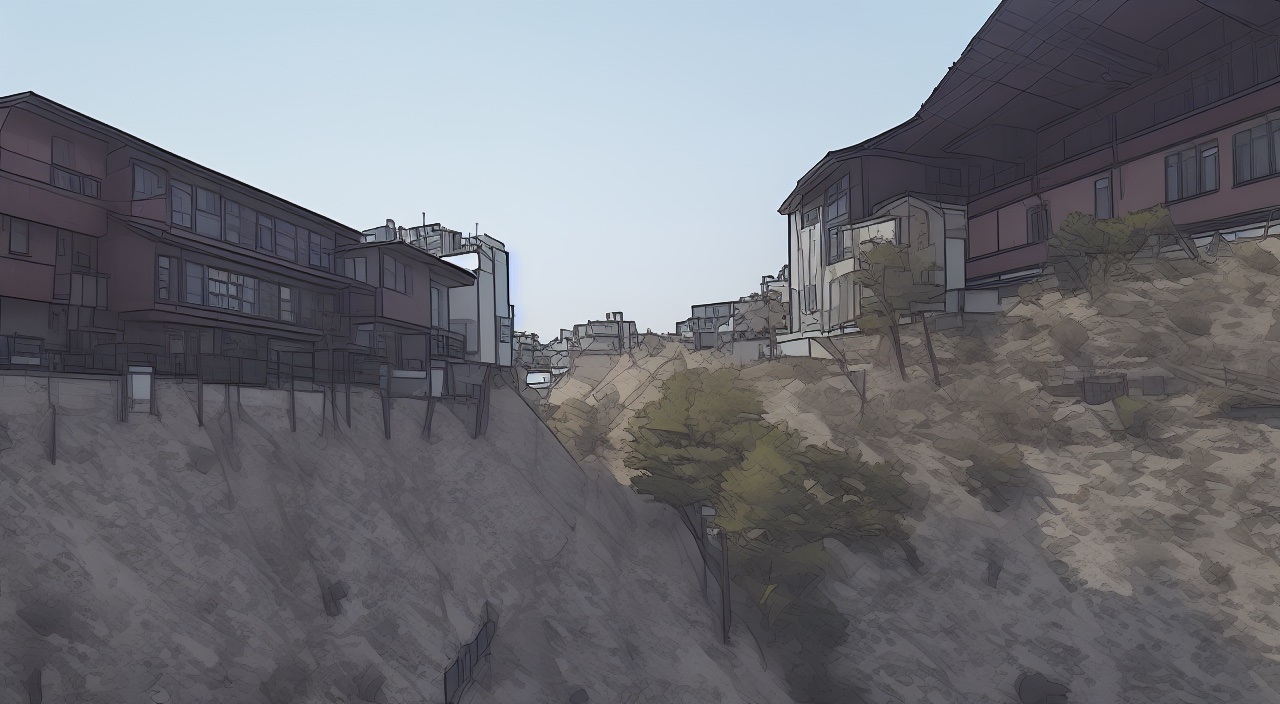}\hspace{\himg}
\includegraphics[width=\h]{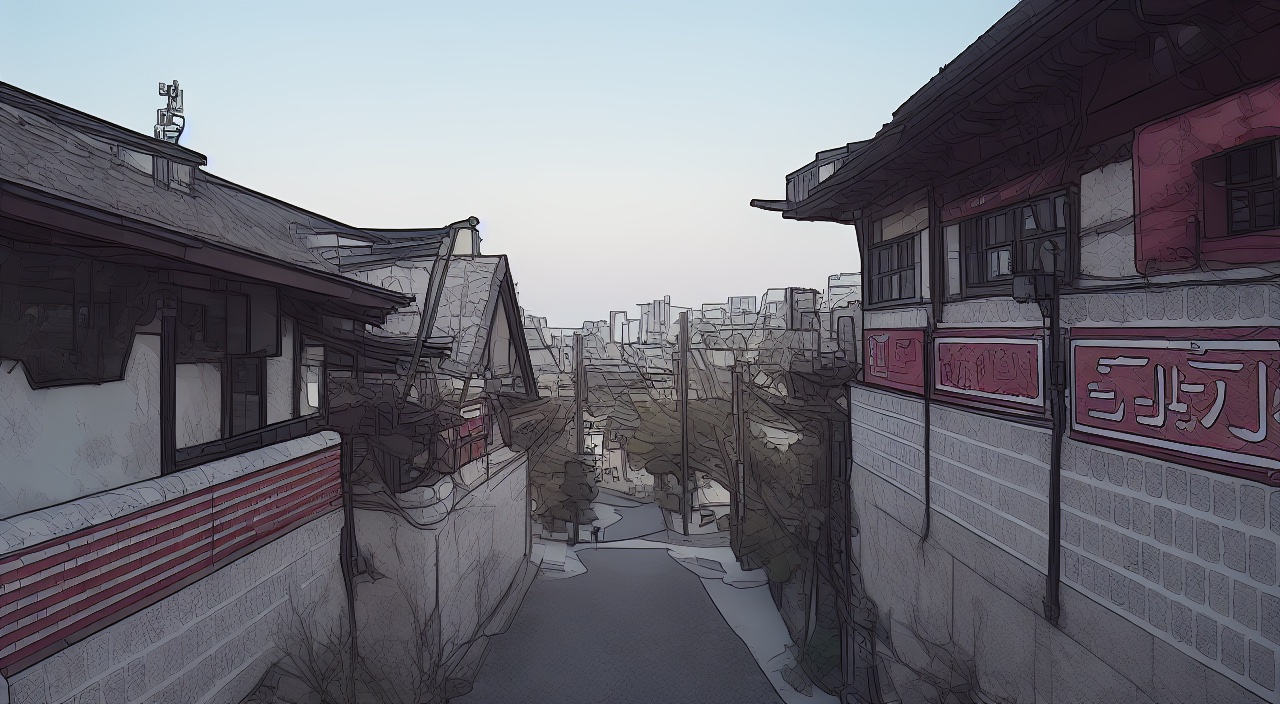}\hfill
\caption{\textbf{Comparison to Stable diffusion (SD)~\cite{rombach2022high} based image-to-image translation method~\cite{meng2021sdedit}}. Best viewed in zoom.}
\label{fig:vs_sd}
\end{figure*}
}

%% file: sections/0_abstract.tex
\begin{abstract}
Cartoonization is a task that renders natural photos into cartoon styles. Previous deep cartoonization methods only have focused on end-to-end translation, which may hinder editability.
Instead, we propose a novel solution with editing features of texture and color based on the cartoon creation process.
To do that, we design a model architecture to have separate decoders, texture and color, to decouple these attributes.
In the texture decoder, we propose a texture controller, which enables a user to control stroke style and abstraction to generate diverse cartoon textures.
We also introduce an HSV color augmentation to induce the networks to generate diverse and controllable color translation.
To the best of our knowledge, our work is the first deep approach to control the cartoonization at inference while showing profound quality improvement over to baselines.
\end{abstract}

%% file: sections/1_introduction.tex
\section{Introduction}
Cartoons gain steep popularity in a recent, and the number of cartoon creators have also increased.
The universal workflow of cartoon painters is as follows:
character drawing, which is then composed into a background scene. Post-processing such as shading is added afterward.
Professional tools~\cite{clipstudio,adobeae} provide helpful plugins to assist the artist. Despite this, cartoon creation still remains an arduous task even for the more skilled creators.

We follow the observation that cartoon-styled scene generation has received notable attention.
Many artists convert real-world photographs into cartoon styles to utilize as a background scene, dubbed as \textit{image cartoonization}.
This allows creators to more focus on effective decisions in making cartoons, such as character generation.
It is shown that deep learning-based cartoonization approach is able to produce cartoon-stylized output with a prominent quality, that is possible to be utilized in real service production~\cite{chen2018cartoongan,chen2019animegan,wang2020learning}.

\figBackgroundProcess

However, the previous deep methods skip the intermediate procedures of cartoon-making processes, thus disabling the creators from controlling outputs.
The artists follow a series of structured steps when creating a cartoon background from a photo (\Fref{fig:background_process}).
\textbf{1)} Color stylization, where the author changes the color both locally and globally.
Sky synthesis is performed along with this procedure.
\textbf{2)} Texture stylization, where additional sketch lines are drawn, and fine details are selectively removed to achieve the different \textit{levels of abstraction}. \textbf{3)} Post-processing, which includes lighting and image filters.
Unfortunately, due to the end-to-end inference nature of the previous deep cartoonization methods, the artist has no control over the generation process.
The creators may only intervene with a source photo (\Fref{fig:background_process}a) or the final output (\Fref{fig:background_process}e), which harms the usability of the cartoonization methods in artists' workflow.

In this study, we present an effective approach to embedding interactivity in cartoonization.
The proposed solution focuses on building a pipeline for more controllable texture and color. 
We define texture control as the manipulation of stroke thickness and abstractions.
This concept can be utilized in many scenarios; the artist can abstract the details of the far-distance scene to depict the natural perspective or emphasize the details of the character. 
The creators can also choose to change the delicacy of the brushstroke to match the texture of the foreground objects when composing the scene.
As for color control, we aim to build a control system in which the creator freely manipulates arbitrary regions with the desired color.
This is designed to assist the artist in the color stylization procedure (\Fref{fig:background_process}c).

To obtain user controllability in cartoonization, we separately build texture and color decoders to minimize interference across the features (\Fref{fig:model_overview}).
We also found that the decomposed architecture provides a robust and superb quality of texture stylization.
For texture control, we investigated the role of the receptive field and the target image resolution in the level of stroke thickness and abstraction.
Based on these observations, we present a \textit{texture controller}, which adjusts the receptive field of the network through a dynamic replacement of the intermediate features.
For color control, we jointly train the color decoder in a supervised manner with the paired dataset that is built based on the proposed HSV augmentation.
Throughout this training strategy, the color module gains the ability to produce diverse colors.
With the combination of the decoupled texture and color modules, we achieve a two-dimension of control space that can create a variety of cartoonized results upon user communication.
Such a design also provides robust and perceptually high-quality cartoonized outcomes.

To the best of our knowledge, our framework is the first approach that presents interactivity to deep learning-based cartoonization.
Based on the proposed solution, we demonstrate application scenarios that permit user intentions to create cartoonized images with diverse settings.
Extensive experiments demonstrate that the proposed solution outperforms the previous cartoonization methods in terms of perceptual quality, while also being able to generate multiple images based on the user's choices of texture and color.

\figModelOverview

%% file: sections/2_related_work.tex
\section{Related work}
\noindent\textbf{Non-photorealistic rendering.}
Non-photorealistic rendering (NPR) is a computer vision/graphics task that focuses on representing diverse expressive styles for digital art.
Among a variety of subtasks of NPR, we briefly describe the methods that stylize natural domain image to the specific artistic style.
Because of its usefulness in digital art creation, NPR has been expanded into various applicable scenarios such as line drawing~\cite{winnemoller2012xdog}, image abstraction~\cite{xu2011image}, and cartoonization~\cite{wang2004video}.
Style transfer methods~\cite{gatys2016image,johnson2016perceptual} are notable approaches in NPR.
By jointly optimizing the content and style losses, they can generate decent-quality of stylization.

\smallskip

\noindent\textbf{Cartoonization.}
Deep learning-based methods show profound improvement over conventional NPR algorithms on this task.
CartoonGAN~\cite{chen2018cartoongan}, a pioneering study on deep cartoonization, adopts adversarial training~\cite{goodfellow2014generative} along with an edge-promoting loss to improve cartoon style. 
AnimeGAN~\cite{chen2019animegan} enhances CartoonGAN with advanced losses suitable for cartoon style such as Gram-based loss~\cite{gatys2016image}.
With a careful inspection of cartoon drawing process, WhiteboxGAN~\cite{wang2020learning} decomposed cartoon images as surface, structure, and texture representations to tackle each factor with tailored losses. This approach achieves superior cartoonization quality compared to previous methods.

Despite the imposing cartoonization results, none of the deep cartoonization methods support interaction, failing to create diverse conditional outputs. Our method aims to enable user control while maintaining perceptually appealing cartoonization, moving closer to the actual service level.

%% file: sections/3_method.tex
\figResultPreAnalysis

\section{Method}
We describe the interactive cartoonization method dubbed as \textsc{Cartooner}.
It uses separate decoders for texture and color (\Fref{fig:model_overview}), contrary to the single decoder architecture of the previous methods.
The decision was made by observing professional artists' workflow, where they separate color modification from texture editing.
We further inspect that isolated modeling of texture and color produces reliable and high-quality cartoonized results.

The controllable features are defined as texture level vector $\pmb{\alpha}$ and users' color modification $\pmb{c}$.
Given a photo $I_{src}$, the goal is to generate a cartoonized image $\hat{I}_{tgt}$ that follows the user intention $\pmb{\alpha}$ and $\pmb{c}$.
To achieve this, \textsc{Cartooner} encodes an image to the latent feature through $E_{shared}$, then delivers it to the separate decoders, $S_{texture}$ and $S_{color}$.
Note that we use Lab color space instead of RGB, hence the texture module produces an L-channel texture map, while the color module generates an ab-channel color map.
These outputs are finalized by converting back to RGB space.

\subsection{Texture module}
\label{subsec:texture_module}
\noindent\textbf{Analysis of texture level.}
In this module, the primary goal is to provide a fine control mechanism and we define texture control as altering \textit{stroke thickness} and image \textit{abstraction}.
To do that, we first analyze which components influence stroke thickness and abstraction change.
In our preliminary experiments, we observed that increasing the resolution of target cartoon images affects stroke thickness, and expanding the receptive field (RF) of the generator along with the increased resolution inflates the abstraction level (\Fref{fig:result_preanalysis}).

\figPreAnalysis

For stroke thickness change, we argue that the \textit{loss network with a fixed RF} (\textit{e.g.,} VGG or discriminator) is involved as shown in \Fref{fig:preanalysis}a.
Given the fixed loss network, when we increase the resolution of cartoon images (\textcolor{ForestGreen}{green box}), the strokes are enlarged within an RF window, thus inevitably, the generator learns to produce thick strokes at training.
When we decrease the resolution (\red{red box}), the opposite behavior occurs.
This impacts cartoonization more since cartoon images mostly have flat texture regions.

For abstraction change, we argue that \textit{scene complexity} affects this as shown in \Fref{fig:preanalysis}b.
When the RF of the generator is expanded (\textcolor{blue}{blue box}), the network can perceive a wider region of a content image, which results in high scene complexity.
In contrast, when the resolution of a cartoon image grows, its scene complexity becomes lower since the loss network can only see relatively tiny regions (\textcolor{ForestGreen}{green box}).
With these, if we utilize high-resolution cartoon images $I^{HR}_{tgt}$ to train the generator with a large RF (which are \textcolor{ForestGreen}{green} and \textcolor{blue}{blue box}, respectively), the generator is guided to reduce the complexity of the high-complexity scene.
This arises from the loss calculation with the low-complexity scene extracted from the loss network.
As a result, the generator with a large RF gains the ability to ``abstract" the complex details.
For the lower RF (\textcolor{magenta}{pink window}), the contrary behavior happens.
However, the abstraction change is not as dramatic as the high one since in general, the scene complexity of cartoon is lower than the content images.
We also analyzed the scenario where \textit{only RF of the generator is expanded}, however, the results are not drastic as \Fref{fig:result_preanalysis}b since the generator is not guided by the different-complexity cartoon scene.
Note that Jing et al.~\cite{jing2018stroke} inspected the role of the resolution and RF in style transfer literature, however, our in-depth analysis reveals that their behavior patterns are disparate in cartoonization.

\figTextureController

\smallskip
\noindent\textbf{Texture controller.}
The above analysis requires multiple networks to handle diverse levels and it cannot produce consistent styles for each other.
Also, it only supports discrete control levels, making it challenging to be used as a real-world solution.
Hence, based on the analysis, we introduce a simple but effective texture control module, dubbed as \textit{texture controller} (\Fref{fig:texture_controller}).
It consists of the stroke and abstraction control units and we design both units as to be a multi-branch architecture.
In the stroke unit, each branch is composed of two consecutive 3$\times$3 conv layers, and these are fused by the gating module.
The abstraction unit is identical in structure to the stroke unit except it uses conv layers with large kernel size, $K_1 < K_2 < ... < K_N$.

The texture controller is influenced by texture level $\pmb{\alpha}=\{\alpha_s, \alpha_a\}$, specifically, the stroke and abstraction units are guided by stroke thickness $\alpha_s$ and abstraction $\alpha_a$ levels, respectively.
With the feature $f$ from the encoder, the stroke unit generates a feature set $\pmb{g}_{s} = \{g_s^1,...,g_s^N\}$ through conv branches, and the abstraction unit produces $\pmb{g}_{a}$ as the same way.
Then, according to texture levels $\alpha_{\{s,a\}}$, which are a positive rational number, the two features of $\pmb{g}_{\{s, a\}}$ with indices closest to a texture level are chosen.
The chosen features are then interpolated based on the respective distance between $\alpha_{\{s,a\}}$ and indices.
Finally, these are combined by an element-wise addition operation.

We design the stroke control unit to have all 3$\times$3 conv layers since the texture level analysis showed that RF of the generator does not affect the stroke thickness.
Instead, each branch is trained by different resolutions of target cartoon images.
At inference, the feature interpolation via $\alpha_s$ enables continuous control over stroke thickness.
For the abstraction unit, we also construct a single module based on the analysis.
However, unlike the stroke unit, each branch includes conv layers with different kernel sizes (with increasing order) because changing both the RF of the generator and the resolution of target images alters the abstraction.
The output features are interpolated through $\alpha_a$, as identical to the stroke unit.
As we design the decoupled structure of stroke and abstraction in parallel, each unit can concentrate on a different aspect and it provides the ability that can control the texture as a two-dimensional space.

In addition, to incorporate adversarial learning~\cite{goodfellow2014generative} into texture control, we utilize a multi-texture discriminator.
It is based on the multi-task discriminator~\cite{mescheder2018training,choi2020stargan}, which consists of multiple output branches.
Each branch corresponds to a different texture level and learns to distinguish whether a given image is from a real cartoon domain or generated.

\figPreprocessTrain
\figCompQual

\smallskip
\noindent\textbf{Loss function.}
We use adversarial loss $L^{adv}_{texture}$ to guide the model mimicking the texture of the target cartoon.
\begin{equation}
L^{adv}_{texture} = logD_{\pmb{\alpha}}(I^{L, \pmb{\alpha}}_{tgt}) + log(1-D_{\pmb{\alpha}}(G(I^{Lab}_{src}, \pmb{\alpha})))
\end{equation}
where $G$ is the generator and $D_{\pmb{\alpha}}$ denotes the multi-texture discriminator with a given texture factors $\pmb{\alpha}$. $I^{L, \pmb{\alpha}}_{tgt}$ is a resized target cartoon image to fit a respective texture level $\pmb{\alpha}$.
To ensure a cartoonized output well preserves the semantic information of a source photo, we employ content loss.
\begin{equation}
L^{vgg}_{content} = ||VGG(I^{L}_{src}) - VGG(G(I^{Lab}_{src}, \pmb{\alpha}))||_1.
\end{equation}
We use a \textit{conv4\_4} layer of the pre-trained VGG19~\cite{simonyan2014very}.
In addition, we enforce the generator to learn high-level texture representation via Gram-based loss as:
\begin{equation}
L^{vgg}_{texture} = ||Gram(I^{L, \pmb{\alpha}}_{tgt}) - Gram(G(I^{Lab}_{src}, \pmb{\alpha}))||_1.
\end{equation}
$Gram$ indicates the Gram calculation with VGG feature extraction (of \textit{conv4\_4}).
We also use total variation loss~\cite{aly2005image} to impose spatial smoothness on the output.
\begin{equation}
L^{tv}_{texture} = || \nabla x (G(I^{Lab}_{src}, \pmb{\alpha})) + \nabla y (G(I^{Lab}_{src}, \pmb{\alpha})) ||_1
\end{equation}
With balancing parameters $\lambda_{texture}^{1,...,4}$, the final loss of the texture decoder (and the shared encoder) is defined as:
\begin{align}
L_{texture} &= \lambda_{texture}^1 * L^{adv}_{texture} + \lambda_{texture}^2 * L^{vgg}_{content} \nonumber\\
&+ \lambda_{texture}^3 * L^{vgg}_{texture} + \lambda_{texture}^4 * L^{tv}_{texture}.
\end{align}

\subsection{Color module}
\label{subsec:color_module}
The goal is to transfer the color of a given source photo to a provided color intention while reflecting the color \textit{nuance} of the target cartoon.
\textsc{Cartooner} takes an input photo $I^{Lab}_{src}$ as well as an input color map $\bar{C}^{Lab}_{src}$ and generates an ab-channel image $\hat{I}^{ab}_{src}$, which is later concatenated with the texture map $\hat{I}^{L}_{tgt}$ generated from the texture decoder.
To simulate control manipulation, we synthetically generate a color map, $\bar{C}^{Lab}_{src}$ (\Fref{fig:preprocess_train}).
Given an input photo $I^{RGB}_{src}$, we create an initial color map $C^{RGB}_{src}$ by applying a superpixel algorithm.
Without this, fine details of an input image become too noisy and thus not adequate to be utilized as a color cue; a superpixel is used as a noise reduction procedure.
Then, the HSV augmentation changes the color of $I^{RGB}_{src}$ and $C^{RGB}_{src}$, creating color manipulated images $\bar{I}^{RGB}_{src}$ and $\bar{C}^{RGB}_{src}$.
These are converted to Lab space.
Note that we observed that color transfer to either input or output image, unlike ours, cannot achieve faithful visual quality.

HSV augmentation is a simple but effective method that can reflect diverse color control intentions from a user.
It randomly alters all color channels of HSV; hue, saturation, and value (brightness).
To prevent the color shifting from generating perceptually implausible outputs, we further apply the L caching trick~\cite{cho2017palettenet} prior to the color augmentation, which caches the luminance (L) of image and reverts the luminance of the augmented image to a cached one.
We cache L instead of V-channel since V indirectly interferes with L, which is important in regard to diverse cartoonization.

\tableCompQuant

\smallskip
\noindent\textbf{Loss function.}
We use a simple mean squared error-based reconstruction loss as shown below.
\begin{equation}
L_{color} = ||\bar{I}^{ab}_{src} - G(I^{Lab}_{src}, \; \bar{C}^{Lab}_{src})||_2
\end{equation}
In our experiment, additional adversarial loss or regularization shows marginal improvement in color quality.
We suspect that the decomposed color modeling, as well as the color cue ($\bar{C}^{Lab}_{src}$) provision, ease the training difficulty.

\smallskip
\noindent\textbf{Reflecting target cartoon's color.}
We designed to preserve the color information of an input image to increase the controllability, however, one might want to generate an image that has a similar color distribution to the target cartoon.
To handle this scenario, we additionally fine-tune the color decoder with an assist of adversarial loss as in below.
\begin{align}
L^{tgt}_{color} &= \lambda^1_{color} * L_{color} + \lambda^2_{color} * L^{adv}_{color} \;\;\;\; \text{where,}\nonumber\\
L^{adv}_{color} &= logD(I^{ab}_{tgt}) + log(1-D(G(I^{Lab}_{src}, \; C^{Lab}_{src})))
\end{align}
Note that we use a color cue that is generated from an original image ($C^{Lab}_{src}$), instead of $\bar{C}^{Lab}_{src}$.
With the aforementioned color decoder parts, a user can choose which ``color mode" to use interchangeably depending on the situation.

\subsection{Model training}
\label{subsec:training}
Unlike previous deep cartoonization methods, we do not perform network warm-up~\cite{chen2018cartoongan}.
We train the entire framework with a loss of $L = L_{texture} + L_{color}$, except for the abstraction control unit.
Then, the abstraction unit is trained (via $L_{texture}$) while other components are all frozen.
To provide various resolution images to the generator, we resize $I^{RGB}_{tgt}$ according to texture level $\pmb{\alpha}$ (\Fref{fig:preprocess_train}).
We set kernel sizes of the abstraction unit, $\{K_1, K_2,...,K_N\}$, as $\{3, 7, 11, 15, 19\}$ each.
When training \textsc{Cartooner}, we randomly choose $\alpha_{\{s, a\}} \in \{1,...,5\}$, which respectively resize $I^{RGB}_{tgt}$ to be $\{256^2, 320^2, 416^2, 544^2, 800^2\}$ resolutions, but $\alpha_{\{s, a\}}$ can be expanded to arbitrary numbers at inference.
More detailed setups are described in Suppl.

%% file: sections/4_experiment.tex
\section{Experiment}
\smallskip
\noindent\textbf{Baselines.}
We compare \textsc{Cartooner} with the state-of-the-art deep learning-based cartoonization methods, CartoonGAN~\cite{chen2018cartoongan}, AnimeGANv2~\cite{chen2019animegan}, and WhiteboxGAN~\cite{wang2020learning}.
Since they have trained their cartoonization network on different datasets and setups from each other, we retrained using our cartoon datasets using official codes.

\smallskip
\noindent\textbf{Datasets.}
We built datasets focused on landscape, to better target the domain of cartoonizing background scenes.
We used \textit{monet2photo}~\cite{zhu2017unpaired} as the photo domain.
We collected cartoon datasets from Japanese animations and Webtoons.
Specifically, we acquired artworks by Miyazaki Hayao and Shinkai Makoto, and comics of titles ``FreeDraw" and ``Barkhan" from the NAVER Webtoon platform.
Detailed dataset generation protocols are described in Suppl.

\smallskip
\noindent\textbf{Metrics.}
We evaluated the cartoonization with Fre\'chet Inception Distance (FID)~\cite{heusel2017gans} and $\text{FID}_{\text{CLIP}}$~\cite{kynkaanniemi2022role}.
We additionally conducted a user study to measure perceptual quality.
We asked 26 users to select the best results for how well the outputs follow both the cartoon styles and source photos.

\figInteractivity

\subsection{Comparison with state-of-the-art method}
When comparing \textsc{Cartooner} with others, we generate images to reflect the target cartoon since FIDs can be influenced by color information.
In addition, we set the texture levels $(\alpha_s, \alpha_a)$ as zero, which is identical stylization setting to others.
\Tref{table:comp_quant} shows the quantitative comparison.
\textsc{Cartooner} achieves exceeding performance on both FID and $\text{FID}_{\text{CLIP}}$ with significant margins for all the cases.
We also present the visual comparison in \Fref{fig:comp_qual}.
Separation of the texture and color decoders helps prevent image artifacts, for instance, \textsc{Cartooner} produces fewer color bleeding (\Fref{fig:comp_qual}, 2nd row).
The visual quality is also profoundly enhanced and \textsc{Cartooner} can capture adequate stroke and color nuance of the target cartoon.
A user study shows the superiority of \textsc{Cartooner} as well (\Tref{table:user_study}).

\subsection{Interactivity}
\label{sec:interactivity}
As shown in \Fref{fig:interactivity}, \textsc{Cartooner} creates diverse results according to user interaction.
When the artist manipulates the colors to their tastes (with any color adjustment UI), \textsc{Cartooner} automatically reflects the intention.
They can also edit textural details by simply controlling the stroke or abstraction factors to match the output in various cartoon situations.
These editings can be performed locally or globally through a simple mask-based region control UI (shown in Suppl.).
Our cartoonization workflow is more compact than the traditional editing tools, while still maintaining an adequate level of user intervention.
Although \textsc{Cartooner} may not achieve the degree of meticulous editing workflows (which requires the effort of skilled artists), it can provide a broader range of user experiences with suitable quality.
We would like to emphasize that none of the deep cartoonization methods can provide controllability nor produce diverse results of a given source photography.

\figAblColor
\figFeatmap

\subsection{Model analysis}
\noindent\textbf{Color module.}
In \Fref{fig:abl_color}, we present the result where the color change is performed before or after cartoonization, unlike ours that jointly models the color and texture.
The pre-execution of color change (\Fref{fig:abl_color}b) cannot adequately handle the delicate color alters and produces uneven texture level since the model has not observed the re-colorized input image at training, which becomes out-of-distribution.
The pipeline of color change after cartoonization (\Fref{fig:abl_color}c) cannot generate cartoon-style colors at all.

\smallskip
\noindent\textbf{What does texture controller learn?}
We visualize the output feature map of the texture controller in \Fref{fig:featmap}.
The stroke control unit produces features that more concentrate on the high-frequency edge regions, which empirically demonstrates why this can control the stroke thickness.
On the other hand, the abstraction control unit focuses on a wide range of regions including flat texture and some mid-frequency details.
Consequently, this unit can deliver helpful clues about the abstraction change to the decoder.

\smallskip
\noindent\textbf{Stroke control unit.}
We decrease the number of stroke levels at training and examine how the models react at inference.
\Fref{fig:abl_stroke} shows that the model trained with 2-levels cannot capture high stroke thickness; we observed that it produces saturated thickness only.
In contrast, the model with increased stroke levels adequately expresses a wide dynamic range of stroke thickness.
In Suppl., we demonstrate a similar analysis regards on the abstraction control unit.

\figAblStroke

%% file: sections/5_application.tex
\section{Application}
\smallskip
\noindent\textbf{Reference image-based color control.}
In \Sref{sec:interactivity}, we demonstrated a simple interactive cartoonization workflow with \textsc{Cartooner}.
However, unskilled users might struggle to choose appropriate color tones if they have little experience in coloring.
To increase usability for inexpert users, we present reference image-based color control (\Fref{fig:ref_based}).
Instead of direct color manipulation, a user prepares a color guidance image and chooses which regions to be referred via region masking UI.
After this, \textsc{Cartooner} transfers the color information of the selected area to the cartoonized outputs.
To implement this, we first extract a color palette from a reference image and then manipulate the color map, $C_{src}$ using palette-based color transfer algorithm~\cite{chang2015palette}.

\smallskip
\noindent\textbf{Semi-automatic cartoon making.}
As discussed, making background scenes is repetitive and time-consuming.
\textsc{Cartooner} can help to reduce the burden of background creation with interactive texture-color editing so the artists can focus more on other creative tasks.
\Fref{fig:cartoon_making} shows an example where one can use \textsc{Cartooner} to effectively create a cartoon cut, consisting of a background scene blended with character(s) and/or speech balloons, which would have been a strenuous task for previous pipelines.

\figRefBased

\figCartoonMaking

%% file: sections/6_conclusion.tex
\vspace{0.5em}
\section{Conclusion}
We proposed an interactive cartoonization model, \textsc{Cartooner}.
The proposed method accepts user-guided texture control in the form of abstraction and stroke strength levels, which are passed to a \textit{texture controller} to dynamically control the overall texture of the generated image.
The user can also manipulate the color scheme through a color module, which is reinforced by the HSV augmentation.
Experimental results demonstrate \textsc{Cartooner}'s superiority in both quality and usability as applications for cartoon creators.

Although we provided effective control space, there exist more controlling factors, especially for texture control (\Fref{fig:background_process}d).
In the future, it is worth exploring the other aspects of texture editing such as brush stroke's style~\cite{kotovenko2021rethinking}.

%% file: sections/7_appendix.tex
\section{Method details}
In this section, we explain detailed settings and experimental analyzes conducted in the main text.

\smallskip
\noindent\textbf{Texture control analysis.}
Here, we describe detailed setups of the texture control analysis (Section 3.1, main text).
In Figure 4a (main text), each stroke thickness case was generated by the separately trained \textsc{Cartooner}.
We used three models and these are trained with the cartoon image resolutions of $\{256^2, 416^2, 800^2\}$.
We also set the texture controller to only have a single branch.
With these setups, the models trained with $\{256^2, 416^2, 800^2\}$ resolutions generated thin, moderately thick, and thick strokes respectively.

To conduct an abstraction change experiment, as shown in Figure 4b (main text), we trained multiple \textsc{Cartooner} similar to the stroke change scenario except for the receptive field (RF) of the generator.
We differentiated the RF of the network by changing the kernel size of conv layers in the texture controller by $\{3, 11, 19\}$ each, which corresponds to the low, moderate, and high abstraction scenes.

\smallskip
\noindent\textbf{Network architecture.} \Tref{table:network_architecture} presents the network architecture of \textsc{Cartooner}.
We applied LeakyReLU for all conv layers and did not use any normalization layer.
In our experiments, we observed that the existence of a normalization layer (batchnorm~\cite{ioffe2015batch} and instance norm~\cite{ulyanov2016instance}) drops the cartoonization quality.
The cardinality of conv layers in the ResNeXt blocks were set to 32.
We used the bilinear interpolation method for \textit{Upsample} layer.
In \textit{col2} and \textit{col3} layers in the color decoder, the additional 3-channel of each first conv layer is for the color cue injection.

\smallskip
\noindent\textbf{Abstraction control unit.}
We designed this unit to be a shared multi-branch system to increase the quality robustness and reduce the model size.
Specifically, the multi-branch module was composed of conv layers with varying kernel size, $K_1 < ... < K_N$, where $K$ denotes kernel size and $N$ is the number of branches.
Instead of using $N$ conv kernels, we only initialized a single conv layer with $K_N$ kernel size.
All other conv kernels were set to be a subset of $K_N$ kernel as illustrated in \Fref{fig:shared_unit}.
By doing so, the abstraction control unit can produce robust outcomes for the different abstraction levels.
We will demonstrate how crucial this design is in \Sref{sec:suppl_model_analysis}.
Such a design also successfully reduces the model parameters; without a shared scheme, the model parameters of \textsc{Cartooner} becomes 26.5M, while the shared kernel version (ours) is 5.9M.

\tableNetworkArchitecture

\smallskip
\noindent\textbf{Model training.}
We trained \textsc{Cartooner} using Adam~\cite{kingma2014adam} for 100K steps with a batch size of 32 and learning rate of $2\times10^{-4}$.
Unlike previous deep methods~\cite{chen2018cartoongan,chen2019animegan,wang2020learning}, we did not perform network pre-training~\cite{chen2018cartoongan}.
For all results shown in this paper, we used the same hyper-parameters: $\lambda^1_{texture}=$ 1.0, $\lambda^2_{texture}=$ 0.0025, $\lambda^3_{texture}=$ 0.0045, $\lambda^4_{texture}=$ 0.0015.
We utilized the official Caffe-version of VGG19~\cite{simonyan2014very}, and when creating a Gram matrix, we divided it by the product of \# of the channels, width, and height.

When generating an initial color map, $C^{RGB}_{src}$ from an input photo $I^{RGB}_{src}$, we adopted \textit{Zhu et al.}~\cite{zhu2021learning} for both training and inference.
We selected this algorithm since it is GPU-friendly, nevertheless, any off-the-shelf super-pixel algorithm can be adopted.
We tested other algorithms, such as SLIC~\cite{achanta2012slic}, and observed no performance drop even when we use a different approach at train and inference.

\figSharedUnit

\section{Experimental settings}
\noindent\textbf{Dataset.}
In our study, we mainly focused on outdoor scenes and landscapes, to better target the domain of cartoonizing background scenes.
We collected 8,227 real-world outdoor images from the \textit{monet2photo} dataset~\cite{zhu2017unpaired} for the source photo domain.
Then, this was split into 6,227 and 2,000 images for the train and test set. 
For the target cartoon domain, we collected cartoon datasets from Japanese animations and Webtoons.
Specifically, we acquired animation images from `The Garden of Words' and `Your Name' by Shinkai Makato, and `Spirited Away' by Miyazaki Hayao.
For the Webtoon dataset, we collected comics of titles `FreeDraw'\footnote{\url{https://comic.naver.com/webtoon/list?titleId=597447}} and `Barkhan'\footnote{\url{https://comic.naver.com/webtoon/list?titleId=650305}} from the NAVER Webtoon platform.

We resized source domain images as 256$\times$256 resolution.
For cartoon datasets, we cropped images to be a resolution of 512$\times$512 and applied $\times$2 super-resolution~\cite{wang2018esrgan} beforehand when the raw image is $<$1024$\times$1024 resolution.
In total, we gathered images of 5,480 Hayao, 5,647 Shinkai, 5,308 FreeDraw, and 6,186 Barkhan datasets.
In our experiment, we treated the above four style datasets as independent since each artwork has a unique style.

\smallskip
\noindent\textbf{User study.}
We asked 26 participants to pick the best results for how well the outputs follow both the cartoon styles and source photos.
Each of them was asked to vote on 16 questions, thus we collected 416 samples in total.
In every question, we presented source photography, exemplar cartoon image, and the results of the previous and our methods.
For better visibility, we also showed cropped patches for all the results.
We computed the \textit{quality preference} score by the ratio of the voted (as the best) cases.

\figAblPretrain
\figResultPreAnalysisSuppl

\smallskip
\noindent\textbf{Interactive UI.}
\Fref{fig:ui} shows an interactive UI of \textsc{Cartooner}.
In the left panel, selection tools (\textit{e.g.,} selection, quick selection, and eraser) offer mask-based region selection so the user can easily manipulate local region.
In the right panel, style change tools offer texture and color control over the selected region.
Here, the creators can change the \textbf{1)} target cartoon style, \textbf{2)} texture (\textit{i.e.,} stroke thickness and abstraction), and \textbf{3)} color of the cartoonized outcomes.
For the color control, we provide both a color picker and an HSV control slider UIs since we observed that the latter is straightforward to utilize for unskilled users.
Throughout this, the creators easily render given natural photos into the cartoon styles as well as manipulate the results with their own desired texture and color.

\figAblAbsShared

\smallskip
\noindent\textbf{Naive coloring approach.}
In \Fref{fig:abl_color}, we presented a comparison between \texttt{Cartooner} and two alternative approaches that accept color control.
The first one involves executing the re-colorization before the cartoonization (\Fref{fig:abl_color}a), while the second one involves the re-colorization after the cartoonization (\Fref{fig:abl_color}b).
For both, we employed a palette-based re-colorization~\cite{chang2015palette} and \texttt{Cartooner} that is trained without a color decoder.

\smallskip
\noindent\textbf{Reference image-based color control.}
It can be achieved with a simple pipeline.
With a given reference image, we extract the color palette through the K-means clustering.
In our demonstration (Figure 13, main text), we used eight palettes (\textit{i.e.,} clusters).
Note that when the user selects the specific region (via selection tools) to transfer the color, we generate a color palette from that region instead.
Then, the user chooses the region to be changed in a source photo and the framework calculates the average color of the region, denoted as $c$.
In the meantime, the user also picks the color $c'$ from the palette.
Using these colors, $c$ and $c'$, the framework performs a palette-based color transfer algorithm~\cite{chang2015palette} to the initial color map $C_{src}$ and generates $\bar{C}_{src}$.
However, we observed that a simple color transfer in RGB space (\Eref{eq:color_transfer}) also produces robust results.
With the manipulated color map, \textsc{Cartooner} now generates appropriate cartoonized outputs that fulfill the users' color intention.
\begin{equation}
\label{eq:color_transfer}
\bar{C}_{src}^{RGB} = C_{src}^{RGB} + (c' - c)
\end{equation}

\section{Model analysis}
\label{sec:suppl_model_analysis}
\noindent\textbf{Model training.}
We analyzed the pre-training that has been a prevalent strategy on deep cartoonization~\cite{chen2018cartoongan,chen2019animegan,wang2020learning}.
Previous studies reported that the warm-up process, which optimizes the network through the content loss only in advance, guarantees better convergence and cartoonization quality.
However, we found that the network pre-training does not require to \textsc{Cartooner} (\Fref{fig:abl_pretrain}).
We claim that a separate design of texture and color enables robust training since the texture and color decoders can solely concentrate on synthesizing texture and color alone, respectively.

\figAblAbsKsize
\figUI

\smallskip
\noindent\textbf{What affects the abstraction?}
In \Fref{fig:result_preanalysis_suppl}, we present results of the abstraction analysis where we only change the receptive field (RF) of the generator (\Fref{fig:result_preanalysis_suppl}a), or change both RF of the generator and the image resolution of cartoon domain dataset (\Fref{fig:result_preanalysis_suppl}b).
Note that the latter result was shown in the main text.
When we alter the RF of the generator alone, the abstraction seems not much affected since the low-complexity cartoon scene does not guide the generator, hence the generator would not have any incentive to increase the abstraction.
On the other hand, as we discussed in Figure 5 (main text), increasing both RF and the resolution effectively affects the abstraction due to the scene complexity guidance from the cartoon images.

\smallskip
\noindent\textbf{Abstraction control unit.}
We designed this unit to have a shared conv kernel scheme in a multi-branch system.
When the conv kernels are not shared (\Fref{fig:abl_abs_shared}a), it is not guaranteed consistent and smooth abstraction transitions.
For example, the detailed textural gradations near window edges intensify even when we increase the abstraction (2nd column).
This is because each branch learns separate representations without communication, thus, they are not tuned to each other to generate smooth abstraction change.
In contrast, the shared conv kernel approach (\Fref{fig:abl_abs_shared}b) adequately produces continuous abstraction modification.
We claim that robustness can be achieved since the RF of the current abstraction is gradually evolved based on the previous RF (\Fref{fig:shared_unit}).
By doing so, all the branches share the viewpoint near the center point (of RF), and the kernels with larger RF look wider region while maintaining the perspective of the previous ones.
As a consequence, they produce consistent and gradual transitions on the abstraction.

We demonstrate the results of the different kernel size cases.
In \Fref{fig:abl_abs_ksize}a, we excessively expand the kernel sizes and the network generates blurred outputs since the model is guided from too many flat regions.
In our experiment, we found that kernel size in \Fref{fig:abl_abs_ksize}b shows the best abstraction change in terms of perceptual quality.
However, the other settings (such as \{3, 5, 9, 11, 13\}) also make plausible outcomes as long as kernel sizes are in increasing order.

\section{Discussion}
\figVsSD

\smallskip
\noindent\textbf{Comparison to Jing et al.~\cite{jing2018stroke}.}
Unlike style transfer, which utilizes general artistic paintings, expressing abstraction is crucial in cartoonization since cartoon scenes have many flat regions.
Thus, we decomposed stroke into \textit{stroke thickness} and \textit{abstraction}.
For stroke thickness, our observations are similar to the stroke size analysis in Jing et al.~\cite{jing2018stroke}.
However, we argue that the distinctiveness of the cartoon, characterized by its numerous flat regions, renders this distinctive more conspicuous than that of style transfer.
For abstraction, we argue that our analysis based on \textit{scene complexity} is more relevant to the cartoon domain, given the prevalence of flat regions and abstraction in this genre.

\smallskip
\noindent\textbf{Comparison to WhiteBoxGAN~\cite{wang2020learning}.}
We present a comparison between our proposed method and WhiteBoxGAN~\cite{wang2020learning}.
The shared objective of both methods is to decouple features in order to improve the training and synthesize quality.
WhiteBoxGAN achieves this by decomposing representations in its loss design such as utilizing structure, texture, and surface losses.
In contrast, \texttt{Cartooner} separates representations in the model design such as texture and color decoders.
Upon comparing the ``decomposition" in loss and model design, we posit that the latter approach provides a more explicit separation of features, as each decoder can concentrate on its specific task (texture or colormap generation).
In contrast, when multiple losses are incorporated in a single decoder (as WhiteBoxGAN), the decoder may become confused due to the diverse signals emanating from multiple synthesis tasks.
As a result, our framework offers the significant advantage of being more efficient during the training process, leading to superior quality with fewer artifacts, as demonstrated in \Fref{fig:comp_qual}.
Moreover, our model design philosophy is universally applicable, as the texture and color decoders are designed with identical structures that can be integrated into any network architecture.

\noindent\textbf{Comparison to Stable diffusion~\cite{rombach2022high}.}
We tackled image-to-image (I2I) based cartoonization.
It requires maintaining the source photos' structure while altering their style and mood.
In this respect, we compared the proposed method with \textit{unpaired I2I approach}~\cite{chen2018cartoongan,chen2019animegan,wang2020learning} since these faithfully meet the above requirements.
Nevertheless, it is worth noting the application of recent diffusion models~\cite{ho2020denoising} in the cartoonization task considering its unprecedented progress in the image synthesis task~\cite{nichol2021glide,dhariwal2021diffusion,rombach2022high}.
To investigate this, we fine-tuned a pre-trained Stable diffusion (SD)~\cite{rombach2022high} via Dreambooth strategy~\cite{ruiz2022dreambooth} and then incorporated the most popular I2I method, SDEdit~\cite{meng2021sdedit}, to make SD an I2I framework.
As shown in \Fref{fig:vs_sd}, SD with SDEdit struggles to produce satisfactory outcomes compared to \texttt{Cartooner}.
When the denoising strength ($s$) is set to a higher value ($=0.6$), it fails to preserve the source photo's content information, which hinders usability in the background-making process.
When $s$ is set to a lower value ($=0.4$), it produces severe artifacts, which results in inferior visual quality than ours.
We suspect this is due to SDEdit's inability to effectively eliminate high-frequency information when the denoising strength is low.
In the future, the SD-based content-preserving I2I approach would be a valuable research topic.

\smallskip
\noindent\textbf{Comparison to cartoon filters.}
The majority of cartoon filters in commercial software~\cite{photoshop,clipstudio} rely on traditional algorithms so they support very limited cartoon styles.
In this respect, ``deep cartoonization" studies have not included comparisons with filters since filters cannot express various styles that deep methods can produce.
In addition, through interviews with many artists, we found that commercial filters require significant retouching to fit the artists' desired cartoon style and inevitably consume extensive effort.
In contrast, deep cartoonization~\cite{chen2018cartoongan,chen2019animegan,wang2020learning}, including ours, can stylize images into a diverse range of styles with superior quality, thus reducing the need for thorough retouching.

\smallskip
\noindent\textbf{Limitation.}
Although \texttt{Cartooner} successfully demonstrates an interactive to the cartoonization task, other properties could be incorporated to turn into a more artist-friendly solution.
In our study, we have focused on color and texture, as these are (in our early interviews) the most needed aspects by artists.
Nevertheless, the lack of other features might limit its usability in many scenarios.
For example, stroke shape (or style) control, sky synthesis (\Fref{fig:background_process}), vectorization, or layer decomposition which are commonly used by artists, would greatly enhance the workflow.

\section{Additional results}
\Fref{fig:interactivity_suppl} displays interactive scenarios and \Fref{fig:comp_suppl_one}, \ref{fig:comp_suppl_two}, \ref{fig:comp_suppl_three}, \ref{fig:comp_suppl_four} show qualitative comparison.

\figInteractivitySuppl

\figCompSupplOne
\figCompSupplTwo
\figCompSupplThree
\figCompSupplFour